\documentclass[10pt,final,journal]{IEEEtran}
\usepackage{cite}
\usepackage{graphicx}
\usepackage{amsmath,amssymb,amsfonts}
\DeclareMathOperator{\Tr}{Tr}
\usepackage[linesnumbered,ruled,vlined]{algorithm2e}
\usepackage{array}
 \usepackage{mathrsfs}
\usepackage{textcomp}
\usepackage{xcolor}
\usepackage{caption}
\usepackage{bbm}
\usepackage{subcaption}
\usepackage{booktabs}
\usepackage{multirow}
\usepackage{siunitx}
\usepackage{blindtext}
\usepackage{enumitem}
\usepackage[utf8]{inputenc}
\usepackage[T1]{fontenc}
\usepackage{nomencl}
\usepackage{bibunits}
\newcommand{\overbar}[1]{\mkern 3.5mu\overline{\mkern-3.5mu#1\mkern-3.5mu}\mkern 3.5mu}
\makenomenclature
\usepackage{amsmath,amssymb,amsfonts,bm,mathtools}
\usepackage{cancel}
\usepackage{algorithmic}
\usepackage{graphicx}
\usepackage{subcaption}
\usepackage{textcomp}
\usepackage{xcolor}
\usepackage{bbold}
\usepackage{mwe}
\newcommand{\E}{\mathrm{E}}

\usepackage{amsthm}
\theoremstyle{definition}

 \theoremstyle{plain}

 \newtheorem{theorem}{Theorem}[section]

\newtheorem{lemma}[theorem]{Lemma}
\newtheorem{property}{Property}[section]
\theoremstyle{remark}

\newtheorem*{remark*}{Remark}

\usepackage[ruled,vlined]{algorithm2e}
\usepackage{array}
\usepackage{booktabs}
\usepackage{nomencl}
\makenomenclature
\def\BibTeX{{\rm B\kern-.05em{\sc i\kern-.025em b}\kern-.08em
    T\kern-.1667em\lower.7ex\hbox{E}\kern-.125emX}}
    
    \usepackage{amsfonts}

\begin{document}
%\linenumbers
\title{Decentralised Variational Inference Frameworks for Multi-object Tracking on Sensor Networks}
\author{Qing~Li*,~\IEEEmembership{Member,~IEEE}, Runze~Gan*,~\IEEEmembership{Member,~IEEE}, and Simon~J.~Godsill,~\IEEEmembership{Fellow,~IEEE}

%\thanks{This research is sponsored by the US Army Research Laboratory and the UK MOD University Defence Research Collaboration (UDRC) in Signal Processing under the SIGNeTS project. It is accomplished under Cooperative Agreement Number W911NF-20-2-0225. The views and conclusions contained in this document are of the authors and should not be interpreted as representing the official policies, either expressed or implied, of the Army Research Laboratory, the MOD, the U.S. Government or the U.K. Government. The U.S. Government and U.K. Government are authorised to reproduce and distribute reprints for Government purposes notwithstanding any copyright notation herein. For the purpose of open access, the author has applied a Creative Commons Attribution (CC BY) license to any Author Accepted Manuscript version arising.}
\thanks{ Q. Li, R. Gan, and S. J. Godsill are with the Engineering Department, University of Cambridge, Cambridge CB2 1PZ, U.K. e-mail: \{ql289, rg605, sjg30\}@cam.ac.uk. *Q. Li and R. Gan contribute equally to this work.}%
\vspace{-1.2em}
}
%\markboth{IEEE TRANSACTIONS ON Signal Processing}%
\maketitle
%\begingroup\renewcommand\thefootnote{\textsection}

\begin{abstract}
This paper tackles the challenge of multi-sensor multi-object tracking by proposing various decentralised Variational Inference (VI) schemes that match the tracking performance of centralised sensor fusion with only local message exchanges among neighboring sensors. 
We first establish a centralised VI sensor fusion scheme as a benchmark and analyse the limitations of its decentralised counterpart, which requires sensors to await consensus at each VI iteration. 
Therefore, we propose a decentralised gradient-based VI framework that optimises the Locally Maximised Evidence Lower Bound (LM-ELBO) instead of the standard ELBO, which reduces the parameter search space and enables faster convergence, making it particularly beneficial for decentralised tracking.
This proposed framework is inherently self-evolving, improving with advancements in decentralised optimisation techniques for convergence guarantees and efficiency. 
Further, we enhance the convergence speed of proposed decentralised schemes using natural gradients and gradient tracking strategies. Results verify that our decentralised VI schemes are empirically equivalent to centralised fusion in tracking performance. Notably, the decentralised natural gradient VI method is the most communication-efficient, with communication costs comparable to suboptimal decentralised strategies while delivering notably higher tracking accuracy.
%To begin with, we establish a centralised VI sensor fusion scheme as a benchmark; based on it, a fully decentralised counterpart  is developed based on a consensus algorithm, which cuts communication costs by transmitting global statistics instead of raw sensor data. To further reduce  communication load and allow sensors to operate independently without awaiting consensus, we first construct a Locally Maximised Evidence Lower BOund (LM-ELBO) in place of standard ELBO and then present a decentralised gradient-based VI framework to optimise the LM-ELBO. 
%notably, among them, the decentralised natural gradient VI method is the most communicationally efficient, offering cost-effectiveness comparable to suboptimal decentralised fusion strategies while delivering much higher tracking accuracy.
\end{abstract}

\IEEEpeerreviewmaketitle

\begin{bibunit}[IEEEtran]
\section{Introduction} \label{sec:intro}
%The integration of data from multiple sensors offers superior object tracking performance over single-sensor deployments, particularly in complex environments with heavy clutter or low detection probability of targets.Centralised sensor fusion that receives data from all sensors at a fusion center node, is considered the theoretical optimal and ideal, whereas this approach may not always be viable, especially in situations where communication bandwidth is limited. Dynamic changes in sensor networks, such as unexpected disruptions or communication link failures, as well as temporal and spatial misalignments between sensors, add further layers of complexity to data integration tasks.
Integrating data from multiple sensors significantly enhances object tracking performance, especially in complex environments with heavy clutter or low target detection probability. While centralised fusion is optimal, it is often impractical due to limited bandwidth. Several distributed sensor fusion and tracking schemes have been developed, including the decentralised Kalman Filter (KF) \cite{rao1993fully}, which is mathematically equivalent to the centralised KF whilst it is %not scalable and 
confined to a specific complete network with a all-to-all information flow. %Later, more scalable distributed KF algorithms were designed for tracking a single target with only local communications \cite{olfati2005distributed, cattivelli2008diffusion}. These distributed KFs were further extended to handle data association e.g., using the suboptimal joint probabilistic data association (JPDA) \cite{sandell2008distributed} for tracking multiple objects in clutter. A detailed review of distributed KF methods can refer to \cite{mahmoud2013distributed,chong2017forty}. 
Later, more scalable distributed KF algorithms were designed for tracking targets with only local communications \cite{olfati2005distributed, sandell2008distributed,chong2017forty}. 
Distributed particle filters have also been extensively studied for nonlinear and non-Gaussian scenarios, while suboptimal fusion rules and 
approximations such as Gaussian mixtures are applied to alleviate heavy communication overheads \cite{coates2004distributed,oreshkin2010asynchronous}. 
%For  nonlinear and non-Gaussian scenarios, extensive study has been conducted on distributed particle filters;nevertheless, they usually encounter significant communication overheads that necessitates the use of suboptimal fusion rules and/or approximation methods such as Gaussian mixtures to ensure feasibility . 
An alternative way to construct optimal fusion is a single-target optimal track-to-track fusion in \cite{chong1990distributed}, under the assumption of known correlations of the information between sensors. It was later embedded to a Multi-Hypothesis Tracker (MHT) \cite{chong2017forty} and a Random Finite Set (RFS) framework \cite{clark2010robust}. However, these cross-correlations practically are unknown or computationally intractable in real applications. 

To prevent double counting of common information, two suboptimal fusion rules, Generalised Covariance Intersection (GCI) \cite{clark2010robust} and Arithmetic Average (AA) \cite{li2017generalized,li2021distributed}, were introduced and integrated with existing multi-object trackers for sensor fusion with unknown correlations, where local multi-object
distributions are fused using GCI or AA rules.
%Hence, two suboptimal fusion rules, Generalised Covariance Intersection (GCI) \cite{clark2010robust} and Arithmetic Average (AA) \cite{li2017generalized,li2021distributed} were proposed to avoid the double counting of common information, and were integrated into existing multi-object trackers to facilitate sensor fusion with unknown correlations, whereeach sensor employs a local tracker to generate a local posterior, which are then fused using either GCI or AA fusion techniques.
Specifically, the GCI and AA rules have been successfully tailored for RFS trackers including probability hypothesis density (PhD) filter \cite{ueney2013distributed}, multi-Bernoulli (MB) filter \cite{li2020arithmetic} 
and others \cite{li2018computationally}. %wang2016distributed,
%However, a closed form solution of GCI fusion is normally unavailable, making it rely on approximation strategies that may compromise the estimation accuracy. Additionally, it was reported in \cite{li2020arithmetic} that GCI fusion is susceptible to missed detections and may degrade with the increase of the sensor number. 
%In comparison, AA fusion is shown to have lower computational costs and outperformed the GCI fusion in low detection probability environments \cite{li2020arithmetic}. 
%The above-mentioned fusion strategies, however, only apply to scenarios with either two sensors or a predetermined communication regime. Therefore, %
Later, consensus-based algorithms \cite{olfati2004consensus,xiao2005scheme} were introduced to perform GCI and AA fusion in a fully distributed manner without the need for a predefined communication protocol.
%To perform fusion in a fully distributed manner without a predefined communication protocol, consensus-based algorithms \cite{olfati2004consensus,xiao2005scheme} were introduced to GCI and AA fusion. 
Nonetheless, these methods that fuse local sensors' posteriors are suboptimal, often leading to reduced tracking accuracy.

%To advance the tracking and fusion performance beyond the capabilities of existing methods, both the deployment of an accurate multi-object tracking algorithm and an low-cost, decentralised towards-optimal sensor fusion strategy are equally critical.
Advancing multi-sensor multi-object tracking performance requires both deploying accurate trackers and developing low-cost, near-optimal sensor fusion strategies.
With regards to the tracking performance, the Variational multi-object Tracker (VT) \cite{gan2022variational,gan2024variational} demonstrated superior performance over other leading tracking algorithms \cite{meyer2020scalable,granstrom2019poisson,li2023adaptive} in single-sensor scenarios with a fixed object number and Non-Homogeneous Poisson Process (NHPP) measurement model \cite{gilholm2005poisson}. As for sensor fusion methods, a fully decentralised counterpart of the centralised multi-sensor VT was developed in \cite{li2023consensus} which, in theory, achieves results equivalent to the centralised fusion scheme while enabling sensors to operate locally with only communication to neighboring sensors. However, each sensor has to wait for the consensus algorithm to converge at each variational inference (VI) iteration, and thus may lead to a substantial communication cost. %This delay can become problematic if sensors struggle to achieve consensus due to network structure or communication challenges.

This paper presents a comprehensive decentralised variational inference framework for tracking a fixed number of objects in clutter with a dynamic sensor network. Compared to \cite{li2023consensus}, it is much more flexible and enables sensor to work independently without awaiting consensus during variational inference iterations.  A streamlined introduction of this method was presented in our preliminary conference paper \cite{li2024}, which, however, only provided the implementation and evaluation of the decentralised natural gradient VT algorithm with no detailed derivations and analysis. This paper extends this preliminary work with the following novel contributions. 

\vspace{-1.2em}
\subsection{Contribution}
\vspace{-.3em}
We propose a decentralised gradient-based variational inference scheme and extend it to a sequential context for a multi-sensor multi-object tracking application. Relevant work in distributed variational inference in a general setting can be found in \cite{hua2015distributed}, in which it directly decomposed the standard ELBO and adopted a stochastic gradient approximation, %for local calculations
although, as the authors pointed out, it lacks solid theoretical analysis, and indeed is not applied in a dynamic scenario over time or for tracking models.
%{This paper proposes variational Bayes method based upon a Locally Maximised Evidence Lower Bound (LM-ELBO), developing a decentralised gradient-based variational inference scheme within the context of a multi-sensor multi-object tracking application. Relevant work in distributed variational inference in a general setting can be found in \cite{hua2015distributed}, in which it directly decomposed the standard global ELBO and adopted  a stochastic gradient approximation for local calculations, although, as the authors pointed out, it does not have the solid theoretical analysis, and indeed is not applied in a dynamic scenario over time or for tracking models.}
Here, by contrast, we form a decentralised optimisation problem of optimising a Locally Maximised Evidence Lower Bound (LM-ELBO), an objective that we demonstrate to be equivalent to the original ELBO. Then we show how to decompose this LM-ELBO into local LM-ELBOs, and thus decentralised gradient-based methods can be applied with guaranteed convergence under specified conditions, see Section \ref{Decentralised Gradient Descent} for details.
%Here, by contrast, we form a decentralised optimisation problem by first demonstrating the equivalence of optimisation solutions between the LM-ELBO and the original ELBO, and then show how to decompose this into local LM-ELBOs. In this way,  decentralised gradient-based methods can be applied with guaranteed convergence under specified conditions, see Section \ref{Decentralised Gradient Descent} for detailed discussion.
The construction of the LM-ELBO %, which is rigorously justified in comparison to original ELBO, 
is particularly advantageous for reducing communication costs in the distributed sensor fusion and tracking task, since it eliminates the need for communication of high-dimensional data association information, whose dimensionality increases with the number of measurements.

With respect to algorithmic development, we propose three novel implementations of decentralised (natural) gradient-based variational multi-object trackers, compared to the conference paper \cite{li2024}. 
Firstly, we propose a decentralised gradient descent VT that has convergence guarantee with specified diminishing step sizes in \cite{chang2020distributed}. 
Secondly, a gradient tracking scheme is applied to improve its convergence speed, which also guarantees convergence when using a constant stepsize \cite{chang2020distributed,nedic2017achieving}. 
%Firstly, we propose a Decentralised Gradient descent VT with Diminishing Stepsize (DeG-VT-DS) that has theoretical convergence guarantee with specified diminishing step sizes in \cite{chang2020distributed}. Secondly, we propose a Decentralised Gradient descent VT with Gradient Tracking (DeG-VT-GT) to improve its convergence speed; this gradient tracking scheme also guarantees convergence when using a constant stepsize \cite{chang2020distributed}. 
Moreover, we integrate the natural gradients \cite{amari1998natural}, in place of standard gradients to further accelerate convergence. %Moreover, we integrate the natural gradient \cite{amari1998natural} to further speed up convergence. 
All proposed algorithms are provided with detailed algorithmic derivations and performance analysis in this paper.
A minor algorithmic contribution is that we present the detailed derivations of Coordinate Ascent Variational Inference (CAVI) updates of the centralised VT in supplementary documents for our prior work \cite{li2023consensus}.

{Besides practical tracking applications, we contribute to the general variational inference in the following aspects.}
The concept of LM-ELBO has been introduced in \cite{hensman2012fast,hoffman2013stochastic} under different names but with limited discussion and application in the literature. 
Here we provide our definition, connect it to existing notions, and introduce key properties. For LM-ELBO, we provide more concise proofs of known properties and present new ones, including validating its use in place of the original ELBO.
Moreover, we prove for the first time that local LM-ELBOs, decomposed from LM-ELBO, inherit several useful properties, including one that simplifies local gradient computation.
Further, we propose a flexible decentralised gradient-based variational inference framework that
can be directly applied to other general tasks defined in Section \ref{sec:ELBO problem setting} and similar system models with global and local variables e.g., in \cite{hoffman2013stochastic}, beyond %multi-sensor multi-object
tracking applications. Most importantly, this framework is self-evolving, 
%allowing to leverage emerging decentralised optimisation techniques to enhance the convergence guarantees and efficiency of these decentralised gradient-based variational inference algorithms. 
allowing to use emerging decentralised optimisation techniques to enhance convergence guarantees and algorithmic efficiency.

Finally, compared to \cite{li2024}, this paper presents extensive comparative analysis for newly-proposed methods
%in convergence speed, tracking accuracy and communication costs across various scenarios with different object and sensor number, detection environment, as well as connectivity and modalities of a sensor network. C
% we add experiments of newly-proposed methods, and conduct a comprehensive analysis of 
with regard to convergence speeds to demonstrate the benefits of incorporating the natural gradient and gradient tracking strategies. %As a complement to the prior work \cite{li2023consensus}, we validate its convergence in time-varying sensor network scenarios. % whereas in \cite{li2023consensus} it only shows the case of a small number of sensors with fixed connectivity. 
 Moreover, % to test their adaptivity, 
we analyse the proposed methods in heterogeneous sensor networks with varying detection and clutter rates, extending our scenario in \cite{li2024}. %All the experiments are conducted in 50 Monte Carlo runs to verify their robustness.  
Simulation results demonstrate that all proposed decentralised (natural) gradient VTs can achieve empirically equivalent tracking performance to centralised fusion. Particularly, decentralised natural gradient descent VTs require lower communication cost than method  in \cite{li2023consensus} and are much accurate than suboptimal fusion techniques under comparable communication cost. 

\vspace{-1.5em}
\subsection{Paper Outline}
\vspace{-0.2em}
Section II presents problem settings and a variational filtering framework for multi-sensor multi-object tracking. 
Section III introduces VI and the standard ELBO, outlines centralised and decentralised VI for tracking and their limitations. 
%Section III introduces the standard ELBO for general coordinate ascent and gradient-based VI, details centralised and decentralised coordinate ascent VI for tracking tasks, and motivates the use of LM-ELBO over the standard ELBO for decentralised gradient-based VI methods. 
Section IV explores the rationale, concept and properties of LM-ELBO, based on which a flexible decentralised gradient-based VI framework is designed for sensor fusion, and local LM-ELBO properties are presented in Section V. Implementations of the distributed multi-object trackers are given in Section VI. Sections VII and VIII are results and conclusions.

\vspace{-.5em}
\section{Problem Formulation and Modelling} \label{sec:PoissonModel}
This paper considers tracking multiple targets in clutter under a distributed sensor network. % where the communication links between sensors can be time-varying. 
Assume that there are $K$ objects in the surveillance area and $K$ is known. At each time step $n$, their joint state is $X_n=[X_{n,1}^\top,X_{n,2}^\top,...,X_{n,K}^\top]^\top$, where each vector $X_{n,k}, k\in \{1,...,K\}$ denotes the kinematic state for the $k$-th object. Suppose that objects are observed by $N_s$ sensors, each capable of observing the entire surveillance area. The time-varying sensor network at time step $n$ can be modelled as a graph $\mathcal{G}(n) = \{\mathcal{S}, \mathcal{E}(n)\}$, where $\mathcal{S} = \{1, 2, \ldots, N_s\}$ is the sensor set, and $\mathcal{E}(n)$ is the edge set with edge $(i, j)$ meaning that the $i$-th sensor can communicate with the $j$-th sensor. The set of neighbours of sensor $i$ is $\mathcal{N}_i(n) = \{j \mid (i, j) \in \mathcal{E}(n)\}$, and the degree of the $i$-th sensor is $d_i(n) = |\mathcal{N}_i(n)|$. 
% Assume that there are $K$ targets in the surveillance area. At each discrete time step $n$, their joint state is $X_n=[X_{n,1}^\top,X_{n,2}^\top,...,X_{n,K}^\top]^\top$, where each vector $X_{n,k}, k\in \{1,...,K\}$ denotes the kinematic state for the $k$-th target.
% Suppose that the targets are observed by a sensor network consisting of $N_s$ sensors, each capable of observing the entire tracking area. The time-varying sensor network at time $t$ can be modelled as a graph $G(t) = \{\mathcal{S}, \mathcal{E}(t)\}$ at any given continuous time $t$, where the sensor set is denoted by $ \mathcal{S} = \{1, 2, \ldots, N_s\}$, and $\mathcal{E}(t)$ is the set of edges with the existence of edge $(i, j)$ meaning that the $i$-th sensor can communicate with the $j$-th sensor at time $t$. The set of neighbours of sensor $i$ is denoted by $\mathcal{N}_i(t) = \{j \mid (i, j) \in \mathcal{E}(t)\}$. The degree $d_i(t)$ of the $i$-th sensor represents the number of its neighbouring sensors with which it can communicate, i.e., $d_i(t) = |\mathcal{N}_i(t)|$. In a sensor network, the measurements received from all sensors at time step $n$ can be denoted by $Y_n=[Y_{n}^{1},Y_{n}^{2},...,Y_{n}^{N_s}]$. Each $Y_{n}^{s}$ includes measurements acquired by the $s$-th sensor, and $Y_{n}^{s}=[Y_{n,1}^{s},...,Y_{n,M_n^{s}}^{s}]$, where $M_n^{s}$ is the total number of measurements received at the $s$-th sensor ($s =1,..., N_s$). Subsequently, $M_n=[M_n^1,...,M_n^{N_s}]$ records the total number of measurements received from all sensors at time step $n$.
\vspace{-1.4em}
\subsection{Dynamical  Model}\label{dynamic model}
\vspace{-0.2em}
We assume that targets move in a 2D surveillance area with each having state $X_{n,k}=[x^1_{n,k}, \Dot{x}^1_{n,k},x^2_{n,k}, \Dot{x}^2_{n,k}]^T$, where $x^d_{n,k}$ and $\Dot{x}^d_{n,k}$ (here $d\in\{1,2\}$ although extension to higher dimensions is straightforward) indicate the $k$-th target's position and velocity in the $d$-th dimension, respectively. We assume an independent linear Gaussian transition density:% for each target's states:
\vspace{-0.4em}
\begin{equation}\label{eq: dynamic transition}
    p(X_n|X_{n-1})
    =\prod\nolimits_{k=1}^K\mathcal{N}(X_{n,k};F_{n,k}X_{n-1,k},Q_{n,k}). \\[-0.4em]
\end{equation}
where $F_{n,k}=diag(F^1_{n,k},F^2_{n,k})$, $Q_{n,k}=diag(Q^1_{n,k},Q^2_{n,k})$. 
%For a constant velocity (CV) model, $F_{n,k}^d,Q_{n,k}^d$ ($d=1,2$) are
% \vspace{-0.5em}
% \begin{equation} \label{eq:model para}
%     F_{n,k}^d=\begin{bmatrix} 1 & \tau \\ 0& 1
%     \end{bmatrix}, 
%     Q_{n,k}^d=\sigma_k^2\begin{bmatrix} \tau^3/3 & \tau^2/2 \\ \tau^2/2& \tau
%     \end{bmatrix}, 
% \end{equation}
% where $\tau$ is the time interval between time steps.
\vspace{-1em}
\subsection{NHPP Measurement Model and Association Prior}\label{NHPP measurement model and association prior}
Denote the measurements received from all sensors at time step $n$ be $Y_n=[Y_{n}^{1},Y_{n}^{2},...,Y_{n}^{N_s}]$. Each $Y_{n}^{s}$ includes measurements acquired by the $s$-th sensor, and $Y_{n}^{s}=[Y_{n,1}^{s},...,Y_{n,M_n^{s}}^{s}]$, where $M_n^{s}$ is the total number of measurements received at the $s$-th sensor ($s =1,..., N_s$). Subsequently, $M_n=[M_n^1,...,M_n^{N_s}]$ records the total number of measurements received from all sensors at time step $n$.
Here, we assume each sensor independently detects objects in accordance with the NHPP measurement model as detailed in \cite{li2023consensus,gilholm2005poisson}. Notably, the NHPP model may vary for each sensor.
Denote the set of Poisson rates for all sensors as $\Lambda=[\Lambda^{1},\Lambda^{2},...,\Lambda^{N_s}]$. For each sensor $s$, the Poisson rate vector is defined by $\Lambda^{s}=[\Lambda_0^{s},\Lambda_1^{s},...,\Lambda_K^{s}]$, where $\Lambda_0^{s}$ is the clutter rate and $\Lambda_k^{s}$ is the $k$-th object rate, $k=1,...,K$.  
%For each sensor $s$, each target $k$ generates measurements by a NHPP with a Poisson rate $\Lambda_k^{s}$, and the total measurement process is also a NHPP from the superposition of the conditional independent NHPP measurement process from $K$ targets and clutter.
The total number of measurements from the $s$-th sensor follows a Poisson distribution with a rate of $\Lambda_{sum}^{s}=\sum_{k=0}^K\Lambda_k^{s}$. 
Our independent measurement model assumption implies that given $X_{n}$, the measurements of each sensor are conditionally independent, i.e., $p(Y_n|X_n)=\prod_{s=1}^{N_s} p(Y_{n}^{s}|X_{n})$.
% , in other words, the joint likelihood function can be factorised as the product of all local likelihood functions
% Define the local likelihood function at each sensor $s$ as $p(Y_{n}^{s}|X_{n})$. The joint likelihood function is the product of all local likelihood functions since all $Y_{n}^{s}$ are assumed conditionally independent given $X_{n}$, that is, $p(Y_n|X_n)=\prod_{s=1}^{N_s} p(Y_{n}^{s}|X_{n})$.
% Given $Y_n$, the associations at time step $n$ are denoted by $\theta_{n}=[\theta_{n}^{1},\theta_{n}^{2},...,\theta_{n}^{N_s}]$
We denote the associations of all measurements $Y_n$ by $\theta_{n}=[\theta_{n}^{1},\theta_{n}^{2},...,\theta_{n}^{N_s}]$, with each $\theta_{n}^{s}=[\theta_{n,1}^{s},\theta_{n,2}^{s},...,\theta_{n,M_n^{s}}^{s}]$ ($s=1,...,N_s$) representing the association vector for the $s$-th sensor's measurements. Each component $\theta_{n,j}^{s}$ ($j=1,...,M_n^{s}$) gives the origin of the measurement $Y_{n,j}^{s}$; $\theta_{n,j}^{s}=0$ indicates that $Y_{n,j}^{s}$ is generated by clutter, and $\theta_{n,j}^{s}=k$ ($k=1,...,K$) means that $Y_{n,j}^{s}$ is generated from the target $k$. The adopted conditionally independent NHPP measurement model leads to the following properties. First, $p(Y_n,\theta_n|X_n,M_n)=p(Y_n|\theta_n,X_n)p(\theta_n|M_n)$. Both joint association prior and joint likelihood are conditionally independent across sensors, and measurements are conditionally independent given associations and states, i.e., 
\vspace{-0.5em}
\begin{align}\label{eq: joint association prior conditionally independent}
   & p(\theta_n|M_n) =\prod\nolimits_{s=1}^{N_s}  p(\theta_{n}^{s}|M_n^{s})\\[-0.2em]\label{eq: joint likelihood conditionally independent}
  &  p(Y_n|\theta_n,X_n)=\prod\nolimits_{s=1}^{N_s} p(Y_{n}^{s}|\theta_{n}^{s},X_{n})\\[-.1em]
    \label{eq:obs prior}
&p(Y_{n}^{s}|\theta_{n}^{s},X_{n})=\prod\nolimits_{j=1}^{M_n^{s}}\ell^s(Y_{n,j}^{s}|X_{n,\theta_{n,j}^{s}})\\[-2em]\notag
\end{align}
where $M_n^{s}$ is implicitly known from $\theta_n^{s}$ since $M_n^{s}=|\theta_n^{s}|$, and $\ell^s$ is the probability density function of a single measurement received in sensor $s$ given its originator's state. Here we assume a linear and Gaussian model for object originated measurements and clutter measurements to be uniformly distributed in the observation area of volume $V^{s}$:
\vspace{-0.4em}
\begin{equation} \ell^s(Y_{n,j}^{s}|X_{n,k})=\begin{cases} 
    \mathcal{N}(H X_{n,k},R_{k}^{s}),& \text{$k\neq 0 $; \ \ (object)}\\
     {1}/{V^{s}}, & \text{$k= 0 $; \ \ (clutter)}
\end{cases} 
\\[-0.5em]
\label{measurement model}
\end{equation}
where $H$ is the observation matrix, and $R_{k}^{s}$ indicates the $s$-th sensor noise covariance. Moreover, the joint prior $p(\theta_{n}^{s}|M_n^{s})$ can be factorised as the product of $M_n^{s}$ independent association priors, i.e., $p(\theta_{n}^{s}|M_n^{s})=\prod_{j=1}^{M_n^{s}} p(\theta_{n,j}^{s})$,
where $p(\theta_{n,j}^{s})$ is a categorical distribution with $\theta_{n,j}^{s} \in \{0,...,K\}$
\vspace{-0.3em}
\begin{align}
    \label{eq:single assoc prior}    &p(\theta_{n,j}^{s})=\frac{1}{\sum_{k=0}^K\Lambda_k^{s}}\sum\nolimits_{k=0}^K\Lambda_k^{s}\delta[\theta_{n,j}^{s}=k]. \\[-2em]\notag
\end{align}
\subsection{Variational Filtering for Multi-object Tracking}\label{sec:Sequential Bayesian inference for multi-object tracking}
%With the existence of measurement origin uncertainty,  a
A Bayesian object tracker aims recursively to estimate the posterior $p(X_n,\theta_n|Y_{1:n})$ of object states and associations based on the noisy measurements ${Y}_{1:n}$. Assume that $K,\Lambda,$ and $R_{1:K}^s$ are known parameters. %received at each time step $n$ from all sensors.
%The objective is to sequentially estimate the posterior $p(X_n,\theta_n|Y_{1:n})$ at time step $n$ given observations ${Y}_{1:n}$ from all sensors. 
Accordingly, the exact optimal filtering can be recursively expressed as the following prediction and update steps:
\vspace{-0.3em}
\begin{align} \label{eq:prediction}
    &\hspace{-0.5em} p(X_n|{Y}_{1:n-1})= \int p(X_n|X_{n-1}) p({X}_{n-1}|{Y}_{1:n-1})d{X}_{n-1},\\[-0.2em] 
    &\hspace{-0.5em} p(X_n,\theta_n|{Y}_{1:n})\propto
    p(Y_n|\theta_n,X_n) p(\theta_n|M_n) p(X_n|{Y}_{1:n-1}).\\[-1.8em]\notag
\end{align}
%where $p(X_n|{Y}_{1:n})$ and $p(X_n,\theta_n|{Y}_{1:n})$ are the predictive prior and posterior, respectively. 
However, with the association uncertainty, the exact filtering recursion is intractable even in linear Gaussian systems. Thus approximate inference, here variational filtering, is adopted.
\subsubsection{{Prediction step in the variational filtering}}at time step $n$,  the predictive prior $\hat{p}_n(X_n)$ is computed as follows 
\vspace{-0.5em}
\begin{equation}  \label{eq: predictive prior}
     \hat{p}_n(X_n) = \int p(X_n|X_{n-1})q^*_{n-1}(X_{n-1})dX_{n-1}, \\[-0.6em]
\end{equation}
where we replace $p({X}_{n-1}|{Y}_{1:n-1})$ in \eqref{eq:prediction} with the converged variational distribution $q^*_{n-1}({X}_{n-1})$ obtained with variational inference at time step $n-1$. 
{Specifically, assume that the converged variational distribution is in an independent Gaussian form, i.e., $q_{n-1}^*(X_{n-1})=\prod_{k=1}^K q_{n-1}^*(X_{n-1,k})$, and $q^*_{n-1}(X_{n-1,k})= \mathcal{N}(X_{n-1,k};\mu^{k*}_{n-1|n-1},\Sigma^{k*}_{n-1|n-1})$. Given  the linear Gaussian transition in \eqref{eq: dynamic transition},
its predictive prior $\hat{p}_n(X_n)$ is also in an independent Gaussian form, i.e., $\hat{p}_n(X_n)=\prod_{k=1}^K \hat{p}_n(X_{n,k})$, where for each object $k$, we have}
\vspace{-1.5em}
\begin{align}\label{eq:predictive prior computation}  \hat{p}_n(X_{n,k})=&\mathcal{N}(X_{n,k};\mu^{k*}_{n|n-1},\Sigma^{k*}_{n|n-1}),
    \\[-.2em]\notag
    \mu^{k*}_{n|n-1}=&F_{n,k}\mu^{k*}_{n-1|n-1},\\[-.2em]\notag
    \Sigma^{k*}_{n|n-1}=&F_{n,k}\Sigma^{k*}_{n-1|n-1}F_{n,k}^\top+Q_{n,k}.
    \\[-2.2em]\notag
\end{align}
\subsubsection{{Update step in the variational filtering}}
Subsequently, the target posterior at the update step is   
\vspace{-0.3em}
\begin{align}  \label{eq:phatjoint}
    \hat{p}_n(X_{n},\theta_{n}|Y_n)  
    &\propto p(Y_n|\theta_n,X_n)p(\theta_n|M_n)\hat{p}_n(X_n).\\[-1.8em]\notag
\end{align}
In the following sections, we will elaborate on the variational inference for inferring this target posterior $\hat{p}_n(X_{n},\theta_{n}|Y_n)$.

\vspace{-0.3em}
\section{Variational Inference with Standard ELBO for Multi-object Tracking}
This section first introduces the standard free-form and fixed-form ELBOs for CAVI and gradient-based variational inference, respectively. Then, we present centralised CAVI for tracking tasks and discuss the limitation of its decentralised version. Lastly, we provide the fixed-form ELBO for tracking and motivate the methods proposed later.
\vspace{-1.em}
\subsection{Variational Inference for General Problem Settings}\label{sec:ELBO problem setting}
\vspace{-0.02em}
% Suppose we can represent all variables we wish to infer as two disjoint (multivariate) variables $X,\theta$, and we denote the received measurements as $Y$. 
Consider a general task of inferring the posterior of disjoint multivariate variables $X,\theta$ given measurements $Y$, where the exact posterior $p(X,\theta|Y)$ is intractable but can be evaluated up to a constant, i.e., $p(X,\theta|Y)\propto f(X,\theta,Y)$ and the unnormalised posterior $f(X,\theta,Y)$ is pointwise computable. With a mean-field assumption $q(X,\theta)=q(X)q(\theta)$, $p(X,\theta|Y)$ can be inferred by variational inference \cite{blei2017variational}, which aims to find $q(X)$ and $q(\theta)$ from the posited family that minimises the Kullback–Leibler (KL) divergence, or equivalently, maximises the evidence lower bound (ELBO) \cite{blei2017variational} as follows
\vspace{-0.5em}
\begin{align} \label{eq: ELBO general 1}
    \mathcal{F}(q(X),q(\theta))\coloneqq\E_{q(X)q(\theta)}\log\frac{ f(X,\theta,Y)}{q(X)q(\theta)}\,. \\[-2em]\notag
\end{align}
%since the ELBO is $-\text{KL}(q(X)q(\theta)||p(X,\theta|Y))$ plus a constant.
% can be expressed as a constant minus the $\text{KL}(q(X)q(\theta)||p(X,\theta|Y))$.
% $-\text{KL}(q(X)q(\theta)||p(X,\theta|Y))$ plus a constant. 
% Applying Jensen's inequality yields:
% \vspace{-0.5em}
% \begin{align} \label{eq: jensen}
%     \mathcal{F}(q(X),q(\theta))\leq \log \int f(X,\theta,Y) dXd\theta, \\[-2em]\notag
% \end{align}
% where the right hand side equals $\log p(Y)$ if $f$ represents $p(X,\theta,Y)$, highlighting the term ELBO. 
%Note that the ELBO in \eqref{eq: ELBO general} is a functional with arguments $q(X),q(\theta)$. 
Such a definition of free-form ELBO allows variational distributions $ q(X),q(\theta) $ taking any form, %to find the $ q(X),q(\theta) $ that maximise the ELBO, 
and we can apply Coordinate Ascent Variational Inference (CAVI) to find the $ q(X),q(\theta) $ that maximise the ELBO, where the ELBO in \eqref{eq: ELBO general 1} is optimised by iteratively updating one of the variational distributions while keeping the other fixed. 
%In this case, even though each variational distribution, e.g., $q(\theta)$, is allowed to take any form, there will exist a unique global optimiser for $\mathcal{F}$ while fixing $q(X)$ \cite{blei2017variational}:
{For example, for variational distribution $q(\theta)$,  according to \cite{blei2017variational}, the global optimiser while fixing $q(X)$ is:}
\vspace{-0.5em}
\begin{align} \label{eq:optimal CAVI update}
    &q^*(\theta)\propto \exp \left(E_{q(X)} \log f(X,\theta,Y)\right),\\ \label{eq: global optimal maximum}
    &\mathcal{F}(q(X),q^*(\theta))=\max_{q(\theta)}\mathcal{F}(q(X),q(\theta)), \\[-2.3em]\notag
\end{align}
with the maximisation spanning all possible distributions $q(\theta)$. %It is noteworthy that the uniqueness is in terms of the distribution itself other than its parameters, as the same distribution can have different parametrisations. 
Such a CAVI update, however, is not always applicable or easy to implement since it requires calculating the closed form global optimiser as in \eqref{eq:optimal CAVI update}. %A detailed development of CAVI update for our tracking task will be given in Section \ref{sec:ca update}.

Alternatively, if we further assume the distribution forms of $q(X),q(\theta)$, and denote their respective governing parameters by vectors $\lambda,\rho$, then the ELBO in \eqref{eq: ELBO general 1} can be reformulated as a conventional function with vector argument $\lambda,\rho$. This leads us to define the fixed-form ELBO, denoted as $\mathcal{F}(\lambda,\rho)$:
\vspace{-0.5em}
\begin{align} \label{eq: ELBO function}
    \mathcal{F}(\lambda,\rho)\coloneqq\E_{q(X;\lambda)q(\theta;\rho)}\log\frac{ f(X,\theta,Y)}{q(X;\lambda)q(\theta;\rho)}. \\[-2em]\notag
\end{align}
% where the parameters $\eta,\rho$ belong to some subset of $\mathbb R^n$. 
This fixed-form ELBO enables more conventional optimisation techniques, such as gradient descent, particularly useful when the standard CAVI update \eqref{eq:optimal CAVI update} is intractable. %Such a fixed-form ELBO will be used in the proposed decentralised (natural) gradient descent algorithms.

% \vspace{-1em}
% \subsection{Variational Filtering for Multi-object Tracking in Clutter}\label{sec:Variational Filtering for Multi-object Tracking in Clutter}
% This section formulates a variational filtering framework specifically for tracking multiple objects in clutter since the exact filtering recursion as in Section \ref{sec:Sequential Bayesian inference for multi-object tracking} is intractable. The parameters $K,\Lambda,$ and $R_{1:K}^s$ in Section \ref{sec:PoissonModel} are assumed to be known. 

% In the variational filtering at time step $n$,  the prediction step is as follows with the predictive prior $\hat{p}_n(X_n)$ being
% \vspace{-0.5em}
% \begin{equation}  \label{eq: predictive prior}
%      \hat{p}_n(X_n) = \int p(X_n|X_{n-1})q^*_{n-1}(X_{n-1})dX_{n-1}. \\[-1em]
% \end{equation}
% where we replace $p({X}_{n-1}|{Y}_{1:n-1})$ in \eqref{eq:prediction} with the converged variational distribution $q^*_{n-1}({X}_{n-1})$ obtained with variational inference \cite{blei2017variational} at time step $n-1$.

% Subsequently, the target posterior at the update step is   
% \vspace{-0.3em}
% \begin{align}  \label{eq:phatjoint}
%     \hat{p}_n(X_{n},\theta_{n}|Y_n)  
%     &\propto p(Y_n|\theta_n,X_n)p(\theta_n|M_n)\hat{p}_n(X_n),\\[-1.8em]\notag
% \end{align}

\vspace{-1em}
\subsection{Variational Inference Update 
%Methods to Update $\hat{p}_n(X_{n},\theta_{n}|Y_n)$
for Multi-object Tracking}
For tracking tasks with the target posterior $\hat{p}_n(X_{n},\theta_{n}|{Y}_n)$ defined in \eqref{eq:phatjoint}, variational inference introduced in Section \ref{sec:ELBO problem setting} can be applied to approximate $\hat{p}_n(X_{n},\theta_{n}|{Y}_n)$ by a converged variational distribution $q^*_n(X_{n})q^*_n(\theta_{n})$, under a mean-field factorisation of $q_n(X_n,\theta_n)=q_n(X_n)q_n(\theta_n)$.
%, we aim to approximate $\hat{p}_n(X_{n},\theta_{n}|{Y}_n)$ by a converged variational distribution $q^*_n(X_{n})q^*_n(\theta_{n})$ that maximises the ELBO. 

\subsubsection{{Standard free-form ELBO and centralised CAVI for sensor fusion and tracking}}\label{sec:ca update}
In a centralised setup, a central node collects data $Y_n$ from all $N_s$ sensors. For tracking tasks, the free-form ELBO $\mathcal{F}(q(X_n),q(\theta_n))$ in \eqref{eq: ELBO general 1} is 
\vspace{-0.5em}
\begin{align} \label{eq: ELBO general}
    \mathcal{F}(q(X_n),q(\theta_n))&=\E_{q(X_n)q(\theta_n)}\log\frac{ f(X_n,\theta_n,Y_n)}{q(X_n)q(\theta_n)}.\\
    \label{eq:specific unormal posterior}
f(X_n,\theta_n,Y_n)&=p(Y_n|\theta_n,X_n)p(\theta_n|M_n)\hat{p}_n(X_n).\\[-2em]\notag
\end{align}
Subsequently, the optimisation of the ELBO $\mathcal{F}(q(X_n),q(\theta_n))$ in \eqref{eq: ELBO general} can be done by the standard CAVI algorithm \cite{blei2017variational} that iteratively update $q_n(X_n)$ and $q_n(\theta_n)$, which is guaranteed to find a local optimum of the ELBO after convergence. Under the assumptions in Section \ref{sec:PoissonModel}, these updates are all in closed form, and detailed derivations of \eqref{eq: X update}-\eqref{eq:update theta} are given in  Appendix \ref{apx: centralised CAVI derivation} of supplementary material.

\noindent \textbf{Update for $q_n(X_n)$:} 
First, $X_n$ can be updated as follows
\vspace{-0.6em}
\begin{align}  \label{eq: X update}
    q_n(X_n)
 \propto  \hat{p}_n(X_n)\prod\nolimits_{k=1}^{K}\mathcal{N}\left(\overbar{Y}_n^k;HX_{n,k},\overbar{R}^k_n\right),\\[-2.7em]\notag
\end{align}
%$\text{exp}\left(\E_{q_n(\theta_n)}\log p(Y_n|\theta_n,X_n) \right)$$\propto \prod_{k=1}^{K}\mathcal{N}\left(\overbar{Y}_n^k;HX_{n,k},\overbar{R}^k_n\right)$ are given in supplementary materials, and 
%parameters $\overbar{Y}_n^k$ and $\overbar{R}_n^k$ are computed as follows
\begin{align}\label{eq:pseudomeas R c}
 &\overbar{R}_n^k=\Big(\sum\nolimits_{s=1}^{N_s} \big((R_k^{s})^{-1}\sum\nolimits_{j=1}^{M_n^{s}}q_n(\theta_{n,j}^{s}=k) \big) \Big)^{-1}, \\[-0.2em]\label{eq:pseudomeas Y c}
    &\overbar{Y}_n^k=\overbar{R}_n^k \sum\nolimits_{s=1}^{N_s} \Big((R_k^{s})^{-1}\sum\nolimits_{j=1}^{M_n^{s}}q_n(\theta_{n,j}^{s}=k)Y_{n,j}^{s}\Big).\\[-2em]\notag
\end{align}
%%where derivations of parameters $\overbar{Y}_n^k$ and $\overbar{R}_n^k$ are given in supplementary materials.
Given an independent initial Gaussian prior $p(X_0)=\prod_{k=1}^K p(X_{0,k})$ and the transition in \eqref{eq: dynamic transition}, the updated variational distribution can always be in an independent Gaussian form, i.e., $q_n(X_n)=\prod_{k=1}^K q_n(X_{n,k})$. 
Denote the converged variational distribution for the $k$-th target at time step $n-1$ as $q^*_{n-1}(X_{n-1,k})= \mathcal{N}(X_{n-1,k};\mu^{k*}_{n-1|n-1},\Sigma^{k*}_{n-1|n-1})$. 
Then, according to \eqref{eq: predictive prior}-\eqref{eq:predictive prior computation}, the predictive prior $\hat{p}_n(X_n)=\prod_{k=1}^K \hat{p}_n(X_{n,k})$, and $\hat{p}_n(X_{n,k})=\mathcal{N}(X_{n,k};\mu^{k*}_{n|n-1},\Sigma^{k*}_{n|n-1})$. 
Finally. by using equations \eqref{eq: X update}, the variational distribution $q_n(X_{n,k})=\mathcal{N}(X_{n,k};\mu^{k}_{n|n},\Sigma^{k}_{n|n})$ for each object $k$ can be updated independently and in parallel by Kalman filtering.
% \vspace{-0.2em}
% \begin{align}\notag
%     q_n(X_{n,k})&=\mathcal{N}(X_{n,k};\mu^{k*}_{n|n-1},\Sigma^{k*}_{n|n-1})\mathcal{N}\left(\overbar{Y}_n^k;HX_{n,k},\overbar{R}^k_n\right)\\&=\mathcal{N}(X_{n,k};\mu^{k}_{n|n},\Sigma^{k}_{n|n})\\[-2em]\notag
% \end{align} 
% \begin{align}\notag
%      &\mu^{k}_{n|n}=(\text{I}-GH)\mu^{k*}_{n|n-1}+G\overbar{Y}_{n}^k, \quad  G=\Sigma^{k*}_{n|n-1}H^\top S^{k^{-1}}_{n}
%     \\\label{eq:KF update} 
%     &\Sigma^{k}_{n|n} =(I-GH)\Sigma^{k*}_{n|n-1}, ~S^k_{n}=H\Sigma^{k*}_{n|n-1}H^\top+\overline{R}_n^k
%     .\\[-2em]\notag
% \end{align}
% where $\mu^{k*}_{n|n-1}=F_{n,k}\mu^{k*}_{n-1|n-1}$, $\Sigma^{k*}_{n|n-1}=F_{n,k}\Sigma^{k*}_{n-1|n-1}F_{n,k}^\top + Q_{n,k}$. 
%The variational distribution $q_n(X_{n,k})=\mathcal{N}(X_{n,k};\mu^{k}_{n|n},\Sigma^{k}_{n|n})$ can then be updated by Kalman filtering.
%, i.e.
%\begin{align}
%\begin{aligned} \label{eq:KF update}
 %   q_n(X_{n,k})&=\mathcal{N}(X_{n,k};\mu^{k}_{n|n},\Sigma^{k}_{n|n}),\\
 %   T^k_{n}&=\overbar{Y}_{n}^k-H\mu_{n|n-1}^{k*},\\
 %   S^k_{n}&=H\Sigma^{k*}_{n|n-1}H^\top+\overline{R}_n^k,\\
 %   G&=\Sigma^{k*}_{n|n-1}H^\top S^{k^{-1}}_{n},\\
 %   \mu^{k}_{n|n}&=\mu^{k*}_{n|n-1}+GT^k_{n},\\
 %   \Sigma^{k}_{n|n}& =(I-GH)\Sigma^{k*}_{n|n-1}.
%\end{aligned}
%\end{align}
%Such an update can be independently carried out for all targets.

\noindent \textbf{Update for $q_n(\theta_n)$:}
Next, we derive the update for $\theta_n$
\vspace{-0.5em}
\begin{align}\notag
     q_n(\theta_n) \propto&\prod_{s=1}^{N_s} \prod_{j=1}^{M_n^{s}}p(\theta_{n,j}^{s})\text{exp}\left(\E_{q_n(X_n)}\log \ell^s(Y_{n,j}^{s}|X_{n,\theta_{n,j}^{s}}) \right)\\[-0.5em] \label{eq:update for theta}   \propto&\prod_{s=1}^{N_s}\prod_{j=1}^{M_n}q_n(\theta_{n,j}^{s})\\[-2.5em]\notag
\end{align}
From it, we can directly obtain that $q_n(\theta_n)=\prod_{s=1}^{N_s}  q_n(\theta_n^{s})$, and $q_n(\theta_n^{s})=\prod_{j=1}^{M_n^{s}}q_n(\theta_{n,j}^{s})$, meaning that each sensor can update individually, and at each sensor, the update can also be performed in parallel as follows:  
\vspace{-0.7em}
\begin{align}  \label{eq:update theta}
   & q_n(\theta_{n,j}^{s})
    \propto\frac{\Lambda_0^{s}}{V^{s}}\delta[\theta_{n,j}^{s}=0]+\sum_{k=1}^K\Lambda_k^{s} l_k^{s}\delta[\theta_{n,j}^{s}=k],\\[-0.3em]  \notag
   & l_k^{s}=\mathcal{N}(Y_{n,j}^{s};H\mu_{n|n}^k,R_k^{s})\text{exp}(-0.5\text{Tr}({(R_k^{s})}^{-1}H\Sigma_{n|n}^k H^\top)).\\[-2em]\notag
\end{align}

\paragraph{Decentralised consensus-based CAVI} \label{sec: decentralised consensus}
In our prior work \cite{li2023consensus}, we decentralised the centralised CAVI method above using an average consensus algorithm \cite{olfati2004consensus}, which can in theory converge exactly to  to the centralised fusion result, see \cite{li2023consensus} for details. However, this requires a fully converged average consensus routine at each CAVI update iteration. 
Specifically, during each iteration of the CAVI update, each sensor $s$ independently updates $q_n(\theta_n^s)$ as per \eqref{eq:update theta}, while the update of $q(X_n)$ involves an additional iterative average consensus algorithm to communicate information for calculating \eqref{eq:pseudomeas R c} and \eqref{eq:pseudomeas Y c} with their neighbouring nodes across sensor network, which are essential for every sensor to accurately update $q_n(X_n)$ according to \eqref{eq: X update}. Hence, each sensor node has to wait for the consensus algorithm to converge before proceeding to the next iteration of the coordinate ascent update, which may potentially lead to a substantial communication cost. By contrast, the proposed methods here will not require consensus to be achieved at each iteration, see Section \ref{sec Decentralised Gradient Descent Variational Multi-object Tracker}.

\subsubsection{{Standard fixed-form ELBO and gradient-based variational inference for sensor fusion and tracking}}
%{This paper develops decentralised gradient-based variational inference methods that do not require waiting for consensus as discussed in \ref{sec: decentralised consensus}.}
Alternatively, we can apply the gradient-based variational inference, which requires defining a fixed-form ELBO and for our tracking task, $\mathcal{F}(\lambda_n,\rho_n)$ in \eqref{eq: ELBO function} is specified in the dynamic form as% in the dynamic form required for our tracking task:
\vspace{-0.3em}
\begin{align} \label{eq: fix-form ELBO tracking}
   \hspace{-0.5em} \mathcal{F}(\lambda_n,\rho_n)&=\E_{q(X_n;\lambda_n)q(\theta_n;\rho_n)}\log\frac{ f(X_n,\theta_n,Y_n)}{q(X_n;\lambda_n)q(\theta_n;\rho_n)}, \\[-1.8em]\notag
\end{align}
where $f(X_n,\theta_n,Y_n)$ is defined in \eqref{eq:specific unormal posterior}, and $\lambda_n,\rho_n$ are the governing variational parameters of variational distributions $q_n(X_{n})$ and $q_n(\theta_{n})$, whose specific forms are defined in section \ref{sec Gradient Descent distributed}.
%To approximate target distribution $\hat{p}_n(X_{n},\theta_{n}|{Y}_n)$, existing decentralised variational inference algorithm \cite{hua2015distributed} can be adopted to  maximise the standard ELBO in \eqref{eq: fix-form ELBO tracking}. However, as authors in \cite{hua2015distributed} pointed out, it does not provide theoretical analysis but just interpret it as a distributed implementation of the stochastic gradient descent variational inference. In the following sections, we will show that it require no stochastic gradient descent but can be rigorously derive by constructing a locally maximised ELBO, which proves in this paper to be an equivalent objective  to the standard  ELBO in \eqref{eq: fix-form ELBO tracking}, see detailed rationale in Section \ref{sec:lmelbo}. 
{It would then be possible to approximate the target distribution $\hat{p}_n(X_{n},\theta_{n}|{Y}_n)$, by an adaptation of the decentralised variational inference algorithm given in  \cite{hua2015distributed}.
%to maximise the standard ELBO in \eqref{eq: fix-form ELBO tracking} for this tracking task. 
We do not adopt this approach to derive our decentralised algorithm, owing to the lack of theoretical analysis and informal stochastic optimisation interpretation provided in \cite{hua2015distributed}.
%this approach lacks theoretical analysis and is only informally interpreted as a distributed implementation of stochastic gradient descent variational inference. 
Instead
%In this paper, we will demonstrate that the stochastic approximation procedure in \cite{hua2015distributed}
% for the same general variational inference task, e.g., in Section \ref{sec:ELBO problem setting}, 
%is not necessary. Instead,
we adopt a more rigorous formulation of the task by optimising a locally maximised ELBO, which we prove to be an equivalent objective to the standard ELBO in \eqref{eq: fix-form ELBO tracking}, as now detailed  in Section \ref{sec:lmelbo}.}

%quote from \cite{hua2015distributed}
%''Instead of providing theoretical analysis, we show that the procedure (21) can be interpreted as a distributed implementation of the stochastic variational inference [25]. Each node runs a gradient ascent step (21a) using only the local data, which is similar to the stochastic approximation based on the subsample [25]. The combination step (21b), which diffuses all local estimates over the entire network, can be considered as a procedure gradually collecting global (all local) sufficient statistics (since φθ,i is a function of expectations of related natural sufficient statistics as shown in Lemma 2) with the iterations of the VB procedure(17)''

%using the mean-field variational inference with $q_n(X_n,\theta_n)=q_n(X_n)q_n(\theta_n)$. In the considered tracking scenario,  $\lambda_n,\rho_n$ are defined as the governing variational parameters of variational distribution $q_n(X_{n})$, $q_n(\theta_{n})$ with a specific form defined in sections \ref{sec Gradient Descent distributed} and \ref{sec Natural Gradient Descent distributed}. Thus, the free-form ELBO  $\mathcal{F}(q(X_n),q(\theta_n))$ and fix-form ELBO $\mathcal{F}(\lambda_n,\rho_n)$ have the same form of \eqref{eq: ELBO general} and \eqref{eq: ELBO function}, with $f(X_n,\theta_n,Y_n)$ replaced by $\hat{p}_n(X_n,\theta_n,Y_n)$:

\vspace{-0.5em}

\vspace{-0.02em}
\section{Locally Maximised ELBO for General  Variational Inference Tasks}\label{sec:lmelbo}
%This section introduces a strategy of locally maximised ELBO (LM-ELBO) to enhance optimisation efficiency for general inference problems defined in Section \ref{sec:ELBO problem setting}, with the objective of optimising the fixed-form ELBO in \eqref{eq: ELBO function}. 
%Optimising the standard ELBO in \eqref{eq: ELBO function} can become inefficient with high-dimensional parameters $\lambda,\rho$. Hence, 
In order to achieve efficient decentralised inference, we first introduce a locally maximised ELBO (LM-ELBO) in the general setting. 
%which then will facilitate a very effective decentralised tracking algorithm based on gradient descent. 
%This section introduces an efficient strategy for construction of a locally maximised ELBO (LM-ELBO) in place of the conventional ELBO in \eqref{eq: ELBO function} to enhance optimisation efficiency.  The motivation is that optimising the standard ELBO in \eqref{eq: ELBO function} can be inefficient with high-dimensional parameters $\lambda,\rho$, where in contrast, the LM-ELBO with fewer parameters can lead to more efficient updates and potentially faster convergence. This LM-ELBO strategy could be applied to general inference problems as in Section \ref{sec:ELBO problem setting}. %, with the objective of optimising the fixed-form ELBO in \eqref{eq: ELBO function}. 
{Here, we unify existing analogous notions of LM-ELBO, introduce new properties, and offer simpler proofs for established properties of LM-ELBO.}
We will see later in Section \ref{sec Gradient Descent distributed} that the construction of LM-ELBO is particularly beneficial for our decentralised fusion and multi-object tracking applications, leading to more rapid convergence and a lower dimensional parameter search space. This section follows the notation in Section \ref{sec:ELBO problem setting} for the general setting of variational inference.

%Optimising the fixed-form ELBO in \eqref{eq: ELBO function} can become inefficient with high-dimensional parameters $\lambda,\rho$.This inefficiency motivates the exploration of alternative objectives with fewer parameters, enabling more efficient updates and potentially faster convergence. 

%Therefore, this section introduces the locally maximised ELBO (LM-ELBO), which will be used as an alternative objective to the conventional ELBO in \eqref{eq: ELBO function} in our task to enhance optimisation efficiency. We will see later that this is particularly beneficial in our decentralised fusion scheme in Section \ref{sec Gradient Descent distributed} and \ref{sec Natural Gradient Descent distributed}. 

%\textcolor{red}{SJG: Do we know that $q^*$ is unique or could there be a class of optimal distributions that achieve the global optimum of {\cal{F}}?
%Lily:In most case it is, but we remove it and rewrite into the following (above equation 12): For example, for variational distribution $q(\theta)$,  according to \cite{blei2017variational}, the global optimiser while fixing $q(X)$ is:}
\vspace{-1.em}
\subsection{Definition of LM-ELBO}\label{sec: LM-ELBO def and assump}
Here we present our definition of LM-ELBO and clarify its connection to other locally maximised ELBO objective functions. 
%\subsubsection{Definition and Assumption}
%\label{sec: LM-ELBO def and assump}
The idea is to eliminate $\rho$ from the joint ELBO of  \eqref{eq: ELBO function}.  The LM-ELBO $\mathcal{L}(\lambda)$ is then obtained %in terms of the fixed-form ELBO in \eqref{eq: ELBO function},
simply by replacing $q(\theta;\rho)$ in \eqref{eq: ELBO function} by the optimal form $q^*(\theta)$, as follows,
\vspace{-1.2em}
\begin{align} \label{eq: our LM-ELBO definition}
    \mathcal{L}(\lambda)\coloneqq&\E_{q(X;\lambda)q^*(\theta)}\log\frac{f(X,\theta,Y)}{q(X;\lambda)q^*(\theta)}, \\ \label{eq: our LM-ELBO CAVI local optimum}
    q^*(\theta)\propto& \exp\left(\E_{q(X;\lambda)} \log f(X,\theta,Y)\right), \\[-2em]\notag
\end{align}
noting that $q^*(\theta)$ is implicitly  a function of $\lambda$. In our tracking task, $q^*(\theta)$ is available in closed form.
%$q^*(\theta)$ is available in closed form for the NHPP observation model that we apply here in tracking. 
%The LM-ELBO $\mathcal{L}(\lambda)$ in \eqref{eq: our LM-ELBO definition} is a well-defined function even if $q^*(\theta)$ cannot be evaluated in a closed-form.
{The LM-ELBO $\mathcal{L}(\lambda)$ is then optimised with respect to the single parameter $\lambda$, thus reducing the parameter search space and (as shown later) enabling an efficient decentralised algorithm.}
The LM-ELBO $\mathcal{L}(\lambda)$ is thus used in place of the conventional fixed-form ELBO $\mathcal{F}(\lambda,\rho)$ in \eqref{eq: ELBO function}. 
It is assumed of course that the optimal distribution $q^*(\theta)$ in \eqref{eq: our LM-ELBO CAVI local optimum} is a member of the assumed distributional class $q(\theta;\rho)$.
%, which is easily verified in our scenarios.
%in $\mathcal{F}(\lambda,\rho)$ in \eqref{eq: ELBO function}, includes .
% where we also assume the distribution form of $q(\theta;\rho)$ encompasses the optimal distribution $q^*(\theta)$ in \eqref{eq: our LM-ELBO CAVI local optimum}.
We will denote by
$\rho^*(\lambda)$ the parameter value (or set of values) that reproduces $q^*(\theta)$ in \eqref{eq: our LM-ELBO CAVI local optimum} with $\lambda$ held fixed, i.e., $q^*(\theta)=q(\theta;\rho^*(\lambda))$.

%\vspace{-1.5em}
%\subsubsection{Existing Similar Definitions of LM-ELBO and connections} 
%Although the concept of LM-ELBO is not new, it is a less known strategy that was adopted across various variational inference studies under different names \cite{hoffman2013stochastic,durante2019conditionally,king2006fast,lazaro2012overlapping,hensman2012fast}. 
The concept of LM-ELBO has been adopted across various variational scenarios under different names \cite{hoffman2013stochastic,durante2019conditionally,king2006fast,lazaro2012overlapping,hensman2012fast}, although it has not found extensive usage compared with the standard ELBO approach.
%Nevertheless, it is a less known strategy, and here we unify existing analogous notions and present some new insights to justify its usage.
%Here, we unify existing analogous notions and present some new insights to justify its application. 
%The existing literature on variational inference has introduced two such objectives: 
Two versions in the literature include the  original LM-ELBO (abbreviated here as OLM-ELBO to distinguish from our LM-ELBO) in \cite{hoffman2013stochastic,hoffman2015structured,durante2019conditionally} and the KL-corrected (KLC) bound (also known as marginalised variational bound) \cite{king2006fast,lazaro2012overlapping,hensman2012fast,bonnevie2017difference}. 
These two approaches have been further developed (e.g. \cite{hoffman2015structured,durante2019conditionally} building on the OLM-ELBO and \cite{lazaro2012overlapping,hensman2012fast,bonnevie2017difference} on the KLC bound), although we are not aware of discussion in the literature on their connections. 
%Despite their conceptual similarity and independent influence on subsequent research (e.g., works such as \cite{hoffman2015structured,durante2019conditionally} build on the OLM-ELBO and \cite{lazaro2012overlapping,hensman2012fast,bonnevie2017difference} on the KLC bound), to our knowledge, a discussion on their connections is absent. 

Our investigations find that, compared to OLM-ELBO in \cite{hoffman2013stochastic}, the KLC bound in \cite{hensman2012fast} offers implementational advantages, % that justify its use. 
and hence our LM-ELBO closely adheres to the KLC bound in \cite{hensman2012fast}.
%The literature on variational inference has introduced two such objectives: the locally maximised ELBO \cite{hoffman2013stochastic,hoffman2015structured,durante2019conditionally} and the KL-corrected (KLC) bound (or marginalised variational bound) \cite{king2006fast,lazaro2012overlapping,hensman2012fast,bonnevie2017difference}. Despite their conceptual similarity and independent influence on subsequent research (e.g., works such as \cite{hoffman2015structured,durante2019conditionally} build on the former and \cite{lazaro2012overlapping,hensman2012fast,bonnevie2017difference} on the latter), to our knowledge, a discussion on their connections is absent. Our study finds that, compared to the original LM-ELBO (abbreviated as OLM-ELBO to distinguish from our defined LM-ELBO) in \cite{hoffman2013stochastic}, the KLC bound in \cite{hensman2012fast} is more convenient to implement and offers additional advantages that justify its use. Consequently, our definition of LM-ELBO is closely adhere the definition of the KLC bound in \cite{hensman2012fast}. 
Specifically, when $f$ in \eqref{eq: our LM-ELBO definition} is exactly the joint density $p(X,\theta,Y)$, then our LM-ELBO is equivalent through simple manipulation to the original KLC bound, Eq. (4) of \cite{hensman2012fast}. In addition, our LM-ELBO qualifies as an OLM-ELBO. This is because the properties of our LM-ELBO described in \eqref{eq: first property} and \eqref{eq: zero derivative property} meet the criteria of the OLM-ELBO in \cite{hoffman2013stochastic}.

%The following subsection will further show that our LM-ELBO, and consequently the KLC bound \cite{hensman2012fast}, qualifies as an OLM-ELBO (defined in \cite{hoffman2013stochastic}), given an appropriate form/family for $q(\theta;\rho)$ is assumed.

%The literature on variational inference has introduced two such objectives: the locally maximised ELBO \cite{hoffman2013stochastic,durante2019conditionally} and the KL-corrected (KLC) bound (or marginalised variational bound) \cite{king2006fast,lazaro2012overlapping,hensman2012fast}. Despite their conceptual similarity and independent influence on subsequent research, to our knowledge, a discussion on their connections is absent.  Our study finds that, compared to the original LM-ELBO in \cite{hoffman2013stochastic}, the KLC bound in \cite{hensman2012fast} is more convenient to implement and offers additional advantages that justify its use. Consequently, our definition of LM-ELBO is closely adhere the definition of the KLC bound in \cite{hensman2012fast}. The following subsection will further show that our LM-ELBO, and consequently the KLC bound \cite{hensman2012fast}, qualifies as an Original LM-ELBO (defined in \cite{hoffman2013stochastic}), given an appropriate form/family for $q(\theta;\rho)$ is assumed.

\vspace{-1em}
\subsection{Properties of LM-ELBO } \label{sec: LM-ELBO properties}
The LM-ELBO has a number of reassuring properties that ensure reasonable behaviour of the variational optimisation. First, from definitions in Section \ref{sec: LM-ELBO def and assump}, we have:
%is that the LM-ELBO corresponds to the full ELBO with $\rho$ fixed to its optimal value.

\noindent \textbf{Property 1:} 
%From definitions in Section \ref{sec: LM-ELBO def and assump}, we have%The above definitions and assumptions directly result in
\vspace{-1em}
\begin{align} \label{eq: first property}
    \mathcal{L}(\lambda)=&\mathcal{F}(\lambda,\rho=\rho^*(\lambda)),\\ \label{eq: equivalent definition}
    \mathcal{L}(\lambda)=&\max_\rho \mathcal{F}(\lambda,\rho), \\[-2.2em]\notag
\end{align}
where \eqref{eq: first property} is obtained by comparing the definitions in \eqref{eq: ELBO function} and \eqref{eq: our LM-ELBO definition}; \eqref{eq: equivalent definition} is derived using \eqref{eq: global optimal maximum} and $q(\theta;\rho^*(\lambda))=q^*(\theta)$, where $q^*(\theta)$ represents the global optimum that satisfies \eqref{eq: global optimal maximum}.
These properties play a key role in offering simpler and more intuitive derivations of existing properties in \cite{hensman2012fast,hoffman2013stochastic}, and in establishing Property 5 that justifies the use of LM-ELBO.

Secondly, properties related to derivatives of $\mathcal{L}(\lambda)$ and $\mathcal{F}(\lambda,\rho)$ are given in Properties 2-4, assuming sufficient regularity of the functions for the derivatives to exist. %Despite the properties 2.2 and 2.3 have been mentioned in \cite{hoffman2013stochastic}, we present a simpler way to prove them in this paper. 
%Note that it is assumed that the necessary derivatives exist for these properties to be valid.

\noindent \textbf{Property 2:} 
\vspace{-1em}
\begin{align} \label{eq: zero derivative property}
    \nabla_{\rho} \mathcal{F}(\lambda,\rho)|_{\rho=\rho^*(\lambda)}=0. \\[-2em]\notag
\end{align}
This directly follows from Property 1: since $\rho^*(\lambda)$ is the global maximiser  of $\mathcal{F}(\lambda,\rho)$ when $\lambda$ is held fixed, the gradient $\nabla_{\rho} \mathcal{F}(\lambda,\rho)|_{\rho=\rho^*(\lambda)}$, if it exists, must be zero. See also Appendix \ref{Alternative proof of Property 2.1} for an alternative derivation of Property 2.
%See also supplementary material I.A where an alternative derivation is presented.
%, and \cite{durante2019conditionally} where such a property is derived for the exponential family.% we also provide an alternative way to prove \eqref{eq: zero derivative property} by only using $\eqref{eq: ELBO function}$ and $q(\theta;\rho^*(\lambda))=q^*(\theta)$.

%Although existing work such as \cite{durante2019conditionally}, demonstrates similar conclusion that the optimal CAVI update coincides with updates derived from setting the ELBO's gradient to zero, their derivations is specific to an exponential family model.
%Although existing work such as \cite{durante2019conditionally}, demonstrates that the optimal CAVI update coincides with updates derived from setting the ELBO's gradient to zero, their derivations are specific to an exponential family model. In contrast, property 2.1 suggests that the optimal CAVI update automatically sets the ELBO's gradient to zero. This holds true under any system model assumption, provided that an appropriate form for $q(\theta;\rho)$ is chosen as assumed in Section \ref{sec: LM-ELBO def and assump}. 
% This property \eqref{eq: zero derivative property} will be used to show our LM-ELBO $\mathcal{L}(\lambda)$ is also an OLM-ELBO \cite{hoffman2013stochastic}, and to illustrate the next property. 

\noindent \textbf{Property 3:}
\vspace{-1em}
\begin{align}
\label{eq: derivative simplification}
    \nabla_{\lambda} \mathcal{L}(\lambda)=\nabla_{\lambda} \mathcal{F}(\lambda,\rho)|_{\rho=\rho^*(\lambda)} \\[-2em]\notag
\end{align}
This property simplifies the gradient computation: instead of directly calculating $\nabla_{\lambda} \mathcal{L}(\lambda)$ via \eqref{eq: our LM-ELBO definition}, which is a complex task since $q^*(\theta)$ is a function of $\lambda$, we compute the partial derivative of the fixed-form ELBO \eqref{eq: ELBO function}, treating $q(\theta;\rho)=q^*(\theta)$ as $\lambda$-independent during gradient evaluation.

\noindent \textbf{Property 4:}
This property highlights the curvature discrepancy between the two objectives: 
\vspace{-0.5em}
\begin{align} \label{eq: curvature inequality}
    {\nabla}_{\lambda}^2 \mathcal{L}(\lambda) = {\nabla}_{\lambda}^2 \mathcal{F}(\lambda,\rho)|_{\rho=\rho^*(\lambda)}+P,  \\[-2em]\notag
\end{align}
where $P$ is a positive semi-definite matrix, ${\nabla}_{\lambda}^2 \mathcal{L}(\lambda)$ is the Hessian of $\mathcal{L}(\lambda)$, and ${\nabla}_{\lambda}^2 \mathcal{F}(\lambda,\rho)$ is the Hessian of $\mathcal{F}(\lambda,\rho)$ considered as a function of $\lambda$ alone. 
% This property was proved in section 3.2 in \cite{hensman2012fast} through term expansion and mathematical manipulation. Here we revisited it from a new angle: it is a natural consequence of the properties in \eqref{eq: equivalent definition}, \eqref{eq: first property} and \eqref{eq: derivative simplification}. Using Taylor expansions on $\mathcal{L}(\lambda+\Delta\lambda)$ and $\mathcal{F}(\lambda+\Delta\lambda,\rho)$, and subtracting them yields
% \vspace{-0.5em}
% \begin{align} \notag
%     &\mathcal{L}(\lambda+\Delta\lambda)-\mathcal{F}(\lambda+\Delta\lambda,\rho)\\ \notag
%     =&\mathcal{L}(\lambda)-\mathcal{F}(\lambda,\rho)+(\nabla_{\lambda} \mathcal{L}(\lambda)- \nabla_{\lambda} \mathcal{F}(\lambda,\rho))^\top\Delta\lambda\\ \notag
%     &+0.5\Delta\lambda^\top (\bm{\nabla}_{\lambda}^2 \mathcal{L}(\lambda)-\bm{\nabla}_{\lambda}^2 \mathcal{F}(\lambda,\rho)) \Delta\lambda +o(||\Delta\lambda||^2).  \\[-2em]\notag
% \end{align}
% When $\rho=\rho^*(\lambda)$, the second line is $0$ owing to \eqref{eq: first property} and \eqref{eq: derivative simplification}. However, \eqref{eq: equivalent definition} implies the first line is always non-negative. Consequently, the $\bm{\nabla}_{\lambda}^2 \mathcal{L}(\lambda)-\bm{\nabla}_{\lambda}^2 \mathcal{F}(\lambda,\rho)$ in the third line must be positive semi-definite, thereby validating \eqref{eq: curvature inequality}. 
Property 4 was given by \cite{hensman2012fast} as the reason for the faster convergence observed by \cite{king2006fast} in optimising the KLC bound.

%\noindent \textbf{Simple proof to Property 2.2 and 2.3:}
Here, we present a simple way to  prove simultaneously Properties 3 and 4, showing they hold for any $\lambda=\lambda_0$, i.e.,
%Properties 3 and 4 can readily be shown to  hold for any  $\lambda=\lambda_0$, i.e.% Consider any point $\lambda=\lambda_0$, 
\vspace{-0.3em}
\begin{align} \label{eq: interme proof}
\begin{aligned}
    \nabla_{\lambda} \mathcal{L}(\lambda)|_{\lambda=\lambda_0}=&\nabla_{\lambda} \mathcal{F}(\lambda,\rho\!=\!\rho^*(\lambda_0))|_{\lambda=\lambda_0},\\
    \nabla_{\lambda}^2 \mathcal{L}(\lambda)|_{\lambda=\lambda_0} =&\nabla_{\lambda}^2 \mathcal{F}(\lambda,\rho\!=\!\rho^*(\lambda_0))|_{\lambda=\lambda_0}+P.
\end{aligned}
 \\[-1.8em]\notag
\end{align}
Define $g(\lambda)\coloneqq \mathcal{L}(\lambda)-\mathcal{F}(\lambda,\rho\!=\!\rho^*(\lambda_0))$. Then we have $g(\lambda)\geq0$ by \eqref{eq: equivalent definition}, and $g(\lambda_0)=0$ by \eqref{eq: first property}. Thus $\lambda_0$ is a global minimiser of $g(\lambda)$. By the first and second order necessary optimality conditions,  $\nabla_{\lambda}g(\lambda)|_{\lambda=\lambda_0}$, if it exists, must be zero; and $\nabla^2_{\lambda}g(\lambda)|_{\lambda=\lambda_0}$, if it exists, must be positive semi-definite. This directly leads to \eqref{eq: interme proof}, thus completing the proof.
Properties 3 and 4 were also proved in \cite{hensman2012fast} by expanding $\mathcal{L}$ and $\mathcal{F}$, which requires extra mathematical manipulation, and Property 3 was proved in \cite{hoffman2013stochastic}% in a relatively simple way
, however, requiring an additional assumption that $\nabla_\lambda\rho^*(\lambda)$ exists. 

Next, we introduce an optimality alignment property to validate our LM-ELBO as as a valid alternative objective:
\noindent \textbf{Property 5:} 
\textit{If $\lambda^*$ is a global maximiser, a local maximiser, or a stationary point of $\mathcal{L}(\lambda)$, then $[\lambda^*,\rho^*(\lambda^*)]$ is, respectively, a global maximiser, a local maximiser, or a stationary point of $\mathcal{F}(\lambda,\rho)$.}

This property shows that the LM-ELBO corresponds to the optimiser of the full ELBO: any optimum found by optimising $\mathcal{L}(\lambda)$ is inherently an optimum of the standard ELBO $\mathcal{F}(\lambda,\rho)$. It also ensures that our LM-ELBO does not introduce additional spurious optima. Full proof and analysis are given in Appendix \ref{Proof and analysis of Property 5}.
% \noindent \textbf{Property 3:} LM-ELBO provides a tighter bound on the log evidence than the standard ELBO: %(assuming $f$ is the joint density), i.e.,
% \begin{align} 
%     \mathcal{F}(\eta,\rho)\leq\mathcal{L}(\eta)\leq \log \int f(X,\theta,Y) dXd\theta,
% \end{align}
% Applying Jensen's inequality to \eqref{eq: ELBO general} yields:
% \begin{align} \label{eq: jensen}
%     \mathcal{F}(q(X),q(\theta))\leq \log \int f(X,\theta,Y) dXd\theta,
% \end{align}
% where the right hand side equals $\log p(Y)$ if $f$ represents $p(X,\theta,Y)$, highlighting the terme ELBO (log-Evidence Lower BOund). 

% This follows directly from \eqref{eq: equivalent definition} and \eqref{eq: jensen}

% Furthermore, the property \eqref{eq: zero derivative property}, combined with \eqref{eq: first property}, matches the definition of the OLM-ELBO in \cite{hoffman2013stochastic}, except that the 
% demonstrates that, under the same assumption, our LM-ELBO, and consequently the KLC bound \cite{hensman2012fast}, qualifies as an OLM-ELBO in \cite{hoffman2013stochastic}, whose definition is essentially a combination of \eqref{eq: first property/definition} and \eqref{eq: zero derivative property}.

%\subsubsection{Validating LM-ELBO with Optimality Properties} \label{sec: validating LMELBO}
%The goal of minimising KL divergence justifies the conventional ELBO $\mathcal{F}$ in \eqref{eq: ELBO function} due to its equivalence with this objective. 

Finally, a tighter bound property of LM-ELBO and discussions on convergence assurance for gradient hybrid CAVI are given in Appendix \ref{An additional property of LM-ELBO} and \ref{app:Convergence Assurance for Gradient Hybrid CAVI}.

%\textcolor{red}{I would move this next bit up to section V into the additional material...}

%\noindent \textbf{Property 3:} 
%Assume $f$ is the joint density, LM-ELBO provides a tighter bound on the log evidence than the standard ELBO
% \begin{align} 
%     \mathcal{F}(\lambda,\rho)\leq\mathcal{L}(\lambda)\leq \log \int f(X,\theta,Y) dXd\theta,  \\[-2em]\notag
% \end{align}
%One previously established property in \cite{hensman2012fast} for the KLC bound, suggests:
%indicating that the LM-ELBO provides a tighter bound on the log evidence than the conventional ELBO (assuming $f$ is the joint density). 
%This follows directly from Property 1 and Jensen's inequality to equation (12). When $f$ represents $p(X,\theta,Y)$, the right hand side equals $\log p(Y)$, highlighting the terme ELBO (log-Evidence Lower BOund). 

\vspace{-0.5em}
\section{Decentralised Gradient-based Variational inference Framework for Sensor Fusion }\label{sec Gradient Descent distributed}
%This paper chooses an alternative to the CAVI update, that is, the decentralised gradient-based algorithms to perform variational update step. However, we will see that directly applying decentralised gradient descent to optimise the original ELBO would require more computation and communication costs. Therefore, b
Based on the LM-ELBO strategy in Section \ref{sec:lmelbo}, we show here how to optimise the LM-ELBO  for the multi-sensor multi-object tracking task, presenting a flexible decentralised gradient-based variational inference framework that can readily accommodate several novel and established variants. 
% that can easily accommodate numerous novel established algorithms for this decentralised optimisation task. 
%requires computing gradients of local ELBOs with respect to both object states and association variables, and iteratively passing this gradient information to neighboring sensors, which could be both challenging to compute and inefficient in terms of communication costs.

%In this section, we show how to efficiently optimise the LM-ELBO using decentralised gradient-based variational inference methods, and present full solution of the considered decentralised multi-sensor multi-object tracking task. 

\vspace{-1em}
\subsection{The Rule of Decentralised Gradient Descent}\label{Decentralised Gradient Descent}
First, we present a brief introduction to the Decentralised Gradient Descent (DGD) strategy \cite{nedic2009distributed,zeng2018nonconvex,chang2020distributed}. It addresses the problem where $N_s$ sensors cooperatively maximise  $f(x) = \sum_{s=1}^{N_s} f_s(x)$, with $x \in \mathbb{R}^p$ and each $f_s$ known exclusively to sensor $s$. The DGD algorithm employs consensus ideas for estimating the gradient of the global objective function $\nabla f(x)$. Specifically, each sensor $s$ maintains a local estimate $x^s$ of the variable $x$, and updates it at iteration $i$ using
% the update rule for each sensor $s$ at iteration $i$ is:
\vspace{-0.2em}
\begin{equation}\label{eq:general dgd}
    x^{s}(i + 1) = \sum\nolimits_{j=1}^{N_s} w_{sj}x^{j}(i) + \alpha\nabla_{x^s} f_s(x^{s}(i)),\\[-0.2em]
\end{equation}
where $\alpha$ is the stepsize%, also known as the learning rate
. $w_{sj}$ is nonzero only if $s$ and $j$ are neighbours or $s = j$ and the matrix $W = [w_{sj}] \in \mathbb{R}^{N_s \times N_s}$ is symmetric and doubly stochastic \cite{zeng2018nonconvex}. A common choice is the Metropolis weight whose detailed form is given in \cite{xiao2005scheme}.
%[ Some comment here about how $w_{}$ are computed?]
Each sensor $s$ updates its local estimate $x^{s}$ by combining the average of its neighbours with a local gradient $\alpha \nabla f_s(x^{s})$. Note that the sign of the gradient is positive due to the maximisation task, though we retain the term gradient descent by convention. 

\subsubsection{The choice of the stepsize}
The convergence of the DGD algorithm is influenced by the stepsize  $\alpha$ in \eqref{eq:general dgd}. A stepsize that is too small results in slow convergence, while a large stepsize can prevent convergence or cause divergence. It is shown in \cite{chang2020distributed} that the DGD method guarantees convergence for both convex and non-convex functions. With diminishing step sizes specified in \cite{chang2020distributed}, after convergence, all sensors reach the same solution, which is a stationary point of $f(x)$.

%[Comment: `the' or `a' stationary point?thanks, 'a' would be more appropriate ]

\subsubsection{Gradient tracking strategy}
A gradient tracking strategy  \cite{nedic2017achieving} can be applied to speed up the convergence of the DGD algorithm. It relies on tracking differences of gradients: at each iteration $i$, each sensor $s$ maintains the gradient estimate $ \xi^{s}(i)$ along with the estimate $ x^{s}(i)$. In this setting, the update equations for the gradient tracking strategy at iteration $i$ for each sensor $s$ are modified as follows
\vspace{-.2em}
\begin{align}
   &  x^{s}(i + 1) = \sum\nolimits_{j=1}^{N_s} w_{sj}x^{j}(i) + \alpha \xi^{s}(i)\\[-0.02em]\notag
   &   \xi^{s}(i+1)\!= \!\!\sum\nolimits_{j=1}^{N_s} \!w_{sj}\xi^{j}(i)+\nabla_{x^s}\! f_s(x^{s}(i\!+\!1)) \! -\!\nabla_{x^s}\! f_s(x^{s}(i))\\[-1.8em]\notag
\end{align}
%Note that for time-varying sensor network, the weight $w_{sj}$ here could be time-varying depending on the connectivity of the sensor network at different time.
%With a constant stepsize, this gradient tracking can guarantee to converge to a stationary point  for both convex and nonconvex functions, and also for both time-invariant and time-varying graphs \cite{chang2020distributed}. Thus, it is particularly advantageous as it has a notably rapid convergence speed with convergence guarateen, and  constant stepsize simplifying the tuning process for practical applications. 

With a constant stepsize, this gradient tracking can guarantee convergence to a stationary point for both convex and nonconvex functions, as well as for both time-invariant and time-varying graphs \cite{chang2020distributed}. Thus, it is particularly advantageous due to its notably rapid convergence speed, guaranteed convergence, and the simplification of the tuning process in practical applications provided by the constant stepsize.

%Although these approaches can reach an exact solution at a constant stepsize, they necessitate the transfer of both the optimization variable and an extra auxiliary variable during each iteration. This requirement effectively increases the communication demands in each cycle, compared to traditional decentralized gradient-based algorithms.

%Here we adopt a gradient tracking strategy in \cite{nedic2017achieving} to speed up the convergence of DNGD algorithm since it is shown to have a notably rapid convergence speed. It is particularly advantageous as it ensures convergence with a constant step size \cite{chang2020distributed}, simplifying the tuning process for practical applications. 

\vspace{-0.5em}
\subsection{Justification for use of LM-ELBO over Original ELBO} \label{sec: LM-ELBO over O-ELBO}
% in Decentralised Variational Inference for Tracking and Sensor Fusion: Why Not Optimise the Original ELBO}
To maximise the original ELBO $\mathcal{F}(\lambda_n,\rho_n)$ in \eqref{eq: fix-form ELBO tracking}, we can directly apply the DGD rule in Section \ref{Decentralised Gradient Descent}.
% First, under the model assumptions in Section \ref{sec:PoissonModel}, the full expression of $\mathcal{F}(\lambda_n,\rho_n)$ in \eqref{eq: fix-form ELBO tracking} can be derived as follows using \eqref{eq: joint association prior conditionally independent}-\eqref{eq:obs prior} 
From now on, we assume $q_n(\theta_n;\rho_n)=\prod_{s=1}^{N_s}q_n(\theta_n^s;\rho_n^s)$, $q_n(X_n;\lambda_n)=\prod_{k=1}^{K}q_n(X_{n,k};\lambda_{n,k})$, where $\rho_n=[\rho_n^1,\rho_n^2,...,\rho_n^{N_s}]$, $\lambda_n=[\lambda_{n,1},\lambda_{n,2},...,\lambda_{n,K}]$.
% fully determines all $\rho_n^s$ for $s=1, 2, \dots, N_s$, and similarly for $\lambda_n$.
These expressions naturally result from the optimal CAVI updates, as shown in Section \ref{sec:ca update}.
The $\mathcal{F}(\lambda_n,\rho_n)$ in \eqref{eq: fix-form ELBO tracking} can then be written as follows using \eqref{eq: joint association prior conditionally independent}-\eqref{eq:obs prior}
\vspace{-1.2em}
\begin{align} \label{eq:full fix-form ELBO}
&\mathcal{F}(\lambda_n,\rho_n)= \sum_{s=1}^{N_s} \E_{q_n(X_n)q_n(\theta_n^{s};\rho_n^{s})}\log p(Y_{n}^{s}|\theta_{n}^{s},X_{n}) \\[-0.5em]\notag
&+ \sum_{s=1}^{N_s}  \E_{q_n(\theta_n^{s};\rho_n^{s})}\log  \frac{p(\theta_{n}^{s}|M_n^{s})}{q_n(\theta_{n}^{s};\rho_n^{s})} +\E_{q_n(X_n; \lambda_{n})}\log\frac{\hat{p}_n(X_n)}{q_n(X_n; \lambda_{n})} \\[-2.em]\notag
\end{align}
Subsequently, the global original ELBO in \eqref{eq:full fix-form ELBO} can be directly rewritten as the sum of local ELBO: $\mathcal{F}(\lambda_n,\rho_n)= \sum_{s=1}^{N_s} \mathcal{F}_s(\lambda_n,\rho_n^s)$ 
where each local ELBO $\mathcal{F}_s(\lambda_n,\rho_n^s)$ is
%\textcolor{red}{I can't find the full specification of the ELBO in the paper. Where is it? and is it then obvious how to split it as below?}
\vspace{-0.3em}
\begin{align}    \label{eq: OELBO each sensor} 
   & \mathcal{F}_s(\lambda_n,\rho_n^s)
     \coloneqq \E_{q_n(X_n; \lambda_{n})q_n(\theta_n^{s}; \rho_n^{s})}\log p(Y_{n}^{s}|\theta_{n}^{s},X_{n}) \\\notag
     &+   \E_{q_n(\theta_n^{s}; \rho_n^{s})}\log  \frac{p(\theta_{n}^{s}|M_n^{s})}{q_n(\theta_{n}^{s}; \rho_n^{s})} +\frac{1}{N_s}\E_{q_n(X_n; \lambda_{n})}\log\frac{\hat{p}_n(X_n)}{q_n(X_n; \lambda_{n})} \\[-1.9em]\notag
\end{align}
According to DGD rule in \eqref{eq:general dgd},  we need to calculate and transmit the gradient of the local ELBO $\nabla_{\lambda_n,\rho_n} \mathcal{F}_s(\lambda_n,\rho_n^s)$ with respect to both $\lambda_n$ and $\rho_n$ for this optimisation task. %This was the approach adopted in \cite{gan2022variational}. \textcolor{red}{SJG: am I right here?}

However, we can see that directly applying the DGD update to the original ELBO in the considered tracking tasks can be inefficient and costly, since sensors need to communicate extensive high-dimensional data association information through $\rho_n$. This motivates us to construct the LM-ELBO in Section \ref{sec:lmelbo} which will require fewer parameters. By optimising the LM-ELBO with the decentralised gradient-based methods, the computation of gradients is simplified and and we need only to exchange object state information $\lambda_n$, thus significantly reducing the communication overhead.

%However, for the considered tracking tasks, directly applying the DGD update \eqref{eq:general dgd} as such would force sensors to share extensive high-dimensional data association information since the optimisation variable includes parameters of $q_n(\theta_{n}^{s}; \rho_n)$, which would require a high communication burden for sensor fusion. This motivates us to develop the LM-ELBO in Section \ref{sec:ELBO problem setting}.  We will see that the LM-ELBO can be optimised in a decentralised manner, enabling the exchange of only object state information, thus greatly reducing communication overhead.

%Hence, This paper develops a decentralised gradient descent variational inference framework for this optimisation task, in which way sensors can run independently and no longer need to wait for the other sensors to converge. In particular, it not only has closed-form update rule, but also can be implemented in a decentralised fashion with theoretically guaranteed convergence towards centralised gradient descent objective.  Our proposed algorithm is enabled based on two main strategies, that is the construction of locally maximised ELBO and decentralised gradient descent optimisation methodology, which will be discussed as follows.

\vspace{-0.8em}
\subsection{Decentralisation of LM-ELBO for Multi-sensor Fusion}\label{sec:Decentralisation of LM-ELBO }
By the definition of LM-ELBO in \eqref{eq: our LM-ELBO definition}, for our tracking task, the LM-ELBO can be derived by replacing $q(\theta_{n};\rho_{n})$ in the ELBO in \eqref{eq: fix-form ELBO tracking} by the optimal form $q_n^*(\theta_{n})$ in \eqref{eq: our LM-ELBO CAVI local optimum}, i.e.,
%\textcolor{red}{THis needs some derivation, or at least equation references to all the required terms:}
\vspace{-0.5em}
\begin{align} \notag
   &\mathcal{L}(\lambda_{n})=\E_{q(X_n;\lambda_n)q_n^*(\theta_n)}\log\frac{ p(Y_n|\theta_n,X_n)p(\theta_n|M_n)\hat{p}_n(X_n)}{q(X_n;\lambda_n)q_n^*(\theta_n)}
   \\[-2em]\notag
\end{align}
where the $q_n^*(\theta_{n})$ follows the same derivation in \eqref{eq:update for theta}, i.e., 
% $q_n^*(\theta_n)=\prod_{s=1}^{N_s}  q_n^*(\theta_n^{s})$, $q_n^*(\theta_n^{s})=\prod_{j=1}^{M_n^{s}}q_n^*(\theta_{n,j}^{s})$, 
% \begin{align} \label{eq: local optimal theta}
% q_n^*(\theta_{n,j}^{s})\! \propto\! p(\theta_{n,j}^{s})\text{exp}(\E_{q_n(X_n;\lambda_n)}\!\log \ell^s(Y_{n,j}^{s}|X_{n,\theta_{n,j}^{s}}))
% \end{align}
\vspace{-0.5em}
\begin{align} \label{eq: local optimal theta}
    \begin{aligned}
        &q_n^*(\theta_n)=\prod\nolimits_{s=1}^{N_s}  q_n^*(\theta_n^{s}), \ \ \ \ q_n^*(\theta_n^{s})=\prod\nolimits_{j=1}^{M_n^{s}}q_n^*(\theta_{n,j}^{s}),\\[-0.2em]
        &q_n^*(\theta_{n,j}^{s})\! \propto\! p(\theta_{n,j}^{s})\text{exp}(\E_{q_n(X_n;\lambda_n)}\!\log \ell^s(Y_{n,j}^{s}|X_{n,\theta_{n,j}^{s}})).
    \end{aligned}\\[-2.em]\notag
\end{align}
Note that $q_n^*(\theta_{n,j}^{s})$ is also a function of $\lambda_{n}$. Subsequently,
\begin{align}\label{eq:locally maximised ELBO form}
   &\mathcal{L}(\lambda_{n})= \sum\nolimits_{s=1}^{N_s} \E_{q_n(X_n; \lambda_{n})q_n^*(\theta_n^{s})}\log p(Y_{n}^{s}|\theta_{n}^{s},X_{n})\\[-0.2em] \notag
   &+ \sum\nolimits_{s=1}^{N_s} \!\! \E_{q_n^*(\theta_n^{s})}\!\log  \frac{p(\theta_{n}^{s}|M_n^{s})}{q_n^*(\theta_{n}^{s})} + \E_{q_n\!(\!X_n; \lambda_{n}\!)}\!\log\!\frac{\hat{p}_n(X_n)}{q_n(X_n;\lambda_{n})} 
   \\[-1.8em]\notag
\end{align}
where \eqref{eq: joint association prior conditionally independent}-\eqref{eq:obs prior} are applied.
Next, we decompose $\mathcal{L}(\lambda_{n})$ in \eqref{eq:locally maximised ELBO form} into a sum of local LM-ELBOs $\mathcal{L}_s(\lambda_{n})$ at $s$-th sensor
%To formulate a consensus optimisation problem that many algorithms can be applied, we decentralise the LM-ELBO in \eqref{eq:locally maximised ELBO form} into a sum of the local LM-ELBOs $\mathcal{L}_s(\lambda_{n}^{s})$ at $s$-th sensor:
\vspace{-0.5em}
\begin{align}  \label{eq:local LMelbo sum}
&\mathcal{L}(\lambda_{n}) = \sum\nolimits_{s=1}^{N_s}\mathcal{L}_s(\lambda_{n})\\[-2.6em]\notag
\end{align}
\begin{align} \label{eq:local LM ELBO}
  & \mathcal{L}_s(\lambda_{n})
     \coloneqq \E_{q_n(X_n; \lambda_{n})q_n^*(\theta_n^{s})}\log p(Y_{n}^{s}|\theta_{n}^{s},X_{n})   \\\notag
    & + \E_{q_n^*(\theta_n^{s})}\log  \frac{p(\theta_{n}^{s}|M_n^{s})}{q_n^*(\theta_{n}^{s})}+ \frac{1}{N_s}\E_{q_n(X_n; \lambda_{n})}\log\frac{\hat{p}_n(X_n)}{q_n(X_n;\lambda_{n})}\\[-2em]\notag
\end{align}
%Under this design, it is now a decentralised optimisation problem, 
Thus, it is transformed into a decentralised optimisation problem, where each local $\mathcal{L}_s(\lambda_{n})$ depends only on local data $Y_n^s$, and computations with $\mathcal{L}_s(\lambda_{n})$ (e.g., gradients) can be performed fully locally. This design enables the usage of numerous established decentralised optimisation algorithms from the growing field to optimise the overall objective $\mathcal{L}(\lambda_{n})$. 

\subsubsection{Properties of local LM-ELBO $\mathcal{L}_s(\lambda_{n})$} \label{sec: local LM-ELBO properties}
% By our construction, the $\mathcal{L}(\lambda_n)$ automatically possesses the properties described in Section \ref{sec: LM-ELBO properties}, with Property 5 justifying it as an equally valid objective as $\mathcal{F}(\lambda_n,\rho_n)$.
% Derived from the LM-ELBO definition, the $\mathcal{L}(\lambda_n)$ now automatically possesses the properties described in Section \ref{sec: LM-ELBO properties}.
% A key advantage of our decentralised sensor fusion framework is that the local LM-ELBO $\mathcal{L}_s(\lambda_{n})$ in \eqref{eq:local LM ELBO} inherits many of the beneficial properties of $\mathcal{L}(\lambda_{n})$. 
While $\mathcal{L}(\lambda_n)$ naturally possesses the properties from Section \ref{sec: LM-ELBO properties} due to its derivation, it is not obvious that the decomposed local LM-ELBO $\mathcal{L}_s(\lambda_n)$ in \eqref{eq:local LM ELBO} would inherit them, but in our framework, it does. 
% However, in our decentralised sensor fusion framework, it does inherit many of these beneficial properties.
Specifically, denote by
$\rho_n^{s*}(\lambda_n)$ the parameter value that reproduces $q_n^*(\theta_n^s)$ in \eqref{eq: local optimal theta} with $\lambda_n$ held fixed -- i.e., $q_n^*(\theta_n^s)=q_n(\theta_n^s;\rho_n^{s*}(\lambda_n))$ -- then, by substituting $\mathcal{L}(\lambda)$ with $\mathcal{L}_s(\lambda_{n})$ in \eqref{eq:local LM ELBO} and $\mathcal{F}(\lambda,\rho)$ with $\mathcal{F}_s(\lambda_{n},\rho_n^s)$ in \eqref{eq: OELBO each sensor}, all properties 1-5 from Section \ref{sec: LM-ELBO properties} still hold.
A detailed list of these properties for $\mathcal{L}_s(\lambda_{n})$ and $\mathcal{F}_s(\lambda_{n},\rho_n^s)$, along with proofs, is provided in Appendix \ref{apx: properties for local LM-ELBO and proofs}.
% To prove these facts, it is suffices to prove property 1 holds true, as properties 2-5 are all proved in Section \ref{sec: LM-ELBO properties} using only property 1. The proof of property 1 for $\mathcal{L}_s(\lambda_{n})$ and $\mathcal{F}_s(\lambda_{n},\rho_n^s)$ is provided in Appendix \ref{apx: property 1 for local LM-ELBO}.
Among these properties, the most important, which greatly simplifies the computation of local gradients (as will be demonstrated in Section \ref{sec Decentralised Gradient Descent Variational Multi-object Tracker}), is
\vspace{-0.3em}
\begin{equation}\label{simplify the gradient computation}
\nabla_{\lambda_{n}} \mathcal{L}_s(\lambda_{n})=\nabla_{\lambda_{n}} \mathcal{F}_s(\lambda_n,\rho_n^s)|_{\rho_n^{s}=\rho_n^{s*}(\lambda_n)}.\\[-0.5em]
\end{equation}
% This will greatly simplifies the computation of the local gradients, as will be shown in Section \ref{sec Decentralised Gradient Descent Variational Multi-object Tracker}.
%thus it can perform local computation with $\mathcal{L}_s(\lambda_{n})$ (e.g., gradients). This formulation allows us to leverage many established decentralised optimisation algorithms from emerging decentralised optimisation field to collectively optimise $\mathcal{L}_s(\lambda_{n})$.

%Under this design it is now a decentralised optimisation problem, where each local $\mathcal{L}_s(\lambda_{n})$ depends only on local data $Y_n^s$, and thus can be locally computed by $s$-th sensor, and share information with neighbours,allowing for many established decentralised optimisation algorithms.

%%% \subsection{Decentralised Gradient Descent for Maximising $\mathcal{L}(\lambda_{n})$}\label{sec:Decentralised Gradient Descent}
% Based on the decentralisation of LM-ELBO in Section \ref{sec:Decentralisation of LM-ELBO }, we can directly adopt the classical DGD method in Section \ref{Decentralised Gradient Descent} to optimise $\mathcal{L}(\lambda_{n})$ such that each sensor $s$ can update its variational parameters $\lambda_{n}^{s}$ using \eqref{eq:general dgd} with local gradients $\nabla_{\lambda_{n}^{s}} \mathcal{L}_s(\lambda_{n}^{s})$. This DGD has theoretically guaranteed convergence as discussed in Section \ref{Decentralised Gradient Descent}. Detailed analysis, simplified gradient calculation as in \eqref{eq: derivative simplification}, and closed-form update rules will be presented in our journal paper.

\vspace{-0.8em}
\subsection{Decentralised (Natural) Gradient Descent Variational inference for Maximising LM-ELBO}\label{sec:Decentralised (Natural) Gradient Descent Variational inference for Maximising LM-ELBO}
%Here we present two ways of decentralised gradient descent variational inference methods for maximising $\mathcal{L}(\lambda_{n})$, which have theoretical convergence guarantee for both convex and non-convex objective functions. We also extend its scheme to accommodate the employment of a more efficient natural gradient instead of the standard gradient that are more extensive studied in existing literature, e.g., \cite{xiao2005scheme}.
We present two decentralised gradient descent variational inference methods for maximising $\mathcal{L}(\lambda_{n})$, which are theoretically guaranteed to converge for both convex and non-convex objective functions. %Additionally, we enhance this framework by integrating a more efficient natural gradient to improve convergence speed, as opposed to using the standard gradient that is commonly explored in existing literature, such as in \cite{xiao2005scheme}.
Further, we improve its convergence speed by integrating a more efficient natural gradient into the DGD scheme. Although theoretical studies on decentralised natural gradients are few, we demonstrate their promising performance in multi-object tracking tasks in Section \ref{sec:Results}.
\subsubsection{Decentralised (natural) gradient descent variational inference with diminishing stepsize}
According to DGD rule in Section  \ref{Decentralised Gradient Descent}, the update equation at each iteration $i$ at each sensor $s$ for jointly optimising the LM-ELBO $\mathcal{L}(\lambda_{n})$ is
\vspace{-0.5em}
\begin{align}\label{eq:update equation 1}
   & \lambda_{n}^{s}(i+1)=\sum\nolimits_{j=1}^{N_s} w_{sj}(i) \lambda_{n}^{j}(i) + \alpha_i \boldsymbol{g}_i(\mathcal{L}_s) \\[-2em]\notag
\end{align} 
where $\lambda_{n}^{s}$ is the sensor $s$'s local estimate of $\lambda_{n}$. $\boldsymbol{g}_i(\mathcal{L}_s)$ can represent either the normal gradient ${\nabla}_{\lambda_{n}^{s}} \mathcal{L}_s(\lambda_{n}^{s}(i))$ or the natural gradient $\hat{\nabla}_{\lambda_{n}^{s}} \mathcal{L}_s(\lambda_{n}^{s}(i))$, as detailed in Section \ref{sec Decentralised Gradient Descent Variational Multi-object Tracker}. The weight $w_{sj}(i)$ is chosen as the Metropolis weight in \cite{xiao2005scheme}:
\begin{equation} \label{eq: metro weights}
 w_{sj}(i) =
\begin{cases}
\frac{1}{1 + \max{\{d_s{(i)}, d_j{(i)}\}}} & \text{if } j \in \mathcal{N}_s{(i)}, \\
1 - \sum_{{s,k} \in \mathcal{E}(i)} w_{sk}{(i)} & \text{if } j = s 
\end{cases}
\end{equation}
%\textcolor{red}{This weight needs to be detailed, with some brief description}
Note that $w_{sj}(i)$ depends on the connectivity of the sensor network $\mathcal{G}(i)$, which may be time-varying. %In particular, we apply a diminishing stepsize $\alpha_i$, under which the DGD algorithms have guaranteed convergence for this non-convex objective under several specific form, e.g., $\frac{1}{i}$, and it will be more with development of decentralised optimisation theory. 
In particular, we employ a diminishing stepsize, $\alpha_i$, to ensure guaranteed convergence of the DGD algorithms to this non-convex objective under forms, for example   $\alpha_i=1/i$ as proposed in \cite{chang2020distributed}.% Other improved choices of $\alpha_i$ may well emerge as the theoretical field develops.

\subsubsection{Decentralised (natural) gradient descent variational inference with gradient tracking}
To further improve convergence speed, we can as an alternative maximise $\mathcal{L}(\lambda_{n})$ using gradient tracking methods that track the differences of gradients.
To employ it in our setting, for each sensor $s$ and each iteration $i$, we update both the local estimate $\lambda_{n}^{s}(i)$ of the variational parameter and an additional gradient estimate $ \boldsymbol{\xi}_{n}^{s}(i)$, leading to the following update equations,
\vspace{-0.5em}
\begin{align}\label{eq:update equations for each variational parameter}
   &\hspace{-0.5em} \lambda_{n}^{s}(i+1) = \sum\nolimits_{j=1}^{N_s} w_{sj}(i) \lambda_{n}^{s}(i) + \alpha \boldsymbol{\xi}_{n}^{s}(i),\\[-0.1em]\label{eq:update equation 2}
   &\hspace{-0.5em} \boldsymbol{\xi}_{n}^{s}(i+1) = \!\sum\nolimits_{j=1}^{N_s} \!\!w_{sj}(i) \boldsymbol{\xi}_{n}^{s}(i)+ \boldsymbol{g}_{i+1}(\mathcal{L}_s)-\boldsymbol{g}_{i}(\mathcal{L}_s). \\[-2em]\notag
\end{align}
With a fixed stepsize $\alpha$, the gradient tracking approach for decentralised inference is theoretically guaranteed to converge to a stationary point \cite{chang2020distributed}. To our knowledge this is the first development of such a decentralised (natural) gradient tracking scheme within a tracking application, and Section \ref{sec:Results} demonstrates its empirical convergence and excellent performance, while significantly reducing the communication costs compared to the previous consensus-based approach \cite{li2023consensus}.%\textcolor{red}{Check the citation...}

\vspace{-0.9em}
\section{Decentralised Gradient-based Variational Multi-object Trackers}\label{sec Decentralised Gradient Descent Variational Multi-object Tracker}
{This section provides detailed derivations and implementation steps of the proposed distributed multi-object trackers based on the variational filtering in Section \ref{sec:Sequential Bayesian inference for multi-object tracking} and decentralised (natural) gradient descent variational inference in Section \ref{sec:Decentralised (Natural) Gradient Descent Variational inference for Maximising LM-ELBO}.} Here, we assume an independent Gaussian prior at the initial time step,
% $p(X_0)\!=\!\prod_{k=1}^K p(X_{0,k})$ with $p(X_{0,k})\!=\!\mathcal{N}(X_{0,k};\mu^{k}_{0|0},\Sigma^{k}_{0|0})$.
$p(X_0)\!=\!\prod_{k=1}^K \mathcal{N}(X_{0,k};\mu^{k}_{0|0},\Sigma^{k}_{0|0})$.
We also assume an independent Gaussian variational distribution, which for sensor $s$ with local estimate $\lambda_n^s\!=\![\lambda_{n,1}^s,\!...,\lambda_{n,K}^s]$, is $q_n(X_n;\!\lambda_{n}^{s})\!=\!\prod_{k=1}^K \!q_n(X_{n,k};\!\lambda_{n,k}^s)$, where
$q_n(X_{n,k};\lambda_{n,k}^s)\!=\!\mathcal{N}(X_{n,k};\mu^{k,s}_{n|n},\Sigma^{k,s}_{n|n})$.
Using \eqref{eq: dynamic transition}, this then leads to independent predictive prior $\hat{p}(X_n)=\prod_{k=1}^K\hat{p}(X_{n,k})$.
Finally we denote $q_n^{s,*}(\theta_n^{s})$ as the optimal $q_n^*(\theta_n^{s})$ in \eqref{eq: local optimal theta}, computed using local estimate $\lambda_n^s$. Specifically, $q_n^{s,*}(\theta_{n,j}^{s})$ has the form of \eqref{eq:update theta} with $\mu_{n|n}^k$ and $\Sigma_{n|n}^k$ replaced by $\mu_{n|n}^{k,s}$ and $\Sigma_{n|n}^{k,s}$.
Then, we have $q_n^{s,*}(\theta_n^{s})\!=\!q_n(\theta_n^s;\rho_n^{s*}(\lambda_n^s))$ with $\rho_n^{s*}$ defined in Section \ref{sec: local LM-ELBO properties}.

% i.e., $q_n^{s,*}(\theta_n^{s})=q_n(\theta_n^s;\rho_n^{s*}(\lambda_n^s))$.

\vspace{-1em}
\subsection{Decentralised Gradient  Variational Multi-object Trackers} \label{sec: decentralised gradient trackers}
{Two decentralised trackers, the Decentralised Gradient Variational multi-object Trackers with Diminishing Stepsize (DeG-VT-DS) and Decentralised Gradient Variational multi-object Tracker with Gradient Tracking (DeG-VT-GT), are developed using standard gradient and DGD rule in Section \ref{sec:Decentralised (Natural) Gradient Descent Variational inference for Maximising LM-ELBO}.} In this case, the local estimate $\lambda_{n,k}^s$ for sensor $s$ is defined as
\vspace{-0.5em}
\begin{equation} \label{eq: lambda define normal gradient}
    \lambda_{n,k}^s=[\mu^{k,s}_{n|n},\Sigma^{k,s}_{n|n}], \ \ \ k=1,...,K\\[-0.5em]
\end{equation}
\subsubsection{Prediction and update steps}\label{sec:Prediction and update steps}
% We assume an independent Gaussian prior $p(X_0)=\prod_{k=1}^K p(X_{0,k})$ at initial time step, with $p(X_{0,k})=\mathcal{N}(X_{0,k};\mu^{k}_{0|0},\Sigma^{k}_{0|0})$, and assume independent Gaussian form for variational distribution
% % With the prediction and update step in \eqref{eq: predictive prior}-\eqref{eq:phatjoint}, the updated variational distribution will retain its independent Gaussian form, 
% $q_n(X_n;\lambda_{n}^s)=\prod_{k=1}^K q_n(X_{n,k};\lambda_{n,k}^s)$, where
% $q_n(X_{n,k};\lambda_{n,k}^s)=\mathcal{N}(X_{n,k};\mu^{k,s}_{n|n},\Sigma^{k,s}_{n|n})$, with its variational parameters $\lambda_{n,k}^s$
% \vspace{-0.5em}
% \begin{equation}
%     \lambda_{n,k}^s=[\mu^{k,s}_{n|n},\Sigma^{k,s}_{n|n}]^{\top}, k=1,...,K\\[-0.5em]
% \end{equation}
At time step $n-1$, the converged variational distribution is $q^*_{n-1}(X_{n-1,k};\lambda_{n-1,k}^s)= \mathcal{N}(X_{n-1,k};\mu^{k*,s}_{n-1|n-1},\Sigma^{k*,s}_{n-1|n-1})$. Then, in the prediction step at time step $n$, this local estimate $\lambda_{n-1,k}^s$ is used to compute the predictive prior $\hat{p}_n(X_{n,k})=\mathcal{N}(X_{n,k};\mu^{k*,s}_{n|n-1},\Sigma^{k*,s}_{n|n-1})$ for object $k$, with $\mu^{k*,s}_{n|n-1},\Sigma^{k*,s}_{n|n-1}$ computed according to \eqref{eq:predictive prior computation}. 

Note that if consensus is reached at time step $n-1$, all sensors have the same converged variational distribution $q^*_{n}(X_{n-1,k};\lambda_{n-1,k}^s)$ with all $\{\lambda_{n-1,k}^s\}_{s=1}^{N_s}$ being equal; thus, all sensors have the same predictive prior $\hat{p}_n(X_{n,k})$ as assumed in \eqref{eq:local LM ELBO}. In Section \ref{sec:Robust decentralised tracking}, we also examine cases where sensors have not converged to the same variational distribution due to insufficient iterations.

% In the update step, local sensors execute CAVI update for $q_n^*(\theta_n)$ where for each sensor $s$, $q_n^{s,*}(\theta_{n,j}^{s})$ has the form of \eqref{eq:update theta} with $\mu_{n|n}^k$ and $\Sigma_{n|n}^k$ replaced by $\mu_{n|n}^{k,s}$ and $\Sigma_{n|n}^{k,s}$.

% {Then, for the proposed DeG-VT-DS algorithm, each local sensor executes update equation \eqref{eq:update equation 1}  for variational parameters $\{\lambda_{n,k}^s\}_{s=1}^{N_s}$; for the proposed DeG-VT-GT algorithm, each local sensor executes update equations  \eqref{eq:update equations for each variational parameter}-\eqref{eq:update equation 2} for variational parameters $\{\lambda_{n,k}^s\}_{s=1}^{N_s}$.} 
In the update step, for the proposed DeG-VT-DS algorithm, each local sensor executes iterative update in \eqref{eq:update equation 1} for local estimate $\{\lambda_{n,k}^s\}_{s=1}^{N_s}$; for the proposed DeG-VT-GT algorithm, each local sensor executes iterative update in  \eqref{eq:update equations for each variational parameter}-\eqref{eq:update equation 2} for local estimate $\{\lambda_{n,k}^s\}_{s=1}^{N_s}$. In both algorithms, every update requires pre-computation of $q_n^{s,*}(\theta_{n,j}^{s})$ with the latest $\lambda_{n}^s$ to simplify the computation of gradients $\boldsymbol{g}_i(\mathcal{L}_s)$, which will be used in \eqref{eq:update equation 1}-\eqref{eq:update equation 2}. 
In the next subsection, we present the derivation of $\boldsymbol{g}_i(\mathcal{L}_s)$ at iteration $i$ at each sensor $s$.

Finally, full implementations of the DeG-VT-DS and DeG-VT-GT are given in Algorithm \ref{Algo:DeG-VT-DS} and \ref{Algo:DeG-VT-GT} in Appendix \ref{apx: tracker and pseudocodes}.
%\vspace{-0.8em}
\subsubsection{Derivation of $\boldsymbol{g}_i(\mathcal{L}_s)$} \label{sec: standard gradient derivation}
For DeG-VT-DS and DeG-VT-GT algorithms, $\boldsymbol{g}_i(\mathcal{L}_s)\!=\!{\nabla}_{\lambda_{n}^{s}} \mathcal{L}_s(\lambda_{n}^{s})|_{\lambda_{n}^{s}=\lambda_{n}^{s}(i)}$, i.e., the normal gradient with respect to $\lambda_{n}^{s}$.
% Using \eqref{simplify the gradient computation}, this simplifies to: $\nabla_{\lambda_{n}^{s}} \mathcal{L}_s(\lambda_{n}^{s})\!=\!\nabla_{\lambda_{n}^{s}} \mathcal{F}_s(\lambda_n^{s},\rho_n)|_{\rho_n=\rho_n^*(\lambda_n^s)}$. Thus, it is equivalent to compute $\nabla_{\lambda_{n}^{s}} \mathcal{F}_s(\lambda_n^{s},\rho_n^*(\lambda_n^s))$ ($\mathcal{F}_s$ defined in \eqref{eq: OELBO each sensor}) while treating $q_n(\theta_n^s;\rho_n^{s*}(\lambda_n^s))=q_n^{s,*}(\theta_n^{s})$ independent of $\lambda_n^s$ while evaluating the gradient. Its form is (see Appendix \ref{apx: gradient derivation} for detailed derivations):
Using \eqref{simplify the gradient computation}, this simplifies to $\nabla_{\lambda_{n}^{s}} \mathcal{L}_s(\lambda_{n}^{s})\!=\!\nabla_{\lambda_{n}^{s}} \mathcal{F}_s(\lambda_n^{s},\rho_n^{s*}(\lambda_n^s))$, where $\mathcal{F}_s$ is defined in \eqref{eq: OELBO each sensor}, and $\rho_n^{s*}(\lambda_n^s)$ is considered constant and independent of $\lambda_n^s$ when evaluating the gradient. Its final form is given below (see Appendix \ref{apx: gradient derivation} for detailed derivations):
% Thus, we can first evaluate the $\mathcal{F}_s(\lambda_n^s,\rho_n^s)$ in \eqref{eq: OELBO each sensor} with $q_n(\theta_{n}^{s}; \rho_n^{s})$ replaced by $q_n^{s,*}(\theta_{n,j}^{s})$, then evaluating its gradient while treating $q_n^{s,*}(\theta_{n,j}^{s})$ $\lambda_n^s$-independent.
% Its computation can be simplified using the  property (See proof in Appendix II):
% \vspace{-0.3em}
% \begin{equation}\label{simplify the gradient computation}
% \nabla_{\lambda_{n}^{s}} \mathcal{L}_s(\lambda_{n}^{s})=\nabla_{\lambda_{n}^{s}} \mathcal{F}_s(\lambda_n^{s},\rho_n)|_{\rho_n=\rho_n^*(\lambda_n)}\\[-0.5em]
% \end{equation}
% Thus, we can first derive each component's expectation of $\mathcal{F}_s(\lambda_n,\rho_n)$ in \eqref{eq: OELBO each sensor}; then, using \eqref{simplify the gradient computation}, the final form of the gradient is as follows (see Appendix \ref{apx: gradient derivation} for detailed derivations): 
%To compute the gradient, first, we derive the expectations of   components of the local LM-ELBO in \eqref{eq:local LM ELBO}, and the final form of LM-ELBO is written as
\vspace{-0.5em}
\begin{align} \notag
  & \nabla_{\lambda_{n}^{s}} \mathcal{F}_s(\lambda_n^{s},\rho_n^{s*}(\lambda_n^s))
     \!=\! \frac{-1}{2 N_s}\sum\nolimits_{k=1}^{K}\!\!\!\nabla_{\lambda_{n}^{s}}\!\left [\mathrm{Tr}\Big(\!\big(\Sigma^{k*,s}_{n|n-1}\big)^{\!-1}\Sigma^{k,s}_{n|n}\!\Big) \right.\\[-0.3em]\notag 
   & \left.- \log\big|\Sigma^{k,s}_{n|n}\big| + (\mu^{k,s}_{n|n} \!-\! \mu^{k*,s}_{n|n-1})^{\!\top} (\Sigma^{k*,s}_{n|n-1})^{-1} (\mu^{k,s}_{n|n} \!-\! \mu^{k*,s}_{n|n-1}) \right] \\[-0.5em] \notag
   & -\frac{1}{2} \sum\nolimits_{k=1}^K \!\!\nabla_{\lambda_{n}^{s}}\left [ \big( H\mu^{k,s}_{n|n}\!-\!\overbar{Y}_n^{k,s}\big)^\top \big(\overbar{R}_n^{k,s}\big)^{-1} \big(H\mu^{k,s}_{n|n}\!-\!\overbar{Y}_n^{k,s}\big)  \right.\\[-0.3em]\label{eq:local LM ELBO form}
   &\left.\hspace{6em}+ \mathrm{Tr}\big(H^\top(\overbar{R}_n^{k,s})^{-1}H\Sigma^{k,s}_{n|n}\big)\right] \\[-2em]\notag
\end{align}
where the following local parameters $\overbar{Y}_n^{k,s}$ and $\overbar{R}_n^{k,s}$ are treated as independent of $\lambda_n^s$ during gradient evaluation at sensor $s$
\vspace{-0.3em}
\begin{align}
\label{eq:pseudomeas R}
    \overbar{R}_n^{k,s}=&\frac{R_k}{\sum_{j=1}^{M_n^{s}}q_n^{s,*}(\theta_{n,j}^{s}=k)},\\[-2.6em]\notag
\end{align}
\begin{align}
\label{eq:pseudomeas Y}
\overbar{Y}_n^{k,s}=&\frac{\sum_{j=1}^{M_n^{s}}Y_{n,j}^{s}q_n^{s,*}(\theta_{n,j}^{s}=k)}{\sum_{j=1}^{M_n^{s}}q_n^{s,*}(\theta_{n,j}^{s}=k)}.\\[-2em]\notag
\end{align}
% Thus, each component of the gradients ${\nabla}_{\!\lambda_{n}^{s}} \mathcal{L}_s(\lambda_{n}^{s})=\big\{\nabla_{\!\mu^{k,s}_{n|n}}  \mathcal{L}_s(\lambda_{n}^{s}), \ \nabla_{\!\Sigma^{k,s}_{n|n}}   \mathcal{L}_s(\lambda_{n}^{s})\big\}^{K}_{k=1}$ can be computed with respect to each variational parameter $\mu^{k,s}_{n|n}$ and $\Sigma^{k,s}_{n|n}$:
Then, ${\nabla}_{\!\lambda_{n}^{s}} \mathcal{L}_s(\lambda_{n}^{s})$ is evaluated (detailed in Appendix \ref{apx: gradient derivation}) through its components $\nabla_{\!\mu^{k,s}_{n|n}}  \mathcal{L}_s(\lambda_{n}^{s})$ and $\nabla_{\!\Sigma^{k,s}_{n|n}}   \mathcal{L}_s(\lambda_{n}^{s})$ with respect to each local estimate $\mu^{k,s}_{n|n}$ and $\Sigma^{k,s}_{n|n}$ for $k\!=\!1,2,...,K$:

% ${\nabla}_{\!\lambda_{n}^{s}} \mathcal{L}_s(\lambda_{n}^{s})=\big\{\nabla_{\!\mu^{k,s}_{n|n}}  \mathcal{L}_s(\lambda_{n}^{s}), \ \nabla_{\!\Sigma^{k,s}_{n|n}}   \mathcal{L}_s(\lambda_{n}^{s})\big\}^{K}_{k=1}$ can be computed with respect to each variational parameter $\mu^{k,s}_{n|n}$ and $\Sigma^{k,s}_{n|n}$:
% \begin{align} \notag
% &{\nabla}_{\lambda_{n}^{s}} \mathcal{L}_s(\lambda_{n}^{s}(i))=\begin{bmatrix}
%      \nabla_{\mu^{1,s}_{n|n}}  \mathcal{L}_s(\lambda_{n}^{s}(i)),...,\nabla_{\mu^{K,s}_{n|n}}  \mathcal{L}_s(\lambda_{n}^{s}(i)) \\\nabla_{\Sigma^{1,s}_{n|n}}  \mathcal{L}_s(\lambda_{n}^{s}(i)),...,\nabla_{\Sigma^{K,s}_{n|n}}  \mathcal{L}_s(\lambda_{n}^{s}(i))
% \end{bmatrix}
% \end{align}
\vspace{-0.9em}
\begin{align} \notag
 \nabla_{\mu^{k,s}_{n|n}} \mathcal{L}_s(\lambda_{n}^{s})&=\frac{-1}{N_s} (\Sigma^{k*,s}_{n|n-1})^{-1}(\mu^{k*,s}_{n|n-1}-\mu^{k,s}_{n|n})\\
& \quad +H^\top (\overbar{R}_n^{k,s})^{-1} (\overbar{Y}_n^{k,s}-H \mu^{k,s}_{n|n}) 
\\[-2.2em]\notag
\end{align}
\begin{align} \notag
\nabla_{\Sigma^{k,s}_{n|n}} \mathcal{L}_s(\lambda_{n}^{s})
&=\frac{1}{2N_s}\left( (\Sigma^{k,s}_{n|n})^{-1}- (\Sigma^{k*,s}_{n|n-1})^{-1} \right) \\
& \quad- \frac{1}{2}H^\top(\overbar{R}_n^{k,s})^{-1} H\\[-2.em]\notag
\end{align}

\vspace{-0.1em}
\subsection{Decentralised Natural Gradient Variational Trackers}\label{sec Natural Gradient Descent distributed}
%In the previous section, we explored the decentralised gradient descent variational inference for tracking multiple objects amidst clutter. This approach, while innovative, presents certain inefficiencies, particularly in convergence speed. One possible issue is that the variational parameters are updated by taking small steps in a Euclidean parameter space, whereas for updating parameters of distributions, it ignores the information geometry of the posterior approximation and could lead to slow convergence rate. 
One possible issue with the proposed DeG-VT-DS and DeG-VT-GT in the previous section is that they use the standard gradient descent; as a result, the variational parameters are updated by taking small steps in a Euclidean parameter space, whereas for updating parameters of distributions, it ignores the information geometry of the posterior approximation and could lead to slow convergence rate. 
Natural gradient scales the traditional gradient with the inverse of its Fisher Information Matrix (FIM), $G(\lambda_{n})$, i.e., 
\begin{align}
   \label{eq:natural gradients the Fisher information matrix}
   &\hat{\nabla}_{\lambda_{n}^{s}} \mathcal{L}_s(\lambda_{n}^{s})=G(\lambda_{n}^{s})^{-1}\nabla_{\lambda_{n}^{s}} \mathcal{L}_s(\lambda_{n}^{s}).
\end{align} 
where $\hat{\nabla}$ denotes the natural gradient, and FIM $G(\lambda_{n})$ is a Riemannian metric for computing distance in the distribution:
\begin{align}\notag
   &G(\lambda_{n}^{s})= \E_{q_{n}}\left[ \left(\nabla_{\lambda_{n}^{s}}\ln{q_n(X_{n}; \lambda_{n}^{s})}\right)\left(\nabla_{\lambda_{n}^{s}}\ln{q_n(X_{n};\lambda_{n}^{s})}\right)^\top \right] 
\end{align}
For easier computation of the natural gradient, the optimised distribution parameter is typically defined as the natural parameter of the exponential family, as we will later define for $\lambda_n^s$ in \eqref{eq: natural parameters definition}. The use of the natural gradient is well known to enhance convergence over standard gradients \cite{amari1998natural,hoffman2013stochastic}.
%FIM exploits the Riemannian geometry to adjust the direction of the traditional gradient to the steepest ascent that is more aligned with the underlying statistical manifold, which has been demonstrated to enhance convergence over traditional gradients \cite{amari1998natural,hoffman2013stochastic}.

Hence, we propose a decentralised natural gradient descent scheme where local sensors collaboratively solve the same optimisation task, but replacing the standard gradient with the natural gradient in the update equations in \eqref{eq:update equation 1}-\eqref{eq:update equation 2}, with $\boldsymbol{g}_i(\mathcal{L}_s)=\hat{\nabla}_{\lambda_{n}^{s}} \mathcal{L}_s(\lambda_{n}^{s})|_{\lambda_{n}^{s}=\lambda_{n}^{s}(i)}$.

Subsequently, we propose two decentralised trackers, the Decentralised Natural Gradient Variational multi-object Trackers with Diminishing Stepsize (DeNG-VT-DS) and Decentralised Natural Gradient Variational multi-object Tracker with Gradient Tracking (DeNG-VT-GT), whose full procedures are given in Algorithm \ref{Algo:DeNG-VT-DS} and \ref{Algo:DeNG-VT-GT} in Appendix \ref{apx: tracker and pseudocodes}.

\subsubsection{Prediction and update steps} Similar to Section \ref{sec:Prediction and update steps}, the predictive prior is $\hat{p}_n(X_{n,k})=\mathcal{N}(X_{n,k};\mu^{k*,s}_{n|n-1},\Sigma^{k*,s}_{n|n-1})$, where $\mu^{k*,s}_{n|n-1},\Sigma^{k*,s}_{n|n-1}$ are computed according to \eqref{eq:predictive prior computation}. In the update step, DeNG-VT-DS follows the update in \eqref{eq:update equation 1}, while DeNG-VT-GT uses \eqref{eq:update equations for each variational parameter}-\eqref{eq:update equation 2}, where each iterative update also requires pre-computing $q_n^{s,*}(\theta_{n,j}^{s})$ in parallel using \eqref{eq:update theta} to facilitate computing $\boldsymbol{g}_i(\mathcal{L}_s)$. Here, $\boldsymbol{g}_i(\mathcal{L}_s)$ is the natural gradient, with detailed derivations provided below.
\vspace{-0.1em}
\subsubsection{Derivation of the natural gradient $\boldsymbol{g}_i(\mathcal{L}_s)$}\label{Derivation of the natural gradient}
%In this section, we rewrite the distributions using canonical exponential family distributions for the convenience of deriving the natural gradient of variational parameters.
First, we can rewrite the predictive prior $\hat{p}_n(X_{n,k})$ and the variational distribution $q_n(X_{n,k})$, $k=1,...,K$ at time step $n$ of the $s$-th sensor into the form of canonical exponential family distributions  $\hat{p}_n(X_{n,k}; \eta_{n,k}^{s} )$ and $ q_n(X_{n,k}; \lambda_{n,k}^{s})$:
\vspace{-0.5em}
\begin{align} \notag
   & \hat{p}_n(X_{n,k}; \eta_{n,k}^{s} )=  h(X_{n,k}) \exp \left( {\eta_{n,k}^{s}}^\top T(X_{n,k}) - A(\eta_{n,k}^{s}) \right) \\[-0.2em]\notag
 &   q_n(X_{n,k}; \lambda_{n,k}^{s})=  h(X_{n,k}) \exp \left( {\lambda_{n,k}^{s}}^\top T(X_{n,k}) - A(\lambda_{n,k}^{s}) \right)\\[-2.2em]\notag
\end{align}
where $h(\cdot)$ is the base function, $T(\cdot)$ is the sufficient statistic, $A(\cdot)$ is the log partition function, all for the Gaussian distribution. The natural parameters $\eta_{n,k}^{s}$ and 
$\lambda_{n,k}^{s}$ are defined as
\vspace{-1.3em}
\begin{align}  
 &\eta_{n,k}^{s}=\begin{bmatrix}
\eta_{n,k}^{s,1}\\\eta_{n,k}^{s,2}
 \end{bmatrix}
 =\begin{bmatrix}
    (\Sigma^{k*,s}_{n|n-1})^{-1} \mu^{k*,s}_{n|n-1} \\
    -\frac{1}{2} (\Sigma^{k*,s}_{n|n-1})^{-1}
 \end{bmatrix}\\ \label{eq: natural parameters definition}
 &\lambda_{n,k}^{s}=\begin{bmatrix}
\lambda_{n,k}^{s,1}\\\lambda_{n,k}^{s,2}
 \end{bmatrix}
 =\begin{bmatrix}
    (\Sigma^{k,s}_{n|n})^{-1} \mu^{k,s}_{n|n} \\
    -\frac{1}{2} (\Sigma^{k,s}_{n|n})^{-1}
 \end{bmatrix} \\[-2.em]\notag
\end{align}
%where $\mu^{k*,s}_{n|n-1},\Sigma^{k*,s}_{n|n-1}$ are computed according to \eqref{eq: predictive prior} based on the converged variational distribution $q^*_{n-1}(X_{n-1,k})= \mathcal{N}(X_{n-1,k};\mu^{k*,s}_{n-1|n-1},\Sigma^{k*,s}_{n-1|n-1})$ at time step $n-1$. % $q_n(X_{n,k})=\mathcal{N}(X_{n,k};\mu^{k,s}_{n|n},\Sigma^{k,s}_{n|n})$. 

Using \eqref{eq: OELBO each sensor}, \eqref{eq:local LM ELBO}, \eqref{eq:natural gradients the Fisher information matrix} and the property \eqref{simplify the gradient computation}, the natural gradient of the LM-ELBO simplifies into the following two parts after canceling the zero terms in $\nabla_{\lambda_{n}^{s}} \mathcal{F}_s(\lambda_n^s,\rho_n^s)|_{\rho_n^s=\rho_n^{s*}(\lambda_n^s)}$: 
\vspace{-0.1em}
\begin{align} \label{eq: natural gradient the EL-ELB}
\hat{\nabla}_{\lambda_{n}^{s}}\mathcal{L}_s(\lambda_{n}^{s})&= 
\hat{\nabla}_{\lambda_{n}^{s}}\mathcal{L}_s^1(\lambda_{n}^{s}) + \hat{\nabla}_{\lambda_{n}^{s}}\mathcal{L}_s^2(\lambda_{n}^{s}) \\[-1.8em]\notag
\end{align}
where $\mathcal{L}_s^1(\lambda_{n}^{s})=\frac{1}{N_s}\E_{q_n(X_n; \lambda_{n})}\log\frac{\hat{p}_n(X_n)}{q_n(X_n; \lambda_{n})}$, and $\mathcal{L}_s^2(\lambda_{n}^{s})=\E_{q_n(X_n; \lambda_{n})q_n^{s,*}(\theta_n^s)}\log p(Y_{n}^{s}|\theta_{n}^{s},X_{n})$, with $q_n^{s,*}(\theta_n^s)$ treated as constant that independent of $\lambda_n^s$ during gradient evaluation.
% \begin{align} \label{eq: natural gradient the EL-ELB}
% &\hat{\nabla}_{\lambda_{n}^{s}}\mathcal{L}_s(\lambda_{n}^{s})= 
% \hat{\nabla}_{\lambda_{n}^{s}}\mathcal{F}_s^1(\lambda_n^{s}) + \hat{\nabla}_{\lambda_{n}^{s}}\mathcal{F}_s^2(\lambda_n^{s},\rho_n^*(\lambda_n^s)),\\[-0.2em] \notag
% &\mathcal{F}_s^2(\lambda_n^{s},\rho_n^*(\lambda_n^s))=\E_{q_n(X_n; \lambda_{n})q_n(\theta_n; \rho_n^*(\lambda_n))}\log p(Y_{n}^{s}|\theta_{n}^{s},X_{n}),\\[-2em]\notag
% \end{align}
% $\mathcal{F}_s^1(\lambda_{n}^{s})=\frac{1}{N_s}\E_{q_n(X_n; \lambda_{n})}\log\frac{\hat{p}_n(X_n)}{q_n(X_n; \lambda_{n})}$, and the remaining part in $\nabla_{\lambda_{n}^{s}} \mathcal{F}_s(\lambda_n^s,\rho_n^s)|_{\rho_n^s=\rho_n^{s*}(\lambda_n^s)}$ equals zero.
%In the following, we will compute the expectations of each component in EL-ELBO and their gradient with respect to the natural parameters $\lambda_{n,k}^{s}$.

Here we use two different strategies to compute $\hat{\nabla}_{\lambda_{n}^{s}}\mathcal{L}_s^1(\lambda_{n}^{s})$ and $\hat{\nabla}_{\lambda_{n}^{s}}\mathcal{L}_s^2(\lambda_{n}^{s})$ in order to avoid calculating the FIM term $G(\lambda_{n}^{s})^{-1}$. Full derivations and the required exponential family properties are provided in Appendix \ref{apx: natural gradient derivation}, while only a brief derivation is presented in the remainder of this section.
Specifically, to compute $\hat{\nabla}_{\lambda_{n}^{s}}\mathcal{L}_s^1(\lambda_{n}^{s})$, we first compute the standard gradient $\nabla_{\lambda_{n}^{s}}\mathcal{L}_s^1(\lambda_{n}^{s})$, which has the following simple form: 
%where we compute the expectation  of each component in $\mathcal{L}_s^1(\lambda_{n,k}^{s})$ according to its exponential-family form, and then take its gradient with respect to natural parameter $\lambda_{n,k}^{s}$, and the result is 
\vspace{-0.5em}
\begin{align} \notag
\nabla_{\lambda_{n}^{s}}\mathcal{L}_s^1(\lambda_{n}^{s})=&\frac{1}{N_s} \sum\nolimits_{k=1}^{K} \!\!\nabla_{\lambda_{n}^{s}} \Big[(\eta_{n,k}^{s}- \lambda_{n,k}^{s})^\top \nabla_{\lambda_{n,k}^{s}}A(\lambda_{n,k}^{s})\\[-0.3em] \label{eq:natural gradient s1}
% &= \frac{1}{N_s} \sum_{k=1}^{K} \nabla_{\lambda_{n,k}^{s}}\left[ 
% \E_{q_n(X_{n,k}| \lambda_{n,k}^{s})} \left( \eta_{n,k}^{s} \cdot T(X_{n,k}) - A(\eta_{n,k}^{s}) \right) \right.\\ \notag
% & \left. \quad -  \E_{q_n(X_{n,k}| \lambda_{n,k}^{s})}\left( \lambda_{n,k}^{s} \cdot T(X_{n,k}) - A(\lambda_{n,k}^{s}) \right) \right] \\ \notag
& \qquad\qquad\quad+ A(\lambda_{n,k}^{s}))-A(\eta_{n,k}^{s})) \Big]
%  \\
% \label{eq:natural gradient s1}
% &= \frac{1}{N_s}  (\eta_{n,k}^{s}- \lambda_{n,k}^{s}) \nabla^2_{\lambda_{n,k}^{s}}A(\lambda_{n,k}^{s})
\\[-2.2em]\notag
\end{align}
%The derivations are in supplementary material.
%where the last line uses the property of the expectation of the natural sufficient statistics, i.e., $E[T(x)] = \nabla_{\eta} A(\eta)$.
Then, by using \eqref{eq:natural gradients the Fisher information matrix} and the property that $G(\lambda_{n}^{s})=\nabla^2_{\lambda_{n}^{s}} A(\lambda_{n}^{s})$,
the natural gradient $\hat{\nabla}_{\lambda_{n,k}^{s}}\mathcal{L}_s^1(\lambda_{n,k}^{s})$ for each natural parameter $\lambda_{n,k}^{s,1}$ and $\lambda_{n,k}^{s,2}$, $k=1,...,K$ is 
%Finally, We pre-multiply the gradient by the inverse Fisher information to find the natural gradient. We can derive that the natural gradient $\hat{\nabla}_{\lambda_{n,k}^{s}}\mathcal{L}_s^1(\lambda_{n,k}^{s})$ has the following  form:
\vspace{-0.5em}
\begin{align} \label{eq:natural gradient 1}
&\hat{\nabla}_{\lambda_{n,k}^{s,1}}\mathcal{L}_s^1(\lambda_{n,k}^{s})= \frac{1}{N_s}  (\eta_{n,k}^{s,1}- \lambda_{n,k}^{s,1}) \\\label{eq:natural gradient 2}
&\hat{\nabla}_{\lambda_{n,k}^{s,2}}\mathcal{L}_s^1(\lambda_{n,k}^{s})= \frac{1}{N_s}  (\eta_{n,k}^{s,2}- \lambda_{n,k}^{s,2}) \\[-2.2em]\notag
\end{align}
where the FIM term cancels out without needing computation.
% using the property that $G(\lambda_{n}^{s})=\nabla^2_{\lambda_{n}^{s}} A(\lambda_{n}^{s})$ when $q_n(X_{n}| \lambda_{n}^{s})$ is in the exponential family form.
%where we can see the natural gradient is equal to the difference in the natural parameters without the need of computation of the Fisher information matrix.

Next, to compute the second component $\hat{\nabla}_{\lambda_{n}^{s}}\mathcal{L}_s^2(\lambda_{n}^{s})$, we use the method from  \cite{khan2018fast} to avoid a direct computation of the FIM: define $m_{n,k}^{s}=\E_{q(X_{n,k};\lambda_{n,k}^s)} T(X_{n,k})$ as the mean sufficient statistics; then, the natural gradient with respect to $m_{n,k}^{s}$ equals to the gradient with respect to natural parameters (see Appendix \ref{apx: natural gradient derivation}), i.e., $\hat{\nabla}_{\lambda_{n}^{s}} \mathcal{L}_s(\lambda_{n}^{s})= \nabla_{m_{n}^{s}} \mathcal{L}_s(m_{n}^{s})$.
%khan2017variational
%We build upon the natural gradient method in \cite{khan2017conjugate}, which simplifies the update by avoiding a direct computation of the Fisher information matrix. The main idea is to use the expectation parameters of the exponential-family distribution to compute natural gradients in the natural-parameter space. 
%To compute the natural gradient of second part $\mathcal{L}_s^2(\lambda_{n,k}^{s})$, we use a neat trick of exponential families to simplify the computation as introduced in \cite{khan2017conjugate}.
Therefore, we can compute the standard gradient $\nabla_{m_{n}^{s}}\mathcal{L}_s^2(m_{n}^{s})$ instead of $\hat{\nabla}_{\lambda_{n}^{s}}\mathcal{L}_s^2(\lambda_{n}^{s})$. According to \cite{khan2018fast}, each $m_{n,k}^{s}$ has the following relationship with its Gaussian mean and covariance 
\vspace{-0.5em}
\begin{align} \label{eq:mean parameters}
 &m_{n,k}^{s}=\begin{bmatrix}
     m_{n,k}^{s,1}\\m_{n,k}^{s,2}
 \end{bmatrix}
 =\begin{bmatrix}
    \mu^{k,s}_{n|n} \\
    \mu^{k,s}_{n|n} [\mu^{k,s}_{n|n}]^\top+\Sigma^{k,s}_{n|n}
 \end{bmatrix}\\[-2.em]\notag
\end{align}
After substituting $m_{n,k}^{s}$ from \eqref{eq:mean parameters} and computing the expectations in $\mathcal{L}_s^2(m_{n}^{s})$ (see Appendix \ref{apx: natural gradient derivation}), we have
\vspace{-0.2em}
\begin{align} 
&\nabla_{m_{n}^{s}}\mathcal{L}_s^2(m_{n}^{s})\\[-0.3em]\notag
&=-\frac{1}{2} \sum\nolimits_{k=1}^K  \!\!\!\nabla_{m_{n}^{s}}\mathrm{Tr}\Big(\!H^\top\!(\overbar{R}_n^{k,s})^{-1}H(m_{n,k}^{s,2}\!-\!m_{n,k}^{s,1}(m_{n,k}^{s,1})^\top)\!\Big) \\[-0.3em]\notag
& -\frac{1}{2} \sum\nolimits_{k=1}^K \!\!\!\nabla_{m_{n}^{s}}  ( Hm_{n,k}^{s,1}\!-\!\overbar{Y}_n^{k,s})^\top (\overbar{R}_n^{k,s})^{-1} (Hm_{n,k}^{s,1}\!-\!\overbar{Y}_n^{k,s})
\end{align}
Subsequently, the natural gradients with respect to mean parameters $m_{n,k}^{s,1}$ and $m_{n,k}^{s,2}$ are
\vspace{-0.2em}
\begin{align} 
\hat{\nabla}_{m_{n,k}^{s,1}}\mathcal{L}_s^2(m_{n,k}^{s})&= H^\top  (\overbar{R}_n^{k,s})^{-1} \overbar{Y}_n^{k,s} \\[-2.5em]\notag
\end{align}
\begin{align} \label{eq:natural gradient 4}
\hat{\nabla}_{m_{n,k}^{s,2}}\mathcal{L}_s^2(m_{n,k}^{s})&= -\frac{1}{2}H^\top  (\overbar{R}_n^{k,s})^{-1}H\\[-2.2em]\notag
\end{align}
In sum, the total natural gradients can be obtained by using \eqref{eq: natural gradient the EL-ELB}, and \eqref{eq:natural gradient 1}-\eqref{eq:natural gradient 4}:
\vspace{-0.2em}
\begin{align} \notag
\hat{\nabla}_{\lambda_{n,k}^{s,1}}\mathcal{L}_s(\lambda_{n,k}^{s})&= \frac{1}{N_s}  \left[ (\Sigma^{k*,s}_{n|n-1})^{-1} \mu^{k*,s}_{n|n-1}-  (\Sigma^{k,s}_{n|n})^{-1} \mu^{k,s}_{n|n}\right] \\\label{eq:natural gradient total 1}
&\quad +H^\top  (\overbar{R}_n^{k,s})^{-1} \overbar{Y}_n^{k,s} \\[-.5em]\notag
\hat{\nabla}_{\lambda_{n,k}^{s,2}}\mathcal{L}_s(\lambda_{n,k}^{s})&= \frac{1}{2N_s}  [(\Sigma^{k,s}_{n|n})^{-1}-(\Sigma^{k*,s}_{n|n-1})^{-1}]\\\label{eq:natural gradient total 2}
&\quad -\frac{1}{2}H^\top  (\overbar{R}_n^{k,s})^{-1}H \\[-2.5em]\notag
\end{align}

\vspace{-0.5em}
\subsection{Robust decentralised tracking: explainable performance in limited iterations}\label{sec:Robust decentralised tracking}
Ideally, achieving consensus in the previous time step ensures identical distributions $\hat{p}_n(X_n; \eta_{n}^s)$ across different sensors at time step $n$. 
However, when (natural) gradient descent iterations are limited for efficiency before reaching convergence, sensors may in practice compute different priors $\hat{p}_n(X_{n};\eta_{n}^{s})$. In this case, our decentralised trackers still perform sensible inference, optimising the same LM-ELBO in \eqref{eq:locally maximised ELBO form}, but with a different prior, which can be interpreted as the geometric average (GA) \cite{clark2010robust,li2017generalized} fusion of the individual sensor priors: $\hat{p}_{eff}(X_n)\propto \prod_{s=1}^{N_s}  \hat{p}_n(X_{n};\eta_{n}^{s} ) ^{1/N_s}$, as fully derived in Appendix \ref{apx: Tracker Robustness}. Thus, it remains a reasonable fused prior.
%
%remains interpretable since it still maximises LM-ELBO in %\eqref{eq:local LM ELBO} with 
%$\hat{p}_n(X_n; \eta_{n})$ being replaced by an effective prior, which is the
Notably, in our proposed decentralised gradient-based VTs, this GA fusion occurs automatically without extra processing steps. This contrasts with traditional GA fusion approaches which necessitate separate consensus algorithms to implement a fully distributed GA fusion rule.

\vspace{-0.8em}
\section{Results}\label{sec:Results}
This section investigates empirical sensor fusion and tracking performance of the proposed methods under both fixed and time-varying sensor networks, with a detailed comparison to the following methods:
%and compare them with existing methods in tracking accuracy and communication efficiency. Two scenarios are simulated to study the sensor fusion and tracking performance of the compared methods under fixed and time-varying sensor network.  A detailed comparative examination of the following methods is undertaken:
\vspace{-\baselineskip}
\nomenclature{I-VT}{Individual variational multi-object tracker}
\nomenclature{C-VT}{Centralised variational multi-object tracker}
\nomenclature{DeC-VT}{Decentralised consensus-based variational multi-object tracker}
\nomenclature{DeG-VT-DS}{Decentralised gradient variational multi-object tracker with diminishing stepsize}
\nomenclature{DeG-VT-GT}{Decentralised gradient variational multi-object tracker with gradient tracking}
\nomenclature{DeNG-VT-DS}{Decentralised natural gradient variational multi-object tracker with diminishing stepsize}
\nomenclature{DeNG-VT-GT}{Decentralised natural gradient variational multi-object tracker with gradient tracking}
\nomenclature{DeAA-VT}{Decentralised arithmetic average variational multi-object tracker}
\printnomenclature
Specifically, in I-VT, each sensor runs variational multi-object tracker independently. C-VT is a baseline optimal fusion method that receives all measurement from all sensors, detailed in Section \ref{sec:ca update}. Among them, our proposed methods in this paper are the decentralised (natural) gradient variational multi-object trackers, including DeG-VT-DS, DeG-VT-GT, DeNG-VT-DS, and DeNG-VT-GT in Algorithm 1-4 in the supplementary material. In addition, we compare with DeC-VT algorithm in \cite{li2023consensus} to showcase our improvement in communication efficiency. We also include compare with a commonly-used suboptimal distributed arithmetic average (AA) fusion strategy \cite{li2017generalized,li2021distributed}, where each sensor infers a multi-object posterior distribution using the variational tracker in \cite{gan2024variational} based on local measurements, then a distributed average consensus algorithm is implemented to fuse the multi-object posteriors from each sensor using the AA fusion principle. 
%Note that we only compare sensor fusion methods built upon the variational multi-object tracker in \cite{gan2024variational,gan2022variational} since these variational trackers have demonstrated to outperform state-of-the-art trackers such as ET-JPDA filter \cite{yang2018linear}, SPA-based tracker \cite{meyer2020scalable,yang2018linear} and RFS-based PMBM filter \cite{granstrom2019poisson}. More comparison details can be found in  \cite{gan2024variational,gan2022variational}.

\begin{figure}
    \centering
    \includegraphics[width=8cm]{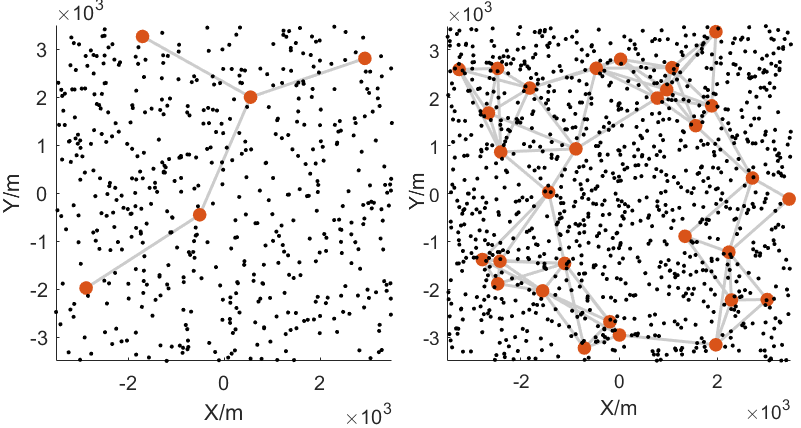}
    \caption{Sensor networks of dataset 1 and 2 in Scene 1; Red circles are sensor nodes, grey lines denote their connectivity, and black dots are an example measurement data of one time step at a single sensor}
    \label{fig:fixed sensor network}
    \vspace{-1.4em}
\end{figure}

\begin{figure}[tp!]
\centerline{\includegraphics[width=8.5cm]{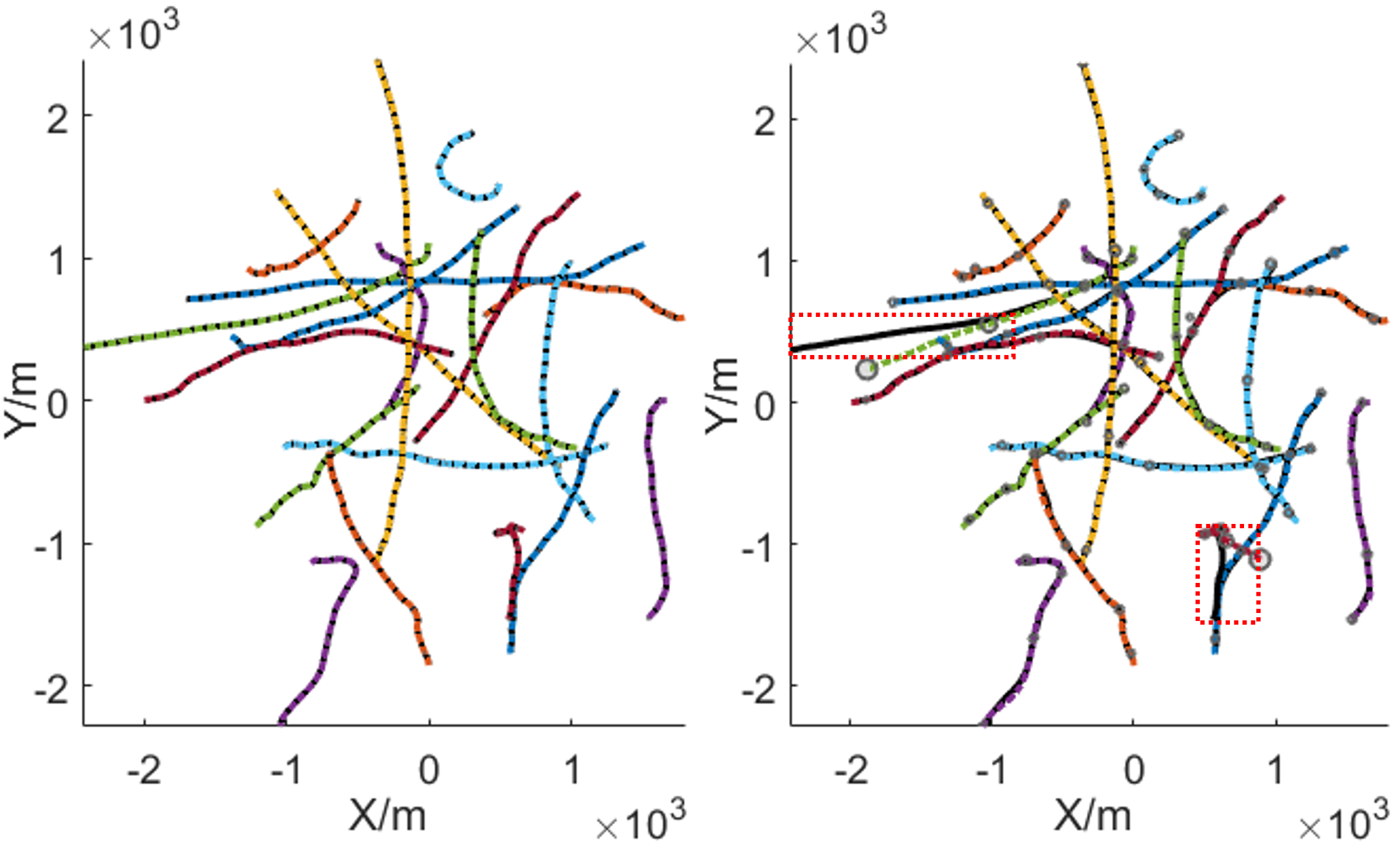}}
\caption{Example tracking performance at one Monte Carlo run of DeNG-VT-GT (left) and DeAA-VT (right); coloured dotted lines are estimate, black lines are ground truth and grey ellipses are 95\% confidence interval. The boxes in the right figure mark the track loss events using DeAA-VT}
\label{fig:estimate_simu1}
\vspace{-1.5em}
\end{figure}
% \begin{figure}
%     \centering
%     \includegraphics[width=8cm]{figure/gttrack_meas2.png}
%     \caption{Sensor network; Red circles are sensor nodes and grey lines indicate their connectivity .}
%     \label{fig:fixed sensor network}
% \end{figure}

\vspace{-1em}
\subsection{Performance Metrics}
We use the following metrics to evaluate the performance. %of proposed algorithms and to compare them with other methods. 
\subsubsection{Generalised optimal sub pattern assignment (GOSPA)}
The GOSPA  distance \cite{rahmathullah2017generalized} is used to evaluate the tracking accuracy, where the order  $p=1$, $\alpha=2$, and the cut-off distance  $c=50$. Concurrently, GOSPA metric returns localisation errors for well-tracked objects, the missed object errors and false object errors. Here, we have a fixed number of objects in the scene; thus, the missed and false object errors denote the track loss rather than the disappearance or appearance of objects. We define a MGOSPA metric, which is the mean GOSPA averaged over all sensors and all time steps.
\subsubsection{Communication Iteration (CI)}
To show the communication cost, we define CI as the total iteration number that sensors pass messages to its neighbours at a time step, averaged over total time steps and Monte Carlo runs. Specifically, for decentralised (natural) gradient-based VB trackers, CI is the total iteration number of the decentralised (natural) gradient descent algorithms, which also equals to the variational update iterations at each time step; For DeC-VT \cite{li2023consensus}, CI equals to the total variational update iterations at each time step multiplies the consensus algorithm iterations at each variational update iteration. For the suboptimal DeAA-VT, CI equals to total iterations of consensus algorithm performed at one time step.

%\subsection{Design and Settings of Simulations and Comparisons}
%We simulated two scenarios to study the sensor fusion and tracking performance of the compared methods. In case 1, 

%the first dataset is a simpler scenario with fewer sensors, a lower clutter rate, and fewer targets, and the second dataset is more challenging with a more complicated sensor network, heavier clutter, and denser targets.

%We design various cases of ....in order to demonstrate the adaptiveness of the proposed method. An extensive comparison analysis of ... is provided, including ...., and  detailed comparisons with .... 

%In Case 1, we validate the efficacy of the proposed .... schemes over other suboptimal methods in tracking scenarios with .... Under this ... setting, . In Case 2, we analyse the advantages and generality of the proposed ..... Case 3 verifies the robustness of the proposed method in t.... In Section ..., we demonstrate the proposed...

%Other general parameter settings are as follows. 

% \begin{figure}
%     \centering
%     \includegraphics[width=9cm]{figure/nggt_conve1.png}
%     \caption{Convergence of DNG-GT-VMOT at one time step}
%     \label{fig:fixed sensor network}
% \end{figure}

% \begin{figure}
%     \centering
%     \includegraphics[width=9cm]{figure/OSPA_ITER.png}
%     \caption{Convergence at one time step}
%     \label{fig:fixed sensor network}
% \end{figure}
\vspace{-0.7em}
\subsection{Scene 1: Distributed Sensor fusion and multi-object tracking under fixed network connectivity}\label{result:case1}
\subsubsection{Simulation settings}\label{Scene 1 Simulation settings}
In Scene 1, we analyse sensor fusion and tracking performance of compared methods with time-invariant sensor network in two datasets with different sensor number and detection environments. Two different sensor networks are simulated as shown in Figure \ref{fig:fixed sensor network}, in which their location and connectivity are randomly generated. All sensors observe the same surveillance area and follow the NHPP measurement model in Section \ref{NHPP measurement model and association prior} with $R_k^{s}=100\text{I}$. Specifically, in dataset 1, there are 5 sensors, and for each sensor, the object Poisson rates are set to 2 and the clutter rate is 500; in dataset 2, we have 30 sensors with object and clutter Poisson rates being 1 and 1000, which is more challenging for a single sensor to track objects properly since there is frequent missed detection and object measurements are buried in clutter. 

For all datasets, we consider the case that there are 20 objects in the surveillance area, moving under the constant velocity dynamical model defined in Section \ref{dynamic model}, with parameters being $F_{n,k}^d=\begin{bmatrix} 1 & \tau \\ 0& 1 \end{bmatrix},Q_{n,k}^d=36\begin{bmatrix} \tau^3/3 & \tau^2/2 \\ \tau^2/2& \tau
\end{bmatrix}$ ($d=1,2$). The total time steps are 50, and the time interval between observations is $\tau=1$s. To verify the robustness of the compared algorithms, we simulate 50 Monte Carlo (MC) runs for each dataset. In particular, for dataset 1, each MC run generate different ground-truth tracks and measurements according to the defined parameter settings, while in dataset 2, we have 50 different measurement data generated with the same ground-truth tracks shown in Figure \ref{fig:fixed sensor network}. 

Other general parameter settings are as follows. For DeNG-VT-GT, the fixed stepsize $\alpha=0.8$ for both dataset 1 and 2. For DeG-VT-GT, $\alpha$ is set to 5 and 10 for dataset 1 and 2, respectively. 
%In the case of DeG-VT-DS, we apply a diminishing step size \(\varepsilon/(i+1)^{\kappa}\), where \(\varepsilon = 1\), \(\kappa = 0.1\), and \(i\) denotes the iteration number. As shown in \cite{zeng2018nonconvex}, the condition \(\kappa \in (0,1]\) ensures convergence to a stationary point. For DeNG-VT-DS, we implement both the diminishing step size \(\varepsilon/(i+1)^{\kappa}\) with \(\varepsilon = 1\), \(\kappa = 0.1\), and a self-tuned diminishing step size with \(\varepsilon = 50\), \(\kappa = 2\). The latter provides potentially faster convergence.
In the case of DeG-VT-DS, we apply a diminishing stepsize \(\alpha_i=1/(i+1)^{\kappa}\), where \(i\) denotes the iteration number. As studied in \cite{zeng2018nonconvex}, the condition \(\kappa \in (0,1]\) ensures convergence to a stationary point, and here we set $\kappa$ to 0.1 and 0.01 for dataset 1 and 2, respectively. We present results of DeNG-VT-DS with two different diminishing stepsizes, denoted as DeNG-VT-DS1 and DeNG-VT-DS2. For DeNG-VT-DS1, we adopt \(\alpha_i=1/(i+1)^{\kappa}\) with \(\kappa =0.5\) and 0.1 for dataset 1 and 2, respectively. For DeNG-VT-DS2, we apply a fine-tuned diminishing stepsize $\alpha_i=\varepsilon/(i+1)^{\kappa}$ that may converge faster, with \(\varepsilon = 20\), \(\kappa = 2\) for both dataset 1 and 2.

\begin{figure}[tp!]
    \centering
    \includegraphics[width=8.5cm]{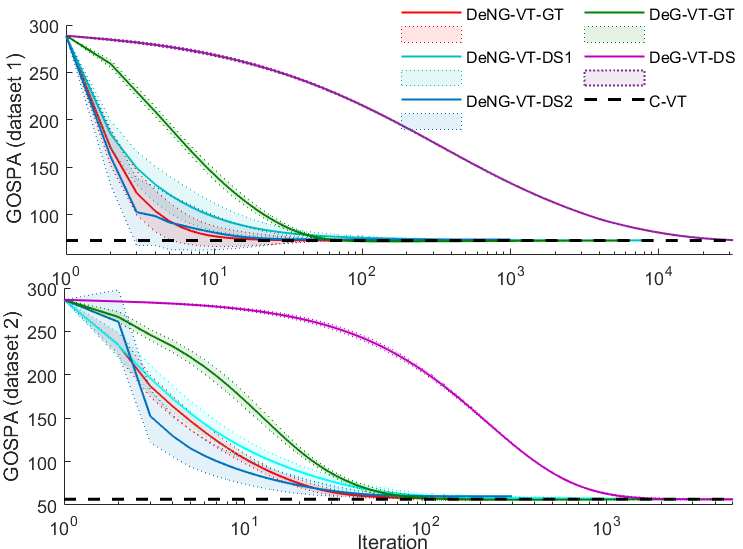}
    \caption{GOSPA over iteration number at a single time step; lines and shaded area are mean and $\pm1$ standard deviation of GOSPA value averaged over all sensors, respectively.}
    \label{fig:GOSPA over iteration}
     \vspace{-1.3em}
\end{figure}

\subsubsection{Result 1: analysis of convergence speed of decentralised gradient-based variational trackers at a single time step}\label{Result 1}
In the first simulation, we select a single time step measurement data from one MC run in both dataset 1 and 2 to perform inference tasks to analyse the convergence performance of the proposed decentralised gradient-based methods, including DeNG-VT-GT, DeNG-VT-DS, DeG-VT-GT, DeG-VT-DS. %In detail, we choose  35th time step measurements data from 2nd MC run in dataset2. 
%Specifically, for DeG-VT-DS, we use a diminising stepsize $\varepsilon/(i+1)^{\kappa}$ with $\varepsilon=1$, $\kappa=0.1$ and $i$ is the iteration number; when $\kappa \in (0,1]$, it guarantees to converge to a stationary point as analysed in \cite{zeng2018nonconvex}. For DeNG-VT-DS, we use both the diminising stepsize $\varepsilon/(i+1)^{\kappa}$ with $\varepsilon=1$, $\kappa=0.1$, and a self-turned diminising stepsize with $\varepsilon=50$, $\kappa=2$ that may fast convergence speed.
To make a fair comparison, we assume the same converged variational distribution at the previous time step for all methods such that they have the same predictive prior. All other settings are the same as in Section \ref{Scene 1 Simulation settings}.

The convergence speed of the proposed methods is evaluated using GOSPA values, with the mean and standard deviation plotted across all local sensors over iterations, as shown in Figure \ref{fig:GOSPA over iteration}. The standard deviations of all compared methods gradually converge to zero, indicating that they reach consensus and each sensor shares the same estimates. Across all datasets, DeNG-VT-GT demonstrates the fastest convergence, followed by DeNG-VT-DS2, DeNG-VT-DS1, DeG-VT-GT, and DeG-VT-DS, with DeG-VT-DS showing significantly slower convergence than the others. While DeNG-VT-DS2 accelerates convergence due to its fine-turned diminishing step size compared to DeNG-VT-DS1, it deviates very slightly from the centralised C-VT solution. Meanwhile, all other methods match the performance of C-VT, empirically demonstrating their equivalence in tracking performance to C-VT.

\begin{figure}[tp!]
    \centering
    \includegraphics[width=8.5cm]{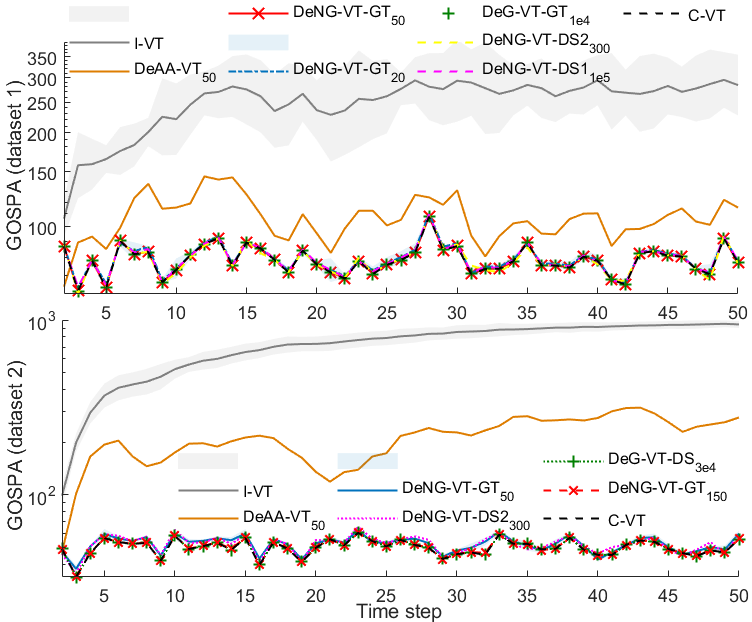}
    \caption{GOSPA over 50 time steps; for all methods, lines are means of GOSPA averaged over all sensors and shaded areas indicate $\pm1$ standard deviation. Y-axis is log-scale.
   }
    \label{fig:Mean GOSPA of all sensors over 50 time steps}
     \vspace{-1.2em}
\end{figure}

\subsubsection{Result 2: comparison of all methods for one single MC run}
Having assessed the performance of the proposed methods at a single time step, we now extend this analysis over all time steps in a single MC run to evaluate convergence and and communication efficiency of DeNG-VT-GT, DeNG-VT-DS, DeG-VT-GT, and compare their tracking accuracy with other methods. We exclude DeG-VT-DS from this evaluation due to its much slower convergence speed, as detailed in Result 1 in Section \ref{Result 1}. 

Figure \ref{fig:estimate_simu1} illustrates mean GOSPA with its one standard deviation over 50 time steps for each compared methods for one MC run in both dataset 1 and 2. The subscript of each method in the figure legend represents the iteration number, i.e., the CI metric, to reflect their communication cost. The results show that all methods except I-VT achieve zero standard deviation at each time step, indicating that all sensor nodes consistently converge to the same values, thus demonstrating their capability to reach a local optimum. 
Most importantly, Figure \ref{fig:Mean GOSPA of all sensors over 50 time steps} confirms empirically the equivalence in tracking performance at every time step between the centralised fusion C-VT and our proposed decentralised solutions, including DeNG-VT-GT, DeNG-VT-DS, and DeG-VT-GT. The significant discrepancy in mean GOSPA between the suboptimal DeAA-VT and our gradient-based methods highlights our superior tracking accuracy. Notably, DeNG-VT-GT not only achieves lower GOSPA values with the same communication cost as DeAA-VT but also matches the performance of C-VT with much lower communication cost compared to other decentralised gradient-based methods.
%It can be seen that the discrepancy in mean GOSPA between the suboptimal DeAA-VT and our proposed gradient-based methods is significant. Remarkably, with the same communication cost as required by DeAA-VT, our DeNG-VT-GT method achieves substantially lower GOSPA values, and DeNG-VT-GT is shown to achieve empirically equivalent performance to the centralised C-VT in a much lower communication cost, compared to other decentralised gradient-based methods.

To show the difference in tracking accuracy more directly,  Figure \ref{fig:estimate_simu1} plots the estimates of DeNG-VT-GT and DeAA-VT. The results demonstrate that DeNG-VT-GT consistently tracks all targets with high accuracy, whereas DeAA-VT frequently loses track and exhibits greater uncertainty in its estimates.

\begin{table}[tp!]
\centering
\caption{Performance of compared methods in dataset 1}
\setlength{\tabcolsep}{2.5pt} % Adjust space bet columns
\begin{tabular}{*6c}
\toprule
method & MGOSPA & location & missed & false    & CI \\
\midrule
C-VT   & 76.9 $\pm$ 1.3 &76.9 $\pm$ 1.3 &0 $\pm$ 0 &0 $\pm$ 0 & --   \\
DeC-VT  & 76.9 $\pm$ 1.3  &76.9 $\pm$ 1.3   &0 $\pm$ 0 &0 $\pm$ 0 & 400    \\
DeNG-VT-GT & 77.7$\pm$ 1.3 &77.7$\pm$ 1.3&0 $\pm$ 0 &0 $\pm$ 0 &  20 \\
\textbf{DeNG-VT-GT} & 76.9 $\pm$ 1.3 &76.9 $\pm$ 1.3 &0 $\pm$ 0 &0 $\pm$ 0 &  50 \\
DeNG-VT-DS$_{2}$ & 77.2 $\pm$ 1.3 &77.2 $\pm$ 1.3&0 $\pm$ 0 &0 $\pm$ 0 &  300 \\
DeG-VT-GT & 78.4 $\pm$ 1.4 &78.4 $\pm$ 1.4&0 $\pm$ 0 &0 $\pm$ 0 &5000 \\
DeG-VT-GT & 76.9 $\pm$ 1.3 &76.9 $\pm$ 1.3 &0 $\pm$ 0 &0 $\pm$ 0 &  1e4 \\
DeG-VT-DS & 76.9 $\pm$ 1.3 &76.9 $\pm$ 1.3 &0 $\pm$ 0 &0 $\pm$ 0 & 1e5\\
DeAA-VT &   103.2 $\pm$ 2.9 &103.2 $\pm$ 2.9 &0 $\pm$ 0&0 $\pm$ 0 &  20  \\
DeAA-VT &   103.1 $\pm$ 2.9 &103.1 $\pm$ 2.9 &0 $\pm$ 0&0 $\pm$ 0 &  100  \\
I-VT  & 218.7 $\pm$ 15.3 & 166.5 $\pm$ 2.7 & 26.1 $\pm$ 8.6 & 26.1 $\pm$ 8.6  & -- \\
\bottomrule 
\end{tabular}
\label{Performance of compared methods over 50 runs1}
\end{table}

\begin{table}[tp!]
\centering
\caption{Performance of compared methods in dataset 2}
\setlength{\tabcolsep}{3pt} % Adjust space bet columns
\begin{tabular}{*6c}
\toprule
method & MGOSPA & location & missed & false    & CI \\
\midrule
C-VT   & 50.1 $\pm$ 0.7 &50.1 $\pm$ 0.7 &0 $\pm$ 0 &0 $\pm$ 0 & --   \\
DeC-VT  & 50.1 $\pm$ 0.7 &50.1 $\pm$ 0.7  &0 $\pm$ 0 &0 $\pm$ 0 & 1200    \\
DeNG-VT-GT & 51.9 $\pm$ 0.7 &51.9 $\pm$ 0.7&0 $\pm$ 0 &0 $\pm$ 0 &  50 \\
\textbf{DeNG-VT-GT }& 50.1 $\pm$  0.7 &50.1 $\pm$ 0.7 &0 $\pm$ 0 &0 $\pm$ 0 &  150 \\
DeNG-VT-DS$_{2}$ & 51.8 $\pm$ 1 &51.8 $\pm$ 1&0 $\pm$ 0 &0 $\pm$ 0 &  300 \\
DeG-VT-GT & 53.2 $\pm$ 1 &53.2 $\pm$ 1&0 $\pm$ 0 &0 $\pm$ 0 &  1e4 \\
DeG-VT-GT & 50.1 $\pm$  0.7 &50.1 $\pm$  0.7&0 $\pm$ 0 &0 $\pm$ 0 &  2e4 \\
DeAA-VT & 193.4 $\pm$ 13 &176.8 $\pm$ 5 &8.3$\pm$7 &8.3$\pm$7 &  100  \\
I-VT  & 734.1 $\pm$ 8 & 108.1 $\pm$ 3 & 313 $\pm$ 5 & 313 $\pm$ 5  & -- \\
\bottomrule 
\end{tabular}
\label{Performance of compared methods over 50 runs2}
\end{table}
\subsubsection{Result 3: Tracking and fusion performance over all 50 runs} 
We verify the robustness of the proposed and compared methods by testing it over 50 Monte Carlo runs in two different datasets under the general settings in Section \ref{Scene 1 Simulation settings}. Table \ref{Performance of compared methods over 50 runs1} and \ref{Performance of compared methods over 50 runs2} show the performance of the compared methods in both tracking accuracy and communication efficiency. We record the mean and one standard deviation of MGOSPA and its submetric (location error, missed object and false object error), averaged over 50 runs. For both datasets, we can see that C-VT and all versions of proposed (natural) gradient based methods show very accurate tracking. In contrast, the tracking accuracy of I-VT and DeAA-VTs is much lower. The estimation results also confirm the equivalence in tracking performance of the proposed DeNG-VT-GT, DeNG-VT-DS, DeG-VT-GT, DeG-VT-DS with the centralised C-VT solution when it converges.

With regards to communication costs, we can see from CI values the great advantage of the proposed DeNG-VT-GT compared with the DeC-VT, DeNG-VT-DS, DeG-VT-GT, and DeG-VT-DS, under the same optimal tracking accuracy. Compared to the suboptimal DeAA-VT method, we can see that our method still greatly outperforms DeAA-VT in tracking accuracy even using the same communication iteration number, which showcases its advantages in both tracking accuracy and communication efficiency.

\vspace{-1em}
\subsection{Scene 2: Distributed Sensor fusion and multi-object tracking under time-varying network connectivity}\label{result:case2}
\begin{figure}[tp!]
    \centering
    \includegraphics[width=8cm]{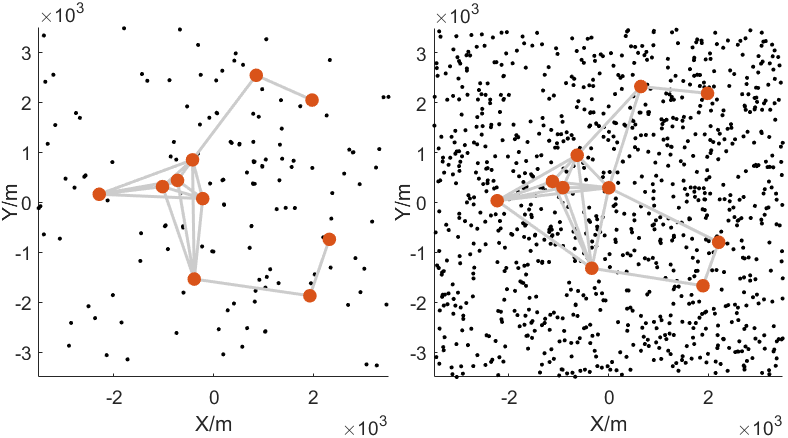}
    \caption{Time varying sensor networks;  Red circles are sensor nodes and grey lines indicate their connectivity. Black dots are measurements received at 15th time step from 1st sensor (left) and 38th time step from 10th sensor (right) }
    \label{fig:Time varying sensor networks}
     \vspace{-1.5em}
\end{figure}

In Scene 2, we simulate a more challenging scenario of a time-varying heterogeneous sensor network in which their location and connectivity are changing over time as shown in Figure \ref{fig:Time varying sensor networks}. In the surveillance area, there are 50 targets moving under the constant velocity model in Section \ref{dynamic model}, with parameters being $F_{n,k}^d=\begin{bmatrix} 1 & \tau \\ 0& 1 \end{bmatrix},Q_{n,k}^d=25\begin{bmatrix} \tau^3/3 & \tau^2/2 \\ \tau^2/2& \tau
\end{bmatrix}$ ($d=1,2$). All sensors observe the same surveillance area and follow the NHPP measurement model in Section \ref{NHPP measurement model and association prior} with $R_k^{s}=100\text{I}$. Specifically, we consider 10 heterogeneous sensors of different detection ability, with their clutter rate ranging from 100 to 1000 while the target rate for all sensors are one, meaning that some sensors' measurements are heavily cluttered. To verify the robustness of the compared algorithms, we simulate 50 MC runs with different ground-truth tracks and measurements according to the parameter settings. For all datasets, the total time steps are 50, and the time interval between observations is $\tau=1$s. 

For DeNG-VT-GT, the fixed stepsize $\alpha=0.8$. For DeG-VT-GT, $\alpha=10$. 
In DeG-VT-DS, we apply a diminishing step size \(1/(i+1)^{\kappa}\), where \(\kappa = 0.5\), and \(i\) denotes the iteration number. For DeNG-VT-DS, we implement both the diminishing stepsize \(1/(i+1)^{\kappa}\) with \(\kappa = 0.5\), and a self-tuned diminishing stepsize $\alpha_i=\varepsilon/(i+1)^{\kappa}$ with \(\varepsilon = 20\), \(\kappa = 1\). The latter provides potentially faster convergence.

% \begin{figure}[tp!]
%     \centering
%     \includegraphics[width=7cm]{figure/est_scene2_ALL1}
%     \caption{Example tracking performance of I-VT at one Monte Carlo run at 1st sensor (left) and 10th sensor (right); coloured dotted lines are estimate, black lines are ground truth and grey ellipses are 95\% confidence interval. }
%     \label{fig:Example tracking performance of I-VT }
%     \vspace{-1em}
% \end{figure}

\subsubsection{Result 1: analysis of convergence speed of decentralised gradient-based variational trackers at a single time step}\label{scene 2 Result 1}
\begin{figure}[tp!]
    \centering
    \includegraphics[width=8.5cm]{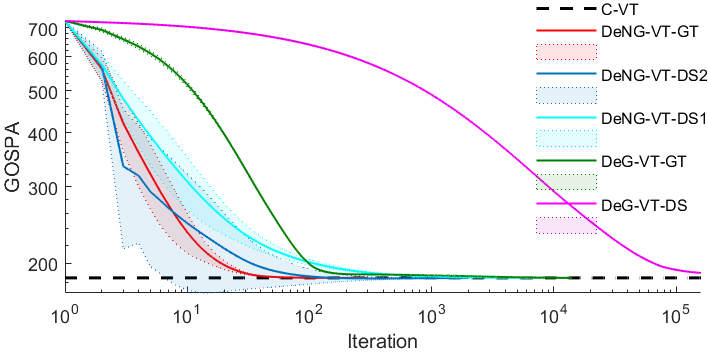}
    \caption{GOSPA over iteration number at a single time step; lines and shaded area are mean and $\pm1$ standard deviation of GOSPA value averaged over all sensors, respectively. Y-axis is log-scale.}
    \label{fig:GOSPA over iteration2}
    \vspace{-1.5em}
\end{figure}

First, we select a single time step measurement data from one MC run to analyse the convergence performance of the proposed decentralised gradient-based methods. % To make a fair comparison, we assume the same converged variational distribution at the previous time step for all methods such that they have the same predictive prior. All other settings are the same as in Section \ref{Scene 1 Simulation settings}.
Figure \ref{fig:GOSPA over iteration2} shows that the standard deviations of all compared methods gradually converge to zero, indicating that they reach consensus and each sensor shares the same estimates. Across all datasets, DeNG-VT-GT converges the fastest, followed by DeNG-VT-DS2, DeNG-VT-DS1, DeG-VT-GT, and DeG-VT-DS. Meanwhile, all methods match the performance of C-VT, empirically demonstrating their equivalence in tracking performance to the centralised C-VT solution.

\subsubsection{Result 2: comparison of all methods for one single MC run}
Having assessed the performance of the proposed methods at a single time step, we now extend this analysis over all time steps in a single MC run to evaluate convergence and and communication efficiency of DeNG-VT-GT, DeNG-VT-DS, DeG-VT-GT, and compare their tracking accuracy with other methods. Since DeG-VT-DS showing significantly slower convergence than the others in Section \ref{scene 2 Result 1}, we exclude DeG-VT-DS from this evaluation. 

Figure \ref{fig:estimate_simu1} illustrates mean GOSPA with its one standard deviation over 50 time steps for each compared methods. The subscript of each method in the figure legend represents the iteration number. The results show that all methods except I-VT achieve zero standard deviation at each time step, indicating that all sensor nodes reach consensus and consistently converge to the same values. As shown in fixed network scenarios in Section \ref{result:case1}, it shows in Figure \ref{fig:Mean GOSPA of all sensors over 50 time steps} that our proposed decentralised solutions are empirically equivalence in tracking performance to the C-VT. Additional, DeNG-VT-GT again shows much better tracking accuracy under the comparable communication cost as DeAA-VT, and are much efficient with regards to communication cost compared to other decentralised gradient-based methods.

\subsubsection{Result 3: Tracking and fusion performance over all 50 runs} 
We verify the robustness of the proposed method by testing it over 50 Monte Carlo runs with different measurement sets. Table \ref{Performance of compared methods over 50 runs3} shows the performance of the compared methods in both tracking accuracy and communication efficiency. We can see that C-VT, DeC-VT, and all versions of DeNG-VTs show very accurate tracking, The centralized method C-VT and several decentralised variants, including DeC-VT, the DeNG-VT-GT, DeNG-VT-DS2, and DeG-VT-GT, all obtain the same performance metrics with the same tracking accuracy and no missed or false targets. DeNG-VT-DS1 shows similar performance to the optimal group but with a marginally higher MGOSPA, indicating a slight decrease in efficiency. In contrast, DeAA-VT and I-VT exhibit significantly poorer performance with much higher MGOSPA values and substantial numbers of missed and false detections. The estimation results also confirm the equivalence of the proposed DeNG-VT with the centralised C-VT solution when it converges.

It is observed that, DeNG-VT-GT, can achieve performance on par with the centralised C-VT, requiring less communication cost compared to other methods, thus highlighting their potential for efficient and accurate tracking in scenarios requiring minimal communication overhead.

\begin{figure}[tp!]
    \centering
    \includegraphics[width=8.5cm]{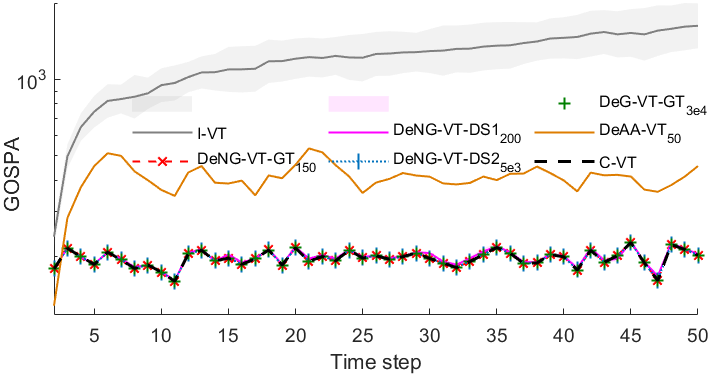}
    \caption{GOSPA over 50 time steps at a single MC run; lines are means of GOSPA averaged over all sensors and shaded areas indicate $\pm1$ standard deviation. Y-axis is log-scale.
   }
    \label{fig:Mean GOSPA over 50 time steps}
    \vspace{-1.em}
\end{figure}

\begin{table}[htp!]
\centering
\caption{Performance of compared methods in Scene 2}
\setlength{\tabcolsep}{3pt} % Adjust space bet columns
\begin{tabular}{*6c}
\toprule
method & MGOSPA & location & missed & false    & CI \\
\midrule
C-VT   & 196.4 $\pm$ 2.3 &196.4 $\pm$ 2.3&0 $\pm$ 0 &0 $\pm$ 0 & --   \\
DeC-VT  & 196.4 $\pm$ 2.3 &196.4 $\pm$ 2.3   &0 $\pm$ 0 &0 $\pm$ 0 & 3e3    \\
% DeNG-VT-GT &198.5 $\pm$ 4.6 &197.9 $\pm$ 2.3 &0.3 $\pm$ 2.1 &0.3 $\pm$ 2.1  &  50 \\
\textbf{DeNG-VT-GT} &196.4 $\pm$ 2.3 &196.4 $\pm$ 2.3&0 $\pm$ 0 &0 $\pm$ 0 &  150 \\
DeNG-VT-DS1 & 198.1 $\pm$ 2.3 &198.1 $\pm$ 2.3&0 $\pm$ 0 &0 $\pm$ 0 &  200 \\
% DeNG-VT-DS2 & 197.1 $\pm$ 5.5 &196.3 $\pm$ 2.3&0.4 $\pm$ 2.5 &0.4 $\pm$ 2.5 &  5e3 \\
DeNG-VT-DS2 & 196.4 $\pm$ 2.3 &196.4 $\pm$ 2.3&0 $\pm$ 0 &0 $\pm$ 0 &  1e4 \\
DeG-VT-GT & 196.4 $\pm$ 2.3 &196.4 $\pm$ 2.3 &0 $\pm$ 0 &0 $\pm$ 0 &3e5 \\
DeAA-VT &  437.6 $\pm$ 24 &419.6 $\pm$ 12 &9.0 $\pm$ 12&9.0 $\pm$ 12 &  50  \\
I-VT  & 1232 $\pm$ 21 & 294.4 $\pm$ 2.8 & 23.4 $\pm$ 11 & 23.4 $\pm$ 11 & -- \\
\bottomrule 
\end{tabular}
\label{Performance of compared methods over 50 runs3}
\end{table}

\vspace{-0.1em}
\section{Conclusion}
This paper presents decentralised multi-object tracking algorithms for cluttered environments in time-varying sensor networks. Our approaches achieve tracking performance on par with centralised fusion, outperform suboptimal distributed fusion strategies in accuracy, and greatly reduce communication costs compared to existing average consensus VT methods. Furthermore, our decentralised trackers remain robust under practical constraints, such as limited gradient descent iterations, while still delivering reliable and explainable inference. Future improvements include the integration of new advanced decentralised optimisation techniques, and extending this framework to accommodate unknown numbers of objects and multimodal sensors with varying spatial coverage.
%This paper presents decentralised multiple object tracking algorithms in cluttered environments under a time-varying sensor network. This approach matches the centralised fusion tracking performance, surpasses the traditional suboptimal distributed VT strategies in tracking accuracy, and demonstrates significant reductions in communication costs when compared to existing average consensus VT methods. There is scope to include future improvements in decentralised optimisation algorithms, such as gradient (tracking) approaches with improved properties, within our general framework. Current developments are extending the work to  accommodate unknown numbers of objects and multimodal sensors with varying spatial coverage.  

% \vspace{-10pt}
% \newpage

% \bibliographystyle{IEEEtran}
% \bibliography{./IEEEabrv,IEEEexample}
\putbib[IEEEexample] 

\end{bibunit}

\onecolumn

\begin{center}
{
\LARGE\bfseries Decentralised Variational Inference Frameworks \\\vspace{0.2em} for Multi-object Tracking on Sensor Networks: Additional Notes}
\end{center}

\appendices
\begin{bibunit}[IEEEtran]
\setcounter{subsection}{0}
\renewcommand{\thesubsection}{\thesection.\arabic{subsection}}

\makeatletter
\renewcommand{\subsection}[1]{%
  \refstepcounter{subsection}% Increment subsection counter
  \vspace{1.5\baselineskip}%
  \noindent%
  \thesubsection\ #1\par% Use the updated subsection number
  \vspace{0.5\baselineskip}%
}
\makeatother

\section{Derivations of centralised CAVI} \label{apx: centralised CAVI derivation}
In this part, we present detailed derivation for the  update step of $q_n(X_n)$ and $q_n(\theta_n)$ in the centralised coordinate
ascent variational inference (CAVI) in Section \ref{sec:ca update}.

\subsection{Update for $q_n(X_n)$} \label{sec:ca update for Xn} 

First we present the update for $q_n(X_n)$. According to the standard CAVI update rule in \cite{blei2017variational}, we have
\begin{equation}
      q_n(X_n)
\propto\text{exp}\left(\E_{q_n(\theta_n)}\log \hat{p}_n(X_n,\theta_n,Y_n)\right)
\end{equation}
where the expression of $\hat{p}_n(X_n,\theta_n, Y_n)$ is written as:
\begin{align} \label{eq: pn}
\hat{p}_n(X_{n},\theta_{n}, Y_n)  &=p(Y_n|\theta_n,X_n)p(\theta_n|M_n)\hat{p}_n(X_n).\\[-1.8em]\notag
\end{align}
Thus, we can further derive the update as follows using \eqref{eq: pn}, \eqref{eq: joint likelihood conditionally independent}, \eqref{eq:obs prior}, and the expression of $\ell({Y}_{n,j}^{s}|X_{n,k})$ in \eqref{measurement model}:
\begin{align} \notag
    q_n(X_n)
  &  \propto \hat{p}_n(X_n)\text{exp}\left(\E_{q_n(\theta_n)}\log p(Y_n|\theta_n,X_n) \right)\\ \notag
   &=  \hat{p}_n(X_n)\text{exp} \sum_{s=1}^{N_s} \sum_{j=1}^{M_n^{s}} \E_{q_n(\theta_{n,j}^{s})}\log \ell({Y}_{n,j}^{s}|X_{n,\theta_{n,j}^{s}})\\ \notag
&=\hat{p}_n(X_n)\text{exp} \sum_{s=1}^{N_s}\sum_{j=1}^{M_n^{s}}\sum_{k=0}^{K}q_n(\theta_{n,j}^{s}=k)\log \ell({Y}_{n,j}^{s}|X_{n,k})\\\notag
& \propto \hat{p}_n(X_n)\text{exp} \sum_{s=1}^{N_s}\sum_{j=1}^{M_n^{s}}\biggl[\sum_{k=1}^{K}q_n(\theta_{n,j}^{s}=k)\biggl[  -\frac{1}{2}(Y_{n,j}^{s}-HX_{n,k})^\top (R_k^{s})^{-1}(Y_{n,j}^{s}-HX_{n,k}) \biggr]+ q_n(\theta_{n,j}^{s}=0) \log \frac{1}{V^{s}}\biggl]
\\\label{eq:der1}
    &\propto  \hat{p}_n(X_n)\text{exp} \sum_{k=1}^{K} \sum_{s=1}^{N_s}\sum_{j=1}^{M_n^{s}} -\frac{1}{2}(Y_{n,j}^{s}-HX_{n,k})^\top \left(\frac{R_k^{s}}{q_n(\theta_{n,j}^{s}=k)}\right)^{-1}(Y_{n,j}^{s}-HX_{n,k})\\\label{eq:der2}
    &\propto  \hat{p}_n(X_n)\text{exp} \sum_{k=1}^{K} \sum_{s=1}^{N_s} -\frac{1}{2}(\overbar{Y}_n^{k,s}-HX_{n,k})^\top{\overbar{R}_n^{k,s}}^{-1}(\overbar{Y}_n^{k,s}-HX_{n,k})\\\label{eq:der3}
    &\propto  \hat{p}_n(X_n)\text{exp} \sum_{k=1}^{K} -\frac{1}{2}(\overbar{Y}_n^{k}-HX_{n,k})^\top{\overbar{R}_n^{k}}^{-1}(\overbar{Y}_n^{k}-HX_{n,k})
    \\\label{eq:der4}
   & \propto  \hat{p}_n(X_n)\prod_{k=1}^{K}\mathcal{N}\left(\overbar{Y}_n^k;HX_{n,k},\overbar{R}^k_n\right),
\end{align}
where the results from lines \eqref{eq:der1} to \eqref{eq:der2}, and from lines \eqref{eq:der2} to \eqref{eq:der3} are computed according to the rule of calculating the
summation of quadratic forms (the Lemma E.1 in Appendix E in \cite{gan2024variationalsuppl}), that is, for symmetric and invertible matrix $C_i\in \mathbb{R}^{D\times D}$, and vectors $x,m_i\in \mathbb{R}^{D\times1}$ ($i=1,2,...,N$), we have 
\begin{align} \label{eq:quadratric sum theorem}
\begin{aligned}
    \sum_{i=1}^N-\frac{1}{2}(x-m_i)^\top& C_i^{-1}(x-m_i)=-\frac{1}{2}(x-\mu)^\top \Sigma^{-1}(x-\mu)+\frac{1}{2}\mu^\top\Sigma^{-1}\mu-\frac{1}{2}\sum_{i=1}^Nm_i^\top C_i^{-1} m_i,\\
    \Sigma&=\left(\sum_{i=1}^N C_i^{-1}\right)^{-1}, \ \ \ \ \ \ \mu=\left(\sum_{i=1}^N C_i^{-1}\right)^{-1}\sum_{i=1}^N C_i^{-1} m_i.
\end{aligned}
\end{align}
The pseudo-measurements and covariances in \eqref{eq:der2}-\eqref{eq:der4} are computed using the above quadratic summation result as follows:
\begin{align}
   & \overbar{R}_n^{k,s}=\frac{R_k^{s}}{\sum_{j=1}^{M_n^{s}}q_n(\theta_{n,j}^{s}=k)},\quad\quad\quad
\overbar{Y}_n^{k,s}=\frac{\sum_{j=1}^{M_n^{s}}Y_{n,j}^{s}q_n(\theta_{n,j}^{s}=k)}{\sum_{j=1}^{M_n^{s}}q_n(\theta_{n,j}^{s}=k)}.
\end{align}

\begin{align}\overbar{R}_n^{k}=&\left(\sum_{i=1}^{N_s} (\overbar{R}_n^{k,s})^{-1}\right)^{-1}=\left(\sum_{s=1}^{N_s} \left((R_k^{s})^{-1}\sum_{j=1}^{M_n^{s}}q_n(\theta_{n,j}^{s}=k) \right) \right)^{-1}, \\
\overbar{Y}_n^{k}=&\left(\sum_{i=1}^{N_s} (\overbar{R}_n^{k,s})^{-1}\right)^{-1}\sum_{i=1}^{N_s} (\overbar{R}_n^{k,s})^{-1} \overbar{Y}_n^{k,s}=\overbar{R}_n^k \sum_{s=1}^{N_s} \Big((R_k^{s})^{-1}\sum_{j=1}^{M_n^{s}}q_n(\theta_{n,j}^{s}=k)Y_{n,j}^{s}\Big).
\end{align}

\subsection{Update for $q_n(\theta_n)$}
Next, we present the derivation for $q_n(\theta_n)$. According to the standard CAVI update rule in \cite{blei2017variational} and expression of $\hat{p}_n(X_n,\theta_n, Y_n)$ in \eqref{eq: pn}, we have:
\begin{align}\notag
     q_n(\theta_n) \propto&\text{exp}\left(\E_{q_n(X_n)}\log \hat{p}_n(X_n,\theta_n,Y_n)\right)\\ \notag
    \propto& \text{exp}\left(\E_{q_n(X_n)}\log p(\theta_n|M_n)p(Y_n|\theta_n,X_n) \right)\\ \notag
=&\prod_{s=1}^{N_s} \prod_{j=1}^{M_n^{s}}\text{exp}\left(\E_{q_n(X_n)}\log p(\theta_{n,j}^{s})p(Y_{n,j}^{s}|X_{n,\theta_{n,j}^{s}})\right)\\ \label{eq:vari theta independent}  \propto&\prod_{s=1}^{N_s}\prod_{j=1}^{M_n}q_n(\theta_{n,j}^{s})
\end{align}
In the following, we present detailed derivations for $q_n(\theta_{n,j}^{s})$, using expressions of $p(\theta_{n,j}^{s})$ in \eqref{eq:single assoc prior} and $\ell({Y}_{n,j}^{s}|X_{n,k})$ in \eqref{measurement model}:
\begin{align}    \notag&q_n(\theta_{n,j}^{s})=\exp\left(\E_{q_n(X_n)}\log p(\theta_{n,j}^{s})p(Y_{n,j}^{s}|X_{n,\theta_{n,j}^{s}})\right)\\ \notag
&\propto \exp \left( \E_{q_n(X_n)} \log \sum_{k=0}^K\Lambda_k^{s}\ell^{s}(Y_{n,j}^{s}|X_{n,k})\delta[\theta_{n,j}^{s}=k]\right)\\ \notag
&=  \exp \left( \E_{q_n(X_n)}  \sum_{k=0}^K\log \Lambda_k^{s}\ell^{s}(Y_{n,j}^{s}|X_{n,k})\delta[\theta_{n,j}^{s}=k]\right)\\ \notag
& = \frac{\Lambda_0^{s}}{V^{s}}\delta[\theta_{n,j}^{s}=0]+  \exp \left( \E_{q_n(X_n)}\sum_{k=1}^K\log \Lambda_k^{s}\mathcal{N}(Y_{n,j}^{s};HX_{n,k},R_k^{s})\delta[\theta_{n,j}^{s}=k]\right)\\ \notag
& = \frac{\Lambda_0^{s}}{V^{s}}\delta[\theta_{n,j}^{s}=0]+  \exp  \sum_{k=1}^K \bigg[\log \Lambda_k^{s}+\E_{q_n(X_n)}\log\mathcal{N}(Y_{n,j}^{s};HX_{n,k},R_k^{s})\delta[\theta_{n,j}^{s}=k]\bigg]
\\ \notag
& = \frac{\Lambda_0^{s}}{V^{s}}\delta[\theta_{n,j}^{s}=0]\\ \notag
&\quad +  \exp  \sum_{k=1}^K \bigg[\log \Lambda_k^{s}+\left( - \frac{1}{2}\E_{q_n(X_{n,k})}(Y_{n,j}^{s}-HX_{n,k})^\top (R_k^{s})^{-1}(Y_{n,j}^{s}-HX_{n,k})+\log \frac{1}{\sqrt{(2\pi)^D{\det R_k^{s}}}}\right)\delta[\theta_{n,j}^{s}=k]\bigg]\
\\ \notag
& = \frac{\Lambda_0^{s}}{V^{s}}\delta[\theta_{n,j}^{s}=0]+  \exp  \sum_{k=1}^K \bigg[\log \Lambda_k^{s}\\ \notag
&\quad +  \bigg(- \frac{1}{2}\left((Y_{n,j}^{s}-H\mu_{n|n}^k)^\top (R_k^{s})^{-1}(Y_{n,j}^{s}-H\mu_{n|n}^k)+\Tr((R_k^{s})^{-1}H\Sigma_{n|n}^kH^\top) \right) +\log \frac{1}{\sqrt{(2\pi)^D{\det R_k^{s}}}}\bigg)\delta[\theta_{n,j}^{s}=k]\bigg]\
\\ \notag
& = \frac{\Lambda_0^{s}}{V^{s}}\delta[\theta_{n,j}^{s}=0]+  \exp  \sum_{k=1}^K \left[\log \Lambda_k^{s}+  \bigg(\log\mathcal{N}(Y_{n,j}^{s};H\mu_{n|n}^k,R_k^{s})-0.5\Tr((R_k^{s})^{-1}H\Sigma_{n|n}^kH^\top)\bigg)\delta[\theta_{n,j}^{s}=k]\right]\
\\ 
&=\frac{\Lambda_0^{s}}{V^{s}}\delta[\theta_{n,j}^{s}=0]+\sum_{k=1}^K\Lambda_k^{s} l_k^{s}\delta[\theta_{n,j}^{s}=k]\\ 
&l_k^{s}=\mathcal{N}(Y_{n,j}^{s};H\mu_{n|n}^k,R_k^{s})\text{exp}(-0.5\text{Tr}({(R_k^{s})}^{-1}H\Sigma_{n|n}^k H^\top))
\end{align}
where the second line to third line follows from the fact that only one of $\delta[\theta_{n,j}=k]$ for $k=0,1,...,K$ equals $1$, with the rest being zero.

% \section*{Appendix I}
\section{Supplementary Properties and Proofs of LM-ELBO} \label{apx: apx for LM-ELBO properties}
%In this section, we develop the newly-proposed multi-sensor multi-object trackers based on the decentralised (natural) gradient descent variational inference in Section \ref{sec:Decentralised (Natural) Gradient Descent Variational inference for Maximising LM-ELBO}, and provide detailed derivations and implementation steps.

\subsection{Alternative proof of Property 2}\label{Alternative proof of Property 2.1}
Here we give a proof of Property 2 using the ELBO definition in \eqref{eq: ELBO function} and $q(\theta;\rho^*(\lambda))=q^*(\theta)$.
First, recall from Section \ref{sec: LM-ELBO def and assump} that the parametric form $q(\theta;\rho)$, as adopted in the ELBO in \eqref{eq: ELBO function}, encompasses the optimal distribution $q^*(\theta)$ in \eqref{eq: our LM-ELBO CAVI local optimum}, i.e.,
\begin{align} \notag
    q^*(\theta)\propto& \exp\left(\E_{q(X;\lambda)} \log f(X,\theta,Y)\right). 
\end{align}
Additionally, in Section \ref{sec: LM-ELBO def and assump}, $\rho^*(\lambda)$ is dented as one parameter value % (or one among several)
that recovers the $q^*(\theta)$ such that 
\begin{align} \label{eq: fixed form for free form update}
    q(\theta;\rho^*)=q^*(\theta)=\frac{\exp\left(\E_{q(X;\lambda)} \log f(X,\theta,Y)\right)}{Z(\lambda)}
\end{align}
where $Z(\lambda)$ is the normalisation constant that does not depend on $\theta$ or $X$. 

Using \eqref{eq: ELBO function}, we have
\begin{align}\label{eq: gradient expansion}
    \nabla_{\rho} \mathcal{F}(\lambda,\rho)=&\nabla_{\rho} \E_{q(X;\lambda)q(\theta;\rho)}\log f(X,\theta,Y)-\nabla_{\rho} \E_{q(\theta;\rho)}\log q(\theta;\rho),
    % \cancelto{0}{\nabla_{\rho} \E_{q(X;\lambda)}\log q(X;\lambda)}
\end{align}
as $\nabla_{\rho} \E_{q(X;\lambda)}\log q(X;\lambda)=0$. The second term in \eqref{eq: gradient expansion} can be further simplified as 
\begin{align} \notag
    \nabla_{\rho} \E_{q(\theta;\rho)}\log q(\theta;\rho)=&\int \nabla_{\rho} \left(q(\theta;\rho)\log q(\theta;\rho)\right) d\theta \\[-3pt] \notag
    =& \int \left(\nabla_{\rho} q(\theta;\rho)\log q(\theta;\rho) + \nabla_{\rho} q(\theta;\rho)  \right)d\theta\\[-4pt] \label{eq:gradient second term}
    =& \int \nabla_{\rho} q(\theta;\rho)\log q(\theta;\rho)d\theta + \cancelto{0}{\nabla_{\rho} 1}.
\end{align}
The first term in \eqref{eq: gradient expansion} is
\begin{align} \notag 
    \nabla_{\rho} \E_{q(X;\lambda)q(\theta;\rho)}\log f(X,\theta,Y)
    =&\int \nabla_{\rho} q(\theta;\rho) \E_{q(X;\lambda)}\log f(X,\theta,Y)d\theta\\[-3pt] \notag
    =& \int \nabla_{\rho} q(\theta;\rho) (\log q(\theta;\rho^*) +\log Z(\lambda))d\theta\\[-4pt] \label{eq:gradient first term}
    =&  \int \nabla_{\rho} q(\theta;\rho) \log q(\theta;\rho^*)d\theta + \cancelto{0}{\nabla_{\rho}\log Z(\lambda)},
\end{align}
where the second last line is obtained using \eqref{eq: fixed form for free form update}. Subtracting \eqref{eq:gradient second term} from \eqref{eq:gradient first term} yields the gradient in \eqref{eq: gradient expansion}: 
\begin{align} \notag
    \nabla_{\rho} \mathcal{F}(\lambda,\rho)=  \int \nabla_{\rho} q(\theta;\rho)( \log q(\theta;\rho^*)-\log q(\theta;\rho))d\theta.
\end{align}
Finally, we conclude the proof as follows
\begin{align} \label{eq: gradient is zero}
    \nabla_{\rho} \mathcal{F}(\lambda,\rho)|_{\rho=\rho^*}= \int \nabla_{\rho} q(\theta;\rho^*)|_{\rho=\rho^*}\times 0 \ d\theta=0.
\end{align}

\subsection{Proof and analysis of Property 5}\label{Proof and analysis of Property 5}
Here we verify this property on a case-by-case basis. The global maximum case is straightforward: If $\lambda^*$ is a global maximum of $\mathcal{L}(\lambda)$, then $\mathcal{L}(\lambda=\lambda^*)=\max_\lambda \mathcal{L}(\lambda)$. Substituting $\mathcal{L}$ on the left and right hand sides with \eqref{eq: first property}
and \eqref{eq: equivalent definition}, respectively, yields $\mathcal{F}(\lambda=\lambda^*,\rho=\rho^*(\lambda^*))=\max_\lambda\max_\rho\mathcal{F}(\lambda,\rho)$, confirming the global maximum of $\mathcal{F}$. The local maximum case also follows from \eqref{eq: first property}
and \eqref{eq: equivalent definition}. 
%A formal proof will be provided in our full paper. 
Intuitively, if $\mathcal{L}(\eta=\eta^*)$ is maximal in a small neighbourhood of $\lambda^*$, then $\mathcal{F}(\lambda=\lambda^*,\rho=\rho^*(\eta^*))$ achieves the maximum of $\mathcal{F}$ for the corresponding vicinity of $\lambda^*$ across all $\rho$, and consequently, in a small neighbourhood of $[\lambda^*,\rho^*(\lambda^*)]$, validating the local maximum. 
Finally, the stationary point case is confirmed by noting that $\nabla_\lambda\mathcal{L}(\lambda)|_{\lambda=\lambda^*}=0$ leads to $\nabla_{\lambda} \mathcal{F}(\lambda,\rho)|^{\lambda=\lambda^*}_{\rho=\rho^*(\lambda^*)}=0 $, as per \eqref{eq: derivative simplification}. Further, \eqref{eq: zero derivative property} ensures that $\nabla_{\rho} \mathcal{F}(\lambda,\rho)|_{\rho=\rho^*(\lambda)}=0$ for all $\lambda$, including $\lambda^*$. Therefore, both $\nabla_{\lambda} \mathcal{F}(\lambda,\rho)$ and $\nabla_{\rho} \mathcal{F}(\lambda,\rho)$ are zero at $[\lambda^*,\rho^*(\lambda^*)]$, verifying $\mathcal{F}$'s stationary point.

This optimality alignment property demonstrates that any optimum found by optimising $\mathcal{L}(\lambda)$ is inherently an optimum within the conventional ELBO $\mathcal{F}(\lambda,\rho)$, thereby validating the optimisation of our LM-ELBO. We further note that this optimality property directly suggests that a distinctive optimal point $\lambda^*$ of $\mathcal{L}(\lambda)$ results in a distinctive optimal point $[\lambda^*,\rho^*(\lambda^*)]$ of $\mathcal{F}$, ensuring our LM-ELBO does not introduce extra suboptimal points—like local maxima or stationary points—where optimisation algorithms could potentially stagnate, and may even mitigate such risks.

\subsection{An additional property of LM-ELBO}\label{An additional property of LM-ELBO}
 One previously established property in \cite{hensman2012fast} for the KLC bound suggests:
\begin{align}  \label{eq: inequality property}
    \mathcal{F}(\lambda,\rho)\leq\mathcal{L}(\lambda)\leq \log \int f(X,\theta,Y) dXd\theta,  \\[-2em]\notag
\end{align}
indicating that the LM-ELBO provides a tighter bound on the log evidence than the conventional ELBO (assuming $f$ is the joint density).

Here we present a simple proof using the Property 1 in Section \ref{sec: LM-ELBO properties}. First, the left inequality follows directly from \eqref{eq: equivalent definition}. To prove the right inequality, we apply Jensen's inequality to the ELBO definition in \eqref{eq: ELBO function}, yielding:
\begin{align} \label{eq: jensen}
    \mathcal{F}(\lambda,\rho)&\leq 
    \log \E_{q(X;\lambda)q(\theta;\rho)}\left[\frac{f(X, \theta, Y)}{q(X;\lambda)q(\theta;\rho)}\right] 
    = 
    \log \int f(X,\theta,Y) dXd\theta,
\end{align}
where the right hand side equals $\log p(Y)$ if $f$ represents $p(X,\theta,Y)$, highlighting that the ELBO is a lower bound on the log marginal likelihood (regardless of the value of $\rho$). Furthermore, \eqref{eq: first property} suggests that $\mathcal{L}(\lambda)=\mathcal{F}(\lambda,\rho=\rho^*(\lambda))$, and is therefore also bounded by the right-hand side of \eqref{eq: jensen}, thus proving the right inequality in \eqref{eq: inequality property}.

%The second property is that our LM-ELBO $\mathcal{L}(\lambda)$ qualifies as an OLM-ELBO (defined in \cite{hoffman2013stochastic}), when an appropriate form/family for $q(\theta;\rho)$ is assumed as in Section V-B. This is because the Property 1 and Property 2.2 of our LM-ELBO meet the criteria of the OLM-ELBO in \cite{hoffman2013stochastic}.

\subsection{Convergence Assurance for Gradient Hybrid CAVI}\label{app:Convergence Assurance for Gradient Hybrid CAVI}
Using LM-ELBO can establish a convergence assurance for a specific class of CAVI algorithm. Standard CAVI iteratively optimises $q(\theta)$ and $q(X)$ with optimal updates like \eqref{eq: our LM-ELBO CAVI local optimum}. Each update ensures a non-negative increment of ELBO and hence guarantees the convergence. However, if the optimal update for one of the two variational distributions, e.g., the $q(X)$, lacks an analytical solution, an intuitive workaround is implementing one step of the gradient ascent update of $q(X;\lambda)$ (using $\nabla_\lambda\mathcal{F}$) while keeping $q(\theta)$ fixed; and then use the optimal update for $q^*(\theta)$ in \eqref{eq: our LM-ELBO CAVI local optimum} for the next update step. The convergence of such a modified algorithm isn't immediately apparent. Nonetheless, by applying Property 3 in Section \ref{sec: LM-ELBO properties}, we recognise that the algorithm essentially performs successive gradient updates $\nabla_\lambda\mathcal{L}(\lambda)$ for $\lambda$, assuring convergence since $\mathcal{L}(\lambda)$ is a valid objective function and gradient ascent ensures the convergence. This convergence assurance can be extended to other hybride CAVI method using different optimisation technique (e.g., the stochastic and/or natural gradient as proved in \cite{hoffman2013stochastic}), provided that a similar property to Property 3 can be established.

\section{Properties of local LM-ELBO and proofs} \label{apx: properties for local LM-ELBO and proofs}
Here we list 5 properties of local LM-ELBO $\mathcal{L}_s(\lambda_{n})$ (defined in \eqref{eq:local LM ELBO}) as mentioned in Section \ref{sec: local LM-ELBO properties}, along with the corresponding proofs. Recall that in Section \ref{sec: local LM-ELBO properties}, $\rho_n^{s*}(\lambda_n)$ is denoted as the parameter value that reproduces $q_n^*(\theta_n^s)$ in \eqref{eq: local optimal theta} with $\lambda_n$ held fixed, i.e., $q_n^*(\theta_n^s)=q_n(\theta_n^s;\rho_n^{s*}(\lambda_n))$. These 5 properties highlights the relationship between $\mathcal{L}_s(\lambda_{n})$ in \eqref{eq:local LM ELBO} and $\mathcal{F}_s(\lambda_n,\rho_n^s)$ in \eqref{eq: OELBO each sensor}, and they mirror the corresponding five properties in Section \ref{sec: LM-ELBO properties}, with $\lambda, \rho, \rho^*(\lambda)$ replaced by $\lambda_n, \rho_n^s, \rho_n^{s*}(\lambda_n)$, and $\mathcal{F}, \mathcal{L}$ replaced by $\mathcal{F}_s, \mathcal{L}_s$.
\begin{property} \label{prop: basic property 1 apx}
    \begin{align} \label{eq: first property local LM-ELBO apx}
\mathcal{L}_s(\lambda_n)=&\mathcal{F}_s(\lambda_n,\rho_n^s=\rho_n^{s*}(\lambda_n)),\\ \label{eq: equivalent definition local LM-ELBO apx}
    \mathcal{L}_s(\lambda_n)=&\max_{\rho_n^s} \mathcal{F}_s(\lambda_n,\rho_n^s). \\[-2.2em]\notag
\end{align}
\end{property}

\begin{property} \label{prop: property 2 apx}
    \begin{align} 
    \nabla_{\rho_n^s} \mathcal{F}_s(\lambda_n,\rho_n^s)|_{\rho_n^s=\rho_n^{s*}(\lambda_n)}=0. \\[-2em]\notag
\end{align}
\end{property}

\begin{property} \label{prop: property 3 apx}
    \begin{align}
\label{eq: derivative simplification local LM-ELBO apx}
    \nabla_{\lambda_n} \mathcal{L}_s(\lambda_n)=\nabla_{\lambda_n} \mathcal{F}_s(\lambda_n,\rho_n^s)|_{\rho_n^s=\rho_n^{s*}(\lambda_n)}. \\[-2em]\notag
\end{align}
\end{property}

\begin{property} \label{prop: property 4 apx}
    \begin{align} 
    {\nabla}_{\lambda_n}^2 \mathcal{L}_s(\lambda_n) = {\nabla}_{\lambda_n}^2 \mathcal{F}_s(\lambda_n,\rho_n^s)|_{\rho_n^s=\rho_n^{s*}(\lambda_n)}+P,  \\[-2em]\notag
\end{align}
where $P$ is a positive semi-definite matrix.
\end{property}

\begin{property} \label{prop: property 5 apx}
    If $\lambda_n^*$ is a global maximiser, a local maximiser, or a stationary point of $\mathcal{L}_s(\lambda_n)$, then $[\lambda_n^*,\rho_n^{s*}(\lambda_n^*)]$ is, respectively, a global maximiser, a local maximiser, or a stationary point of $\mathcal{F}_s(\lambda_n,\rho_n^s)$.
\end{property}

We now present the proof for these five properties. It is sufficient to prove property \ref{prop: basic property 1 apx} (i.e., the \eqref{eq: first property local LM-ELBO apx} and \eqref{eq: equivalent definition local LM-ELBO apx}), as properties \ref{prop: property 2 apx}-\ref{prop: property 5 apx} can all be derived from property \ref{prop: basic property 1 apx} by following the same steps outlined in Section \ref{sec: LM-ELBO properties} for the corresponding properties. Therefore, we refer to property \ref{prop: basic property 1 apx} as the fundamental property.

We now prove the fundamental property \ref{prop: basic property 1 apx}, starting with \eqref{eq: first property local LM-ELBO apx}. By comparing the definitions of $\mathcal{F}_s(\lambda_n,\rho_n^s)$ in \eqref{eq: OELBO each sensor}
and $\mathcal{L}_s(\lambda_{n})$ in \eqref{eq:local LM ELBO}, we observe that if $q_n^*(\theta_n^{s})=q_n(\theta_n^{s};\rho_n^{s})$, then $\mathcal{F}_s(\lambda_n,\rho_n^s)=\mathcal{L}_s(\lambda_{n})$. Moreover, by the definition of $\rho_n^{s*}(\lambda_n)$ in Section \ref{sec: local LM-ELBO properties}, we have $q_n^*(\theta_n^s)=q_n(\theta_n^s;\rho_n^{s*}(\lambda_n))$. Therefore, setting $\rho_n^s=\rho_n^{s*}(\lambda_n)$ gives $\mathcal{F}_s(\lambda_n,\rho_n^s)=\mathcal{L}_s(\lambda_{n})$, i.e., $\mathcal{L}_s(\lambda_n)=\mathcal{F}_s(\lambda_n,\rho_n^s=\rho_n^{s*}(\lambda_n))$, which proves \eqref{eq: first property local LM-ELBO apx}.

Next, we prove \eqref{eq: equivalent definition local LM-ELBO apx}. Since \eqref{eq: first property local LM-ELBO apx} is already established, proving \eqref{eq: equivalent definition local LM-ELBO apx} is equivalent to proving: 
\begin{align} \label{eq: sub ELBO maximum equivalence}
    \mathcal{F}_s(\lambda_n,\rho_n^s=\rho_n^{s*}(\lambda_n))=\max_{\rho_n^s} \mathcal{F}_s(\lambda_n,\rho_n^s).
\end{align}
To prove \eqref{eq: sub ELBO maximum equivalence}, we introduce the following Lemma:

\begin{lemma} \label{lemma: optimum star equivalence}
Recall the assumption from Section \ref{sec: LM-ELBO over O-ELBO}, where $q_n(\theta_n; \rho_n) = \prod_{s=1}^{N_s} q_n(\theta_n^s; \rho_n^s)$ and $\rho_n = [\rho_n^1, \rho_n^2, \dots, \rho_n^{N_s}]$. Let $\rho_n^{*}(\lambda_n)$ be the parameter value of $q_n(\theta_n;\rho_n)$ that yields the optimal distribution $q_n^*(\theta_n)$ in \eqref{eq: local optimal theta} with $\lambda_n$ held fixed, i.e., 
\begin{align} \label{eq: tracking optimal rho for theta}
    q_n(\theta_n;\rho_n^{*}(\lambda_n))=q_n^*(\theta_n),
\end{align}
then we have $$\rho_n^{*}(\lambda_n)=[\rho_n^{1*}(\lambda_n),\rho_n^{2*}(\lambda_n),...,\rho_n^{N_s*}(\lambda_n)],$$
where $\rho_n^{s*}(\lambda_n)$ $(s=1,2,...,N_s)$ is defined in Section \ref{sec: local LM-ELBO properties} as the parameter value that reproduces $q_n^*(\theta_n^s)$ in \eqref{eq: local optimal theta} with $\lambda_n$ held fixed, i.e., $q_n^*(\theta_n^s)=q_n(\theta_n^s;\rho_n^{s*}(\lambda_n))$.
\end{lemma}

\begin{proof}
Let $\rho_n^{s,o}(\lambda_n)$ denote the value of $\rho_n^s$ for $s = 1, 2, \dots, N_s$ when $\rho_n = [\rho_n^1, \rho_n^2, \dots, \rho_n^{N_s}]$ takes the value $\rho_n^{*}(\lambda_n)$. That is, $\rho_n^{*}(\lambda_n) = [\rho_n^{1,o}(\lambda_n), \rho_n^{2,o}(\lambda_n), \dots, \rho_n^{N_s,o}(\lambda_n)]$. Then, we have $q_n(\theta_n;\rho_n^{*}(\lambda_n))=\prod_{s=1}^{N_s}q_n(\theta_n^s;\rho_n^{s,o}(\lambda_n))$. Moreover, from \eqref{eq: tracking optimal rho for theta} and the fact that $q_n^*(\theta_n) = \prod_{s=1}^{N_s} q_n^*(\theta_n^s)$ as stated in \eqref{eq: local optimal theta}, we know that $q_n^*(\theta_n;\rho_n^{*}(\lambda_n))=\prod_{s=1}^{N_s}q_n^*(\theta_n^s)$. Therefore, we have $\prod_{s=1}^{N_s}q_n(\theta_n^s;\rho_n^{s,o}(\lambda_n))=\prod_{s=1}^{N_s}q_n^*(\theta_n^s)$. Subsequently, by marginalising $\theta_n^{s-}$ from both sides, it follows that $q_n(\theta_n^s;\rho_n^{s,o}(\lambda_n))=q_n^*(\theta_n^s)$ for each $s=1,2,...,N_s$. This implies that each $\rho_n^{s,o}(\lambda_n)$ is also the optimal parameter value that reproduces $q_n^*(\theta_n^s)$ with $\lambda_n$ held fixed. This matches the definition of $\rho_n^{s*}(\lambda_n)$ in Section \ref{sec: local LM-ELBO properties}, where $q_n^*(\theta_n^s) = q_n(\theta_n^s; \rho_n^{s*}(\lambda_n))$. Therefore, we conclude that $\rho_n^{s,o}(\lambda_n)=\rho_n^{s*}(\lambda_n)$ for $s = 1, 2, \dots, N_s$, and thus $\rho_n^{*}(\lambda_n)=[\rho_n^{1*}(\lambda_n),\rho_n^{2*}(\lambda_n),...,\rho_n^{N_s*}(\lambda_n)]$.
    
\end{proof}

We now prove \eqref{eq: sub ELBO maximum equivalence}. As mentioned in Section \ref{sec: local LM-ELBO properties}, the LM-ELBO $\mathcal{L}(\lambda_n)$ naturally possesses the properties described in Section \ref{sec: LM-ELBO properties} owing to its derivation, where Property 1 states that
\begin{align} \label{eq: tracking LM-ELBO property 1}
    \mathcal{F}(\lambda_n,\rho_n=\rho_n^{*}(\lambda_n))=\max_{\rho_n} \mathcal{F}(\lambda_n,\rho_n),
\end{align}
where $\rho_n^{*}(\lambda_n)$ denotes the parameter value that yields the optimal distribution $q_n^*(\theta_n)$, as defined in Lemma \ref{lemma: optimum star equivalence}. Using Lemma \ref{lemma: optimum star equivalence}, we have $\mathcal{F}(\lambda_n,\rho_n=\rho_n^{*}(\lambda_n))=\mathcal{F}(\lambda_n,\rho_n=[\rho_n^{1*}(\lambda_n),\rho_n^{2*}(\lambda_n),...,\rho_n^{N_s*}(\lambda_n)])$ where $\rho_n^{s*}(\lambda_n)$ $(s = 1, 2, \dots, N_s)$ is the optimal parameter value of $q_n(\theta_n^s; \rho_n^s)$ such that $q_n^*(\theta_n^s) = q_n(\theta_n^s; \rho_n^{s*}(\lambda_n))$, as defined in Section \ref{sec: local LM-ELBO properties}. Furthermore, using the relation $\mathcal{F}(\lambda_n, \rho_n) = \sum_{s=1}^{N_s} \mathcal{F}_s(\lambda_n, \rho_n^s)$ ($\rho_n = [\rho_n^1, \rho_n^2, \dots, \rho_n^{N_s}]$) from the expression above \eqref{eq: OELBO each sensor}, we obtain:
\begin{align} \notag
    \mathcal{F}(\lambda_n,\rho_n=&\rho_n^{*}(\lambda_n))=\mathcal{F}(\lambda_n,\rho_n=[\rho_n^{1*}(\lambda_n),\rho_n^{2*}(\lambda_n),...,\rho_n^{N_s*}(\lambda_n)])\\ \label{eq: ELBO decomposed with local optimal rho}
    =&\sum_{s=1}^{N_s} \mathcal{F}_s(\lambda_n,\rho_n^s=\rho_n^{s*}(\lambda_n)).
\end{align}
Additionally, using $\mathcal{F}(\lambda_n,\rho_n)= \sum_{s=1}^{N_s} \mathcal{F}_s(\lambda_n,\rho_n^s)$ with $\rho_n=[\rho_n^1,\rho_n^2,...,\rho_n^{N_s}]$ again, we have
\begin{align} \notag
    \max_{\rho_n} \mathcal{F}(\lambda_n,\rho_n)=&\max_{\rho_n^1,\rho_n^2,...,\rho_n^{N_s}} \sum_{s=1}^{N_s} \mathcal{F}_s(\lambda_n,\rho_n^s)\\ \label{eq: sum maximum}
    =& \sum_{s=1}^{N_s} \max_{\rho_n^s} \mathcal{F}_s(\lambda_n,\rho_n^s).
\end{align}
Next, by combining \eqref{eq: tracking LM-ELBO property 1} and \eqref{eq: ELBO decomposed with local optimal rho}, we have 
\begin{align} \notag
    \max_{\rho_n} \mathcal{F}(\lambda_n,\rho_n)=&\mathcal{F}(\lambda_n,\rho_n=\rho_n^{*}(\lambda_n))\\ \notag
    =&\sum_{s=1}^{N_s} \mathcal{F}_s(\lambda_n,\rho_n^s=\rho_n^{s*}(\lambda_n))\\ \label{eq: maximum inequality}
    \leq&\sum_{s=1}^{N_s} \max_{\rho_n^s} \mathcal{F}_s(\lambda_n,\rho_n^s).
\end{align}
The equality in \eqref{eq: maximum inequality} holds if and only if $\mathcal{F}_s(\lambda_n,\rho_n^s=\rho_n^{s*}(\lambda_n))=\max_{\rho_n^s} \mathcal{F}_s(\lambda_n,\rho_n^s)$ for all $s = 1, 2, \dots, N_s$. Since \eqref{eq: sum maximum} implies that this equality holds, we conclude that $\mathcal{F}_s(\lambda_n,\rho_n^s=\rho_n^{s*}(\lambda_n))=\max_{\rho_n^s} \mathcal{F}_s(\lambda_n,\rho_n^s)$ for all $s=1,2,...,N_s$, thus proving \eqref{eq: sub ELBO maximum equivalence}.

Since we have proven both \eqref{eq: first property local LM-ELBO apx} and \eqref{eq: sub ELBO maximum equivalence}, it follows that \eqref{eq: equivalent definition local LM-ELBO apx} is also true, thus completing the proof of the fundamental property \ref{prop: basic property 1 apx} (i.e., \eqref{eq: first property local LM-ELBO apx} and \eqref{eq: equivalent definition local LM-ELBO apx}) for $\mathcal{L}_s(\lambda_n)$ and $\mathcal{F}_s(\lambda_n,\rho_n^s)$. Recall that properties \ref{prop: property 2 apx}–\ref{prop: property 5 apx} can all be derived from the fundamental property \ref{prop: basic property 1 apx} by following the same steps as in Section \ref{sec: LM-ELBO properties} for their corresponding properties. Therefore, we conclude that all properties \ref{prop: basic property 1 apx}–\ref{prop: property 5 apx} are valid.

\section{Derivation of the gradient} \label{apx: gradient derivation}
In this appendix, we derive the gradient ${\nabla}_{\lambda_{n}^{s}} \mathcal{L}_s(\lambda_{n}^{s})$ as presented in Section \ref{sec: standard gradient derivation}. Using the property in \eqref{simplify the gradient computation}, we have $\nabla_{\lambda_{n}^{s}} \mathcal{L}_s(\lambda_{n}^{s})\!=\!\nabla_{\lambda_{n}^{s}} \mathcal{F}_s(\lambda_n^{s},\rho_n^{s})|_{\rho_n^{s}=\rho_n^{s*}(\lambda_n^s)}$. To compute this, we first take the partial derivative of $\mathcal{F}_s(\lambda_n^{s},\rho_n^{s})$ with respect to $\lambda_n^{s}$, and then substitute $\rho_n^{s}$ with $\rho_n^{s*}(\lambda_n^s)$. More compactly, we can express this as $\nabla_{\lambda_{n}^{s}} \mathcal{L}_s(\lambda_{n}^{s})\!=\!\nabla_{\lambda_{n}^{s}} \mathcal{F}_s(\lambda_n^{s},\rho_n^{s*}(\lambda_n^s))$, where $\rho_n^{s*}(\lambda_n^s)$ is treated as independent of $\lambda_n^s$ during the gradient evaluation. We now evaluate this gradient $\nabla_{\lambda_{n}^{s}} \mathcal{F}_s(\lambda_n^{s},\rho_n^{s*}(\lambda_n^s))$. Using \eqref{eq: OELBO each sensor} and the fact that $q_n^{s,*}(\theta_n^{s})\!=\!q_n(\theta_n^s;\rho_n^{s*}(\lambda_n^s))$ (as stated immediately prior to Section \ref{sec: decentralised gradient trackers}), we have

\begin{align}    \label{eq: apx OELBO each sensor} 
     \mathcal{F}_s(\lambda_n^{s},\rho_n^{s*}(\lambda_n^s))
     &= \E_{q_n(X_n; \lambda_{n}^s)q_n(\theta_n^s;\rho_n^{s*}(\lambda_n^s))}\log p(Y_{n}^{s}|\theta_{n}^{s},X_{n})    + \E_{q_n(\theta_n^s;\rho_n^{s*}(\lambda_n^s))}\log  \frac{p(\theta_{n}^{s}|M_n^{s})}{q_n(\theta_n^s;\rho_n^{s*}(\lambda_n^s))} \\\notag
    &\quad + \frac{1}{N_s}\E_{q_n(X_n; \lambda_{n}^s)}\log\hat{p}_n(X_n) - \frac{1}{N_s}\E_{q_n(X_n; \lambda_{n}^s)}\log q_n(X_n;\lambda_{n}^s)\\
    &= \E_{q_n(X_n; \lambda_{n}^s)q_n^{s,*}(\theta_n^{s})}\log p(Y_{n}^{s}|\theta_{n}^{s},X_{n})    + \E_{q_n^{s,*}(\theta_n^{s})}\log  \frac{p(\theta_{n}^{s}|M_n^{s})}{q_n^{s,*}(\theta_n^{s})} - \frac{1}{N_s}\text{KL}(q_n(X_n; \lambda_{n}^s)||\hat{p}_n(X_n)),
\end{align}
where the KL divergence is defined as $\text{KL}(q_n(X_n; \lambda_{n}^s)||\hat{p}_n(X_n))=\E_{q_n(X_n; \lambda_{n}^s)} \log \frac{q_n(X_n; \lambda_{n}^s)}{\hat{p}_n(X_n)} $. To compute ${\nabla}_{\lambda_{n}^{s}} \mathcal{L}_s(\lambda_{n}^{s})=\nabla_{\lambda_{n}^{s}} \mathcal{F}_s(\lambda_n^{s},\rho_n^{s*}(\lambda_n^s))$, where $\rho_n^{s*}(\lambda_n^s)$ (and thus $q_n^{s,*}(\theta_n^{s})=q_n(\theta_n^s;\rho_n^{s*}(\lambda_n^s))$) are treated as independent of $\lambda_n^s$ during the gradient evaluation, we have 
\begin{align} 
    {\nabla}_{\lambda_{n}^{s}} \mathcal{L}_s(\lambda_{n}^{s})=\nabla_{\lambda_{n}^{s}} \mathcal{F}_s(\lambda_n^{s},\rho_n^{s*}(\lambda_n^s))=& - {\nabla}_{\lambda_{n}^{s}}\frac{1}{N_s}\text{KL}(q_n(X_n; \lambda_{n}^s)||\hat{p}_n(X_n))+{\nabla}_{\lambda_{n}^{s}}\E_{q_n(X_n; \lambda_{n}^s)q_n^{s,*}(\theta_n^{s})}\log p(Y_{n}^{s}|\theta_{n}^{s},X_{n})\\ \label{eq: apx L=L1+L2}
    =&{\nabla}_{\lambda_{n}^{s}} \mathcal{L}_s^1(\lambda_n^s)+{\nabla}_{\lambda_{n}^{s}} \mathcal{L}_s^2(\lambda_n^s),
\end{align}
where $\mathcal{L}_s^1(\lambda_n^s),\mathcal{L}_s^2(\lambda_n^s)$ are defined as
\begin{align} \label{eq: apx L1}
\mathcal{L}_s^1(\lambda_n^s)=&- \frac{1}{N_s}\text{KL}(q_n(X_n; \lambda_{n}^s)||\hat{p}_n(X_n))=\frac{1}{N_s} \E_{q_n(X_n; \lambda_{n}^s)}\log \frac{\hat{p}_n(X_n)}{q_n(X_n; \lambda_{n}^s)},\\ \label{eq: apx L2}
\mathcal{L}_s^2(\lambda_n^s)=&\E_{q_n(X_n; \lambda_{n}^s)q_n^{s,*}(\theta_n^{s})}\log p(Y_{n}^{s}|\theta_{n}^{s},X_{n}),
\end{align}
and we note that $q_n^{s,*}(\theta_n^{s})$ is treated as independent of $\lambda_n^s$ during the gradient evaluation.

Recall from Section \ref{sec:Prediction and update steps} and the opening paragraph of Section \ref{sec: decentralised gradient trackers} that at sensor node $s$, the predictive prior is $\hat{p}(X_n)=\prod_{k=1}^K\hat{p}(X_{n,k})$ with $\hat{p}_n(X_{n,k})=\mathcal{N}(X_{n,k};\mu^{k*,s}_{n|n-1},\Sigma^{k*,s}_{n|n-1})$, and $q_n(X_n;\!\lambda_{n}^{s})\!=\!\prod_{k=1}^K \!q_n(X_{n,k};\!\lambda_{n,k}^s)$ with
$q_n(X_{n,k};\lambda_{n,k}^s)\!=\!\mathcal{N}(X_{n,k};\mu^{k,s}_{n|n},\Sigma^{k,s}_{n|n})$. Then, using the multivariate Gaussian KL divergence formula, the $\mathcal{L}_s^1(\lambda_n^s)$ in \eqref{eq: apx L1} is
\begin{align} \notag
    &\mathcal{L}_s^1(\lambda_n^s)=- \frac{1}{N_s}\text{KL}(q_n(X_n; \lambda_{n}^s)||\hat{p}_n(X_n))=- \frac{1}{N_s}\sum_{k=1}^K \text{KL}(q_n(X_{n,k}; \lambda_{n,k}^s)||\hat{p}_n(X_{n,k}))\\ \label{eq: apx L1 derived}
    =& \frac{-1}{2N_s}\sum_{k=1}^K \left [\mathrm{Tr}\left((\Sigma^{k*,s}_{n|n-1})^{-1}\Sigma^{k,s}_{n|n}\right) + (\mu^{k,s}_{n|n} - \mu^{k*,s}_{n|n-1})^\top (\Sigma^{k*,s}_{n|n-1})^{-1} (\mu^{k,s}_{n|n} - \mu^{k*,s}_{n|n-1})-\log|\Sigma^{k,s}_{n|n}|+\log|\Sigma^{k*,s}_{n|n-1}|-d \right]
\end{align}

% Then, the expectations in \eqref{eq: apx OELBO each sensor} can be computed as follows
% \begin{align} \notag
% &\E_{q_n(X_{n,k})}\log\hat{p}_n(X_{n,k})\\
% &=-\frac{1}{2} \left [ \log|\Sigma^{k*,s}_{n|n-1}|+d \log 2\pi + \E_{q_n(X_{n,k})} (X_{n,k} - \mu^{k*,s}_{n|n-1})^\top (\Sigma^{k*,s}_{n|n-1})^{-1} (X_{n,k} - \mu^{k*,s}_{n|n-1}) \right ]\\\notag
%       & =-\frac{1}{2} \left [ \log|\Sigma^{k*,s}_{n|n-1}|+d \log 2\pi+\mathrm{Tr}\left((\Sigma^{k*,s}_{n|n-1})^{-1}\Sigma^{k,s}_{n|n}\right) + (\mu^{k,s}_{n|n} - \mu^{k*,s}_{n|n-1})^\top (\Sigma^{k*,s}_{n|n-1})^{-1} (\mu^{k,s}_{n|n} - \mu^{k*,s}_{n|n-1})  \right ]\\
      % &\E_{q_n(X_{n,k})}\log q_n(X_{n,k})\\ \notag
% &=-\frac{1}{2} \log|\Sigma^{k,s}_{n|n}|-\frac{d}{2} \log 2\pi -\frac{1}{2}\E_{q_n(X_{n,k})} (X_{n,k} - \mu^{k,s}_{n|n})^\top (\Sigma^{k,s}_{n|n})^{-1} (X_{n,k} -\mu^{k,s}_{n|n}) \\\label{eq:expectation 2}
%      & = -\frac{1}{2} \log|\Sigma^{k,s}_{n|n}|-\frac{d}{2} \log 2\pi-\frac{d}{2}
% \end{align}
To compute the $\mathcal{L}_s^2(\lambda_n^s)$ in \eqref{eq: apx L2}, first we derive the inner expectation $\E_{q_n^{s,*}(\theta_n^{s})}\log p(Y_{n}^{s}|\theta_{n}^{s},X_{n})$. The detailed derivation of follows the same steps as in Equations (67)-(70) of Appendix E in \cite{gan2024variationalsuppl}, so we will not repeat it here and instead provide the final form:
\begin{align}
     & \E_{q_n^{s,*}(\theta_n^{s})}\log p(Y_{n}^{s}|\theta_{n}^{s},X_{n}) =\sum_{k=1}^K\log \mathcal{N}(\overbar{Y}_n^{k,s};HX_{n,k},\overbar{R}_n^{k,s}) +C_x^{s} 
\end{align}
where $C_x^{s}$ is a constant that does not depend on $X_{n}$, and pseudo-measurement $\overbar{Y}_n^{k,s}$ and covariance $\overbar{R}_n^{k,s}$ at each sensor $s$ are 
\begin{align} \label{eq: apx pseudo R}
    \overbar{R}_n^{k,s}=&\frac{R_k^s}{\sum_{j=1}^{M_n^{s}}q_n^{s,*}(\theta_{n,j}^{s}=k)},\\ \label{eq: apx pseudo Y}
    \overbar{Y}_n^{k,s}=&\frac{\sum_{j=1}^{M_n^{s}}Y_{n,j}^{s}q_n^{s,*}(\theta_{n,j}^{s}=k)}{\sum_{j=1}^{M_n^{s}}q_n^{s,*}(\theta_{n,j}^{s}=k)}.
\end{align}
Subsequently, the $\mathcal{L}_s^2(\lambda_n^s)$ in \eqref{eq: apx L2} is given by
\begin{align} \notag
\mathcal{L}_s^2(\lambda_n^s)&=\E_{q_n(X_n; \lambda_{n}^s)q_n(\theta_n^{s})}\log p(Y_{n}^{s}|\theta_{n}^{s},X_{n})\\\notag
&=\E_{q_n(X_n; \lambda_{n}^s)} \left[ \sum_{k=1}^K\log \mathcal{N}(\overbar{Y}_n^{k,s};HX_{n,k},\overbar{R}_n^{k,s}) +C_x^s \right] \\ \notag
&=C_x^{s}+ \sum_{k=1}^K \left[ -\frac{1}{2} \log|\overbar{R}_n^{k,s}|-\frac{d}{2} \log 2\pi    -\frac{1}{2} \E_{q_n(X_n; \lambda_{n}^s)}  (\overbar{Y}_n^{k,s} - HX_{n,k})^\top (\overbar{R}_n^{k,s})^{-1} (\overbar{Y}_n^{k,s} - HX_{n,k}) \right] \\ \label{eq: apx L2 derived}
&= -\frac{1}{2} \sum_{k=1}^K  ( H\mu^{k,s}_{n|n}-\overbar{Y}_n^{k,s})^\top (\overbar{R}_n^{k,s})^{-1} (H\mu^{k,s}_{n|n}-\overbar{Y}_n^{k,s})  -\frac{1}{2} \sum_{k=1}^K  \mathrm{Tr}\left(H^\top(\overbar{R}_n^{k,s})^{-1}H\Sigma^{k,s}_{n|n}\right) + C_{xy}^s
\end{align}
where $C_{xy}^s=C_x^{s} -\frac{1}{2} \log \prod_{k=1}^K |\overbar{R}_n^{k,s}| - \frac{dK}{2} \log 2\pi$ is a constant term that does not depend on $\lambda_n^s=[\lambda_{n,1}^s,\lambda_{n,2}^s,...,\lambda_{n,K}^s]$, with $\lambda_{n,k}^s=[\mu^{k,s}_{n|n},\Sigma^{k,s}_{n|n}]$ ($k=1,2,...,K$) as defined in \eqref{eq: lambda define normal gradient}.

Finally, the gradient of the local LM-ELBO ${\nabla}_{\lambda_{n}^{s}} \mathcal{L}_s(\lambda_{n}^{s})=\nabla_{\lambda_{n}^{s}} \mathcal{F}_s(\lambda_n^{s},\rho_n^{s*}(\lambda_n^s))$ can be written as follows using \eqref{eq: apx L=L1+L2}, \eqref{eq: apx L1 derived} and \eqref{eq: apx L2 derived}
\begin{align} \notag
   \nabla_{\lambda_{n}^{s}} \mathcal{L}_s(\lambda_{n}^{s})=&\nabla_{\lambda_{n}^{s}} \mathcal{F}_s(\lambda_n^{s},\rho_n^{s*}(\lambda_n^s))={\nabla}_{\lambda_{n}^{s}} \mathcal{L}_s^1(\lambda_n^s)+{\nabla}_{\lambda_{n}^{s}} \mathcal{L}_s^2(\lambda_n^s) \\ \notag
     =& \frac{1}{2 N_s}\sum_{k=1}^{K}\nabla_{\lambda_{n}^{s}}\left [\log|\Sigma^{k,s}_{n|n}|-\mathrm{Tr}\left((\Sigma^{k*,s}_{n|n-1})^{-1}\Sigma^{k,s}_{n|n}\right) - (\mu^{k,s}_{n|n} - \mu^{k*,s}_{n|n-1})^\top (\Sigma^{k*,s}_{n|n-1})^{-1} (\mu^{k,s}_{n|n} - \mu^{k*,s}_{n|n-1}) \right] \\ \notag
   & -\frac{1}{2} \sum_{k=1}^K \nabla_{\lambda_{n}^{s}}\left [ ( H\mu^{k,s}_{n|n}-\overbar{Y}_n^{k,s})^\top (\overbar{R}_n^{k,s})^{-1} (H\mu^{k,s}_{n|n}-\overbar{Y}_n^{k,s})  + \mathrm{Tr}\left(H^\top(\overbar{R}_n^{k,s})^{-1}H\Sigma^{k,s}_{n|n}\right)\right],
\end{align}
where $\overbar{Y}_n^{k,s}$ and $\overbar{R}_n^{k,s}$ are given in \eqref{eq:pseudomeas Y}, and are treated as independent of $\lambda_n^s$ during gradient evaluation. The gradients ${\nabla}_{\lambda_{n}^{s}} \mathcal{L}_s(\lambda_{n}^{s})$ can then be computed with respect to local estimates of each variational parameter $\mu^{k,s}_{n|n}$ and $\Sigma^{k,s}_{n|n}$, $k=1,...,K$, using the matrix derivative formulas in \cite{petersen2008matrix}, i.e.,
\begin{align}
 \nabla_{\mu^{k,s}_{n|n}} \mathcal{L}_s(\lambda_{n}^{s})&=-\frac{1}{N_s} (\Sigma^{k*,s}_{n|n-1})^{-1}(\mu^{k*,s}_{n|n-1}-\mu^{k,s}_{n|n})+H^\top (\overbar{R}_n^{k,s})^{-1} (\overbar{Y}_n^{k,s}-H \mu^{k,s}_{n|n}) 
\\
\nabla_{\Sigma^{k,s}_{n|n}} \mathcal{L}_s(\lambda_{n}^{s})
&=\frac{1}{2N_s}\left( (\Sigma^{k,s}_{n|n})^{-1}- (\Sigma^{k*,s}_{n|n-1})^{-1} \right)- \frac{1}{2}H^\top(\overbar{R}_n^{k,s})^{-1} H
\end{align}

\section{Derivation of the natural gradient} \label{apx: natural gradient derivation}
\subsection{ The exponential family and some properties}
%In this section, we formulate our task of variational update into the same optimisation problem of maximising the LM-ELBO while rewriting the LM-ELBO using canonical exponential family distributions for the convenience of deriving the natural gradient of variational parameters.
The general form of canonical exponential family distributions can be expressed as follows,
\begin{equation}\label{eq:general form of canonical exponential family}
    q(x ; \lambda) = h(x) \exp \left( \lambda^\top  T(x) - A(\lambda) \right)
\end{equation}
where $x$ is the random variable, $h(x)$ is the base function, $\lambda$ is the natural parameter of the distribution. $T(x)$ is the sufficient statistic, and $A(\lambda)$ is the log partition function that ensures $q(x ; \lambda)$ integrating to $1$. This general form covers a wide range of probability distributions, including the Gaussian, Poisson, and Binomial distributions. Taking the multivariate Gaussian distribution	$\mathcal{N}(x; \mu, \Sigma)$ as an example, the exponential family components defined in \eqref{eq:general form of canonical exponential family} are 
\begin{align}\notag
& h(x)=(2 \pi)^{-\frac{d}{2}},\quad T(x)=\begin{bmatrix} x \\xx^\top\end{bmatrix} \\\notag
&\lambda=\begin{bmatrix}\lambda_{1}\\ \lambda_{2}\end{bmatrix} = \begin{bmatrix}\Sigma^{-1}\mu\\ -\frac{1}{2}\Sigma^{-1}\end{bmatrix}\\\notag
&A(\lambda)=-\frac{1}{4} (\lambda_{1})^\top(\lambda_{2})^{-1}\lambda_{1}-\frac{1}{2} \log |\!-\!2\lambda_{2}|
\end{align}

A useful property is that the expectation of the natural sufficient statistics is the gradient the log-partition function $A(\lambda)$ with respect to the natural parameter $\lambda$:
\begin{equation}\label{eq:the natural sufficient statistics}
\E_{q(x ; \lambda)}[T(x)] = \nabla_{\lambda} A(\lambda)
\end{equation}
Another useful property is that the covariance matrix of the sufficient statistics 
$T(x)$ is the Hessian of the log-partition function $A(\lambda)$ with respect to the natural parameter $\lambda$.
\begin{equation} \label{eq:covariance sufficient statistics}
    \E_{q(x ; \lambda)}\left[\left(T(x)-\E_{q(x ; \lambda)}[T(x)]\right)\left(T(x)-\E_{q(x ; \lambda)}[T(x)]\right)^\top\right] = \nabla_{\lambda}^2 A(\lambda).
\end{equation}
Subsequently, the Fisher information matrix $G(\lambda)$ is also the Hessian of the log-partition function $A(\lambda)$, i.e.,
\begin{align}\notag
     G(\lambda)&= \E_{q(x ; \lambda)}\left[ \left(\nabla_{\lambda}\log{p(X; \lambda)}\right)\left(\nabla_{\lambda}\log{p(X; \lambda)}\right)^\top \right]  \\\notag
     &= \E_{q(x ; \lambda)}\left[ \left( T(x) - \nabla_{\lambda} A(\lambda)\right)\left(T(x) - \nabla_{\lambda} A(\lambda)\right)^{\top} \right]  \\\notag
     &= \E_{q(x ; \lambda)} \left[\left( T(x) - \E_{q(x ; \lambda)}[T(x)]\right)\left(T(x) - \E_{q(x ; \lambda)}[T(x)]\right)^{\top} \right] \\\label{eq:Fisher information matrix1}
     &= \nabla^2_{\lambda} A(\lambda),
\end{align}
where the second last line is obtained by using \eqref{eq:the natural sufficient statistics}, and the last line is obtained by using \eqref{eq:covariance sufficient statistics}.
\subsubsection{Natural gradient and the expectation parameter}
A useful strategy used in this paper to avoid the computation of the inversion of Fisher information matrix is the variable transformation \cite{khan2018fast}. This allows the natural gradient with respect to the natural parameters to be computed via the gradient with respect to the expectation of the sufficient statistics.

Specifically, let the parameter $m$ denote the expectation of the sufficient statistics. Then, \eqref{eq:the natural sufficient statistics} defines a mapping between $\lambda$ and $m$:
\begin{align}\label{eq: apx m is gradient log partition}
    m=\E_{q(x;\lambda)}[T(x)]=\nabla_{\lambda} A(\lambda).
\end{align}
For an exponential family in a \textit{minimal} representation (commonly used and applicable in this paper), there exists a one-to-one mapping between the natural parameter $\lambda$ and the expectation parameter $m$ (see \cite{khan2018fast} for details). Thus one can derive a unique reverse mapping from \eqref{eq: apx m is gradient log partition} and express $f(\lambda)$ in terms of $m$. Subsequently, for a function $f(\lambda)$ of the natural parameter $\lambda$, its gradient $\nabla_{\lambda} f(\lambda)$ can be related to its gradient with respect to the sufficient statistics expectation parameter $m$ as follows:
\begin{align}
   & \nabla_{\lambda} f(\lambda)= (\mathbf{J}_{\lambda}m )\nabla_{m} f(m)=\nabla^2_{\lambda} A(\lambda)\nabla_{m} f(m)=G(\lambda)\nabla_{m} f(m)
\end{align} 
where $f(m)$ is the is the reparameterised form of $f(\lambda)$ using the reverse relationship in \eqref{eq: apx m is gradient log partition}. $\mathbf{J}_{\lambda}m$ is the Jacobian matrix of $m$ with respect to $\lambda$, arising from the application of the chain rule. The last two equalities follow from \eqref{eq: apx m is gradient log partition} and \eqref{eq:Fisher information matrix1}.

Finally, using the definition of the natural gradient
   $\hat{\nabla}_{\lambda} f(\lambda)=G(\lambda)^{-1}\nabla_{\lambda} f(\lambda)$
, we observe an important property: the natural gradient with respect to natural parameter equals to the gradient with respect to the sufficient statistics expectation parameter:
\begin{align} \label{eq: apx natural gradient is expectation gradient}
    \hat{\nabla}_{\lambda} f(\lambda)= \nabla_{m} f(m),
\end{align}
Thus, the Fisher information matrix is no longer required in the natural gradient computation. This variable transformation will be applied in the next section to simplify the natural gradient calculation.

\subsection{Calculate the natural gradients}
In the following, we will compute the natural gradient $\hat{\nabla}_{\lambda_{n}^{s}} \mathcal{L}_s(\lambda_{n}^{s})$ as presented in Section \ref{Derivation of the natural gradient}. 

Recall from Section \ref{Derivation of the natural gradient} that both the predictive prior and variational distribution at sensor $s$ are independent Gaussian distributions: $\hat{p}(X_n)=\prod_{k=1}^K\hat{p}(X_{n,k}; \eta_{n,k}^{s})$ and $q_n(X_n;\!\lambda_{n}^{s})\!=\!\prod_{k=1}^K \!q_n(X_{n,k};\!\lambda_{n,k}^s)$, expressed in the exponential family form:
\begin{align} \notag
   & \hat{p}_n(X_{n,k}; \eta_{n,k}^{s} )=  h(X_{n,k}) \exp \left( {\eta_{n,k}^{s}}^\top T(X_{n,k}) - A(\eta_{n,k}^{s}) \right) \\\notag
 &   q_n(X_{n,k}; \lambda_{n,k}^{s})=  h(X_{n,k}) \exp \left( {\lambda_{n,k}^{s}}^\top T(X_{n,k}) - A(\lambda_{n,k}^{s}) \right)
\end{align}
where $\eta_{n,k}^{s}$ and $\lambda_{n,k}^{s}$ are the natural parameters of $\hat{p}_n(X_{n,k}; \eta_{n,k}^{s} )$ and $q_n(X_{n,k}; \lambda_{n,k}^{s})$, respectively. Since both are Gaussian distributions, they share the same base function $ h(X_{n,k})$, sufficient statistics $T(X_{n,k})$, and log partition function $A(\lambda_{n,k}^{s})$. 
Additionally, the sufficient statistics expectation parameter $m_{n,k}^{s}=\E_{q(X_{n,k};\lambda_{n,k}^s)} T(X_{n,k})$ is defined in Section \ref{Derivation of the natural gradient}. The relationship between the expectation parameter $m_{n,k}^{s}$, the natural parameter $\eta_{n,k}^{s},\lambda_{n,k}^{s}$, and the Gaussian mean and covariance are given in \eqref{eq: natural parameters definition} and
 \eqref{eq:mean parameters}, summarised below:
\begin{align}\label{eq: apx natural , expectation parameters summary}
 &\eta_{n,k}^{s}=\begin{bmatrix}
\eta_{n,k}^{s,1}\\\eta_{n,k}^{s,2}
 \end{bmatrix}
 =\begin{bmatrix}
    (\Sigma^{k*,s}_{n|n-1})^{-1} \mu^{k*,s}_{n|n-1} \\
    -\frac{1}{2} (\Sigma^{k*,s}_{n|n-1})^{-1}
 \end{bmatrix}, \quad \lambda_{n,k}^{s}=\begin{bmatrix}
\lambda_{n,k}^{s,1}\\\lambda_{n,k}^{s,2}
 \end{bmatrix}
 =\begin{bmatrix}
    (\Sigma^{k,s}_{n|n})^{-1} \mu^{k,s}_{n|n} \\
    -\frac{1}{2} (\Sigma^{k,s}_{n|n})^{-1}
 \end{bmatrix},
  \quad 
  m_{n,k}^{s}=\begin{bmatrix}
     m_{n,k}^{s,1}\\m_{n,k}^{s,2}
 \end{bmatrix}
 =\begin{bmatrix}
    \mu^{k,s}_{n|n} \\
    \mu^{k,s}_{n|n} [\mu^{k,s}_{n|n}]^\top+\Sigma^{k,s}_{n|n}
 \end{bmatrix}
\end{align}

We now begin the computation. Using \eqref{eq: apx L=L1+L2} and the natural gradient definition from \eqref{eq:natural gradients the Fisher information matrix}, we have 
\begin{align} \label{eq: apx natural l=l1+l2}
    \hat{\nabla}_{\lambda_{n}^{s}} \mathcal{L}_s(\lambda_{n}^{s})=\hat{\nabla}_{\lambda_{n}^{s}} \mathcal{L}_s^1(\lambda_n^s)+\hat{\nabla}_{\lambda_{n}^{s}} \mathcal{L}_s^2(\lambda_n^s),
\end{align}
where $\mathcal{L}_s^1(\lambda_n^s),\mathcal{L}_s^2(\lambda_n^s)$ are given in \eqref{eq: apx L1} and \eqref{eq: apx L2}, respectively:
\begin{align} \notag
\mathcal{L}_s^1(\lambda_n^s)=&\frac{1}{N_s} \E_{q_n(X_n; \lambda_{n}^s)}\log \frac{\hat{p}_n(X_n)}{q_n(X_n; \lambda_{n}^s)},\\ \notag
\mathcal{L}_s^2(\lambda_n^s)=&\E_{q_n(X_n; \lambda_{n}^s)q_n^{s,*}(\theta_n^{s})}\log p(Y_{n}^{s}|\theta_{n}^{s},X_{n}),
\end{align}
with $q_n^{s,*}(\theta_n^{s})$ treated treated as independent of $\lambda_n^s$ during the (natural) gradient evaluation. 

We will now first compute $\hat{\nabla}_{\lambda_{n}^{s}}\mathcal{L}_s^1(\lambda_n^s)$. Note that
% First, for the computation convenience, we rewrite $\mathcal{L}_s(\lambda_{n}^{s})$ into following parts:
% \begin{align} 
% \mathcal{L}_s(\lambda_{n}^{s})&= 
% \mathcal{L}_s^1(\lambda_{n}^{s}) + \mathcal{L}_s^2(\lambda_{n}^{s}) + \mathcal{L}_s^3(\lambda_{n}^{s})
% \end{align}
% where 
% \begin{align} 
% &\mathcal{L}_s^1(\lambda_{n}^{s})=\frac{1}{N_s}\E_{q_n(X_n; \lambda_{n})}\log\frac{\hat{p}_n(X_n)}{q_n(X_n; \lambda_{n})}\\
% &\mathcal{L}_s^2(\lambda_{n}^{s})=\E_{q_n(X_n; \lambda_{n})q_n(\theta_n; \rho_n^*(\lambda_n))}\log p(Y_{n}^{s}|\theta_{n}^{s},X_{n})\\ &\mathcal{L}_s^3(\lambda_{n}^{s})=\E_{q_n(\theta_n; \rho_n^*(\lambda_n))}\log  \frac{p(\theta_{n}^{s}|M_n^{s})}{q_n(\theta_{n}^{s}; \rho_n^*(\lambda_n))} 
% \end{align}
% In this way, the natural gradient can be computed individually in these three parts:
% \begin{align} 
% \hat{\nabla}_{\lambda_{n}^{s}}\mathcal{L}_s(\lambda_{n}^{s})&= 
% \hat{\nabla}_{\lambda_{n}^{s}}\mathcal{L}_s^1(\lambda_{n}^{s}) + \hat{\nabla}_{\lambda_{n}^{s}}\mathcal{L}_s^2(\lambda_{n}^{s}) 
% \end{align}
% where $\mathcal{L}_s^3(\lambda_{n}^{s})=0$ according to Property 2.1.
% First, let consider the component of calculating the first part $\mathcal{L}_s^1(\lambda_{n}^{s})$, that is:
\begin{align} \notag
&\mathcal{L}_s^1(\lambda_{n}^{s})= \frac{1}{N_s} \sum_{k=1}^{K} \left[ 
\E_{q_n(X_{n,k}; \lambda_{n,k}^{s})}\log p_n(X_{n,k}; \eta_{n,k}^{s} )) -  \E_{q_n(X_{n,k}; \lambda_{n,k}^{s})}\log q_n(X_{n,k}; \lambda_{n,k}^{s}))\right]\\ \notag
& = \frac{1}{N_s} \sum_{k=1}^{K} \left[ 
\E_{q_n(X_{n,k}; \lambda_{n,k}^{s})} \left( {\eta_{n,k}^{s}}^{\top} T(X_{n,k}) - A(\eta_{n,k}^{s}) \right) -  \E_{q_n(X_{n,k}; \lambda_{n,k}^{s})}\left( {\lambda_{n,k}^{s}}^{\top} T(X_{n,k}) - A(\lambda_{n,k}^{s}) \right) \right] \\ \notag
& =\frac{1}{N_s} \sum_{k=1}^{K} \left[({\eta_{n,k}^{s}}^{\top}- {\lambda_{n,k}^{s}}^{\top}) \nabla_{\lambda_{n,k}^{s}}A(\lambda_{n,k}^{s})+ A(\lambda_{n,k}^{s})- A(\eta_{n,k}^{s}) \right]
\end{align}
where the last line uses the property in \eqref{eq:the natural sufficient statistics}.
For each component $\lambda_{n,k}^{s}$,  the gradient can then be calculated as
\begin{align} \notag
\nabla_{\lambda_{n,k}^{s}}\mathcal{L}_s^1(\lambda_{n,k}^{s})&=\frac{1}{N_s} \left( \nabla^2_{\lambda_{n,k}^{s}}A(\lambda_{n,k}^{s}){\eta_{n,k}^{s}} - \nabla^2_{\lambda_{n,k}^{s}}A(\lambda_{n,k}^{s}){\lambda_{n,k}^{s}} - \nabla_{\lambda_{n,k}^{s}}A(\lambda_{n,k}^{s})+ \nabla_{\lambda_{n,k}^{s}}A(\lambda_{n,k}^{s}) \right)   \\ \notag
 &= \frac{1}{N_s}  \nabla^2_{\lambda_{n,k}^{s}}A(\lambda_{n,k}^{s})({\eta_{n,k}^{s}}- {\lambda_{n,k}^{s}})  \\ \notag
 &= \frac{1}{N_s}  G(\lambda_{n,k}^{s})({\eta_{n,k}^{s}}- {\lambda_{n,k}^{s}}),
\end{align}
where the property in \eqref{eq:Fisher information matrix1} is used to obtain the last line. Consequently, the natural gradient is given by: $\hat{\nabla}_{\lambda_{n,k}^{s}}\mathcal{L}_s^1(\lambda_{n,k}^{s})=G(\lambda_{n,k}^{s})^{-1}\nabla_{\lambda_{n,k}^{s}} \mathcal{L}_s(\lambda_{n,k}^{s})= \frac{1}{N_s}  (\eta_{n,k}^{s}- \lambda_{n,k}^{s})$. Then, according to \eqref{eq: apx natural , expectation parameters summary}, each component of the natural gradient has the following form:
\begin{align}  \label{eq: apx l1 natural gradient}
\begin{aligned}
&\hat{\nabla}_{\lambda_{n,k}^{s,1}}\mathcal{L}_s^1(\lambda_{n,k}^{s})= \frac{1}{N_s}  (\eta_{n,k}^{s,1}- \lambda_{n,k}^{s,1})=\frac{1}{N_s}  \left[ (\Sigma^{k*,s}_{n|n-1})^{-1} \mu^{k*,s}_{n|n-1}-  (\Sigma^{k,s}_{n|n})^{-1} \mu^{k,s}_{n|n}\right] \\
&\hat{\nabla}_{\lambda_{n,k}^{s,2}}\mathcal{L}_s^1(\lambda_{n,k}^{s})= \frac{1}{N_s}  (\eta_{n,k}^{s,2}- \lambda_{n,k}^{s,2})= \frac{1}{2N_s}  \left[(\Sigma^{k,s}_{n|n})^{-1}-(\Sigma^{k*,s}_{n|n-1})^{-1}\right] 
\end{aligned}
\end{align}

Next, to compute $\hat{\nabla}_{\lambda_{n}^{s}}\mathcal{L}_s^2(\lambda_n^s)$, we rewrite the expression of $\mathcal{L}_s^2(\lambda_{n,k}^{s})$ given in \eqref{eq: apx L2 derived} in terms of $m_{n,k}^{s}$, using the relationship in \eqref{eq: apx natural , expectation parameters summary}:
\begin{align} 
\mathcal{L}_s^2(m_{n,k}^{s})&=C_{xy}^s-\frac{1}{2} \sum_{k=1}^K  \mathrm{Tr}\left(H^\top(\overbar{R}_n^{k,s})^{-1}H\Sigma^{k,s}_{n|n}\right) -\frac{1}{2} \sum_{k=1}^K  ( H\mu^{k,s}_{n|n}-\overbar{Y}_n^{k,s})^\top (\overbar{R}_n^{k,s})^{-1} (H\mu^{k,s}_{n|n}-\overbar{Y}_n^{k,s})  \\\notag
&=C_{xy}^s -\frac{1}{2} \sum_{k=1}^K  \mathrm{Tr}\left(H^\top(\overbar{R}_n^{k,s})^{-1}H(m_{n,k}^{s,2}-m_{n,k}^{s,1}(m_{n,k}^{s,1})^\top)\right)  -\frac{1}{2} \sum_{k=1}^K  ( Hm_{n,k}^{s,1}-\overbar{Y}_n^{k,s})^\top (\overbar{R}_n^{k,s})^{-1} (Hm_{n,k}^{s,1}-\overbar{Y}_n^{k,s}),
\end{align}
where $\overbar{Y}_n^{k,s}$ and $\overbar{R}_n^{k,s}$ are given in \eqref{eq:pseudomeas Y}, and are treated as independent of $\lambda_n^s$ during gradient evaluation. Subsequently, applying the matrix derivative formulas in \cite{petersen2008matrix}, the gradients with respect to the expectation parameters $m_{n,k}^{s,1}$ and $m_{n,k}^{s,2}$ are
\begin{align} 
{\nabla}_{m_{n,k}^{s,1}}\mathcal{L}_s^2(m_{n,k}^{s})&= H^\top  (\overbar{R}_n^{k,s})^{-1} \overbar{Y}_n^{k,s}
\end{align}
\begin{align} {\nabla}_{m_{n,k}^{s,2}}\mathcal{L}_s^2(m_{n,k}^{s})&= -\frac{1}{2}H^\top  (\overbar{R}_n^{k,s})^{-1}H
\end{align}
Using the property in \eqref{eq: apx natural gradient is expectation gradient}, these correspond to the required natural gradients: $\hat{\nabla}_{\lambda_{n,k}^{s,1}}\mathcal{L}_s^2(\lambda_{n,k}^{s})={\nabla}_{m_{n,k}^{s,1}}\mathcal{L}_s^2(m_{n,k}^{s})$, $\hat{\nabla}_{\lambda_{n,k}^{s,2}}\mathcal{L}_s^2(\lambda_{n,k}^{s})={\nabla}_{m_{n,k}^{s,2}}\mathcal{L}_s^2(m_{n,k}^{s})$. Finally, as indicated in \eqref{eq: apx natural l=l1+l2}, combining these results with the derived $\hat{\nabla}_{\lambda_{n,k}^{s,1}}\mathcal{L}_s^1(\lambda_{n,k}^{s})$, $\hat{\nabla}_{\lambda_{n,k}^{s,2}}\mathcal{L}_s^1(\lambda_{n,k}^{s})$ from \eqref{eq: apx l1 natural gradient} yields the overall natural gradients $\hat{\nabla}_{\lambda_{n,k}^{s,1}}\mathcal{L}_s(\lambda_{n,k}^{s})$, $\hat{\nabla}_{\lambda_{n,k}^{s,2}}\mathcal{L}_s(\lambda_{n,k}^{s})$ as shown in \eqref{eq:natural gradient total 1} and \eqref{eq:natural gradient total 2}.

\section{Details of proposed trackers and pseudocodes}\label{apx: tracker and pseudocodes}
\subsection{Decentralised gradient
variational multi-object trackers with with diminishing stepsize
(DeG-VT-DS)}
Recall from \eqref{eq: lambda define normal gradient} that the local estimate of the optimised variational parameter is defined as $\lambda_{n,k}^s=[\mu^{k,s}_{n|n},\Sigma^{k,s}_{n|n}], k=1,...,K$ for both DeG-VT-DS and DeG-VT-GT algorithms. For DeG-VT-DS algorithm, the update of $\lambda_{n,k}^s$ for jointly optimising the LM-ELBO $\mathcal{L}(\lambda_{n})$ follows \eqref{eq:update equation 1}. Specifically, the update equation for each parameter estimate $\mu^{k,s}_{n|n}$ and $\Sigma^{k,s}_{n|n}$, $k=1,...,K$, at iteration $i$ and each sensor $s$, is given by
\begin{align}\label{eq:update mu 1}
   & \mu^{k,s}_{n|n}(i+1)=\sum_{j=1}^{N_s} w_{sj}(i)  \mu^{k,j}_{n|n}(i) + \alpha_i \nabla_{\mu^{k,s}_{n|n}}  \mathcal{L}_s(\lambda_{n}^{s}(i))\\\label{eq:update sigmal 1}
   & \Sigma^{k,s}_{n|n}(i+1)=\sum_{j=1}^{N_s} w_{sj}(i)  \Sigma^{k,j}_{n|n}(i) + \alpha_i \nabla_{\Sigma^{k,s}_{n|n}}  \mathcal{L}_s(\lambda_{n}^{s}(i))
\end{align} 
where each gradient component is derived in detail in Appendix \ref{apx: gradient derivation} (and briefly outlined in Section \ref{sec: standard gradient derivation}) as
\vspace{-0.5em}
\begin{align} \label{eq: gradient mu}
 \nabla_{\mu^{k,s}_{n|n}} \mathcal{L}_s(\lambda_{n}^{s}(i))&={\nabla}_{\mu^{k,s}_{n|n}} \mathcal{L}_s(\lambda_{n}^{s})|_{\mu^{k,s}_{n|n}=\mu^{k,s}_{n|n}(i)}
 =\frac{1}{N_s} (\Sigma^{k*,s}_{n|n-1})^{-1}(\mu^{k,s}_{n|n}(i)-\mu^{k*,s}_{n|n-1}) +H^\top (\overbar{R}_n^{k,s})^{-1} (\overbar{Y}_n^{k,s}(i)-H \mu^{k,s}_{n|n}(i)) 
\end{align}
\begin{align} \label{eq: gradient sigma}
\nabla_{\Sigma^{k,s}_{n|n}} \mathcal{L}_s(\lambda_{n}^{s}(i))&={\nabla}_{\Sigma^{k,s}_{n|n}} \mathcal{L}_s(\lambda_{n}^{s})|_{\Sigma^{k,s}_{n|n}=\Sigma^{k,s}_{n|n}(i)}
=\frac{1}{2N_s}\left( (\Sigma^{k,s}_{n|n}(i))^{-1}- (\Sigma^{k*,s}_{n|n-1})^{-1} \right)- \frac{1}{2}H^\top(\overbar{R}_n^{k,s}(i))^{-1} H
\end{align}
The local pseudo-measurement $\overbar{Y}_n^{k,s}$ and covariance $\overbar{R}_n^{k,s}$ in \eqref{eq: gradient mu} and \eqref{eq: gradient sigma} at each sensor $s$ are given by
\begin{align}\label{eq:y r}
    \overbar{R}_n^{k,s}(i)=&\frac{R_k^s}{\sum_{j=1}^{M_n^{s}}q_n^{s,*}(\theta_{n,j}^{s}=k)}, \ \ \ \ \
\overbar{Y}_n^{k,s}(i)=\frac{\sum_{j=1}^{M_n^{s}}Y_{n,j}^{s}q_n^{s,*}(\theta_{n,j}^{s}=k)}{\sum_{j=1}^{M_n^{s}}q_n^{s,*}(\theta_{n,j}^{s}=k)},
\end{align}
where $q_n^{s,*}(\theta_{n,j}^{s})$ is computed using the most recent local estimate $\lambda_n^s(i)$ as
\begin{align}  \label{eq:qtheta}
   & q_n^{s,*}(\theta_{n,j}^{s})
    \propto\frac{\Lambda_0^{s}}{V^{s}}\delta[\theta_{n,j}^{s}=0]+\sum_{k=1}^K\Lambda_k^{s} l_k^{s}\delta[\theta_{n,j}^{s}=k],\\
   & l_k^{s}=\mathcal{N}(Y_{n,j}^{s};H\mu_{n|n}^{k,s}(i),R_k^{s})\text{exp}(-0.5\text{Tr}({(R_k^{s})}^{-1}H\Sigma_{n|n}^{k,s}(i) H^\top)).
\end{align}

The full procedure of DeG-VT-DS, including prediction and update steps, can be seen in Algorithm \ref{Algo:DeG-VT-DS}.
\begin{algorithm}[htp!]
% \SetAlgoLined
%\algsetup{linenosize=\small}
 % \scriptsize
\caption{{DeG-VT-DS at time step $n$ for each sensor $s$}}
\label{Algo:DeG-VT-DS}
\textbf{Input}: $q^*_{n-1}(X_{n-1,k};\lambda_{n-1,k}^s)= \mathcal{N}(X_{n-1,k};\mu^{k*,s}_{n-1|n-1},\Sigma^{k*,s}_{n-1|n-1})$, $k=1,...,K$, $Y_n^s$, maximum iteration $I_{max}$.\\
\textbf{Output}: $q^*_{n}(X_{n};\lambda_{n}^{s})=\prod_{k=1}^K q^*_{n,k}(X_{n,k};\lambda_{n,k}^{s}) =\prod_{k=1}^K \mathcal{N}(X_{n,k};\mu^{k*,s}_{n|n},\Sigma^{k*,s}_{n|n})$. \\
\For {$k=1,2,...,K$} {Prediction step: $\hat{p}_n(X_{n,k})=\mathcal{N}(X_{n,k};\mu^{k*,s}_{n|n-1},\Sigma^{k*,s}_{n|n-1})$ using \eqref{eq:predictive prior computation}.} 
{Initialisation}:  
% $q_n(X_{n,k}; \lambda_{n,k}^{s})= \mathcal{N}(X_{n,k};\mu^{k,s}_{n|n}(0),\Sigma^{k,s}_{n|n}(0))$, 
For $k=1,2,...,K$, set $\mu^{k,s}_{n|n}(0)= \mu^{k*,s}_{n|n-1}$, 
$\Sigma^{k,s}_{n|n}(0)=\Sigma^{k*,s}_{n|n-1}$.\\
\For {$i=0,1,...,I_{max}$}  
{
Exchange variables $\mu^{k,s}_{n|n}(i),\Sigma^{k,s}_{n|n}(i)$ $(k=1,2,...,K)$ with the current neighbors of sensor $s$ in $\mathcal{N}_s(i)$.\\
For $j=1,...,M_n$, compute $q^{s,*}_n(\theta_{n,j})$ using \eqref{eq:qtheta}. \\
Compute the gradients $\nabla_{\mu^{k,s}_{n|n}},\nabla_{\Sigma^{k,s}_{n|n}}$ in \eqref{eq: gradient mu}, \eqref{eq: gradient sigma}.\\
\For {$k=1,2,...,K$}
   {Update $\mu^{k,s}_{n|n}(i+1)$,
   $\Sigma^{k,s}_{n|n}(i+1)$ according to \eqref{eq:update mu 1}, \eqref{eq:update sigmal 1}, \eqref{eq: metro weights}.\\
   }
}
After convergence,
$q^*_{n,k}(X_{n,k};\lambda_{n,k}^{s})=\mathcal{N}(X_{n,k};\mu^{k*,s}_{n|n},\Sigma^{k*,s}_{n|n})$, where $\mu^{k*,s}_{n|n},\Sigma^{k*,s}_{n|n}$ are the final updates of $\mu^{k,s}_{n|n}(i),\Sigma^{k,s}_{n|n}(i)$.
\end{algorithm}

\subsection{Decentralised gradient
variational multi-object trackers with with gradient tracking
(DeG-VT-GT)}
Recall from \eqref{eq: lambda define normal gradient} that the local estimate of the optimised variational parameter is defined as $\lambda_{n,k}^s=[\mu^{k,s}_{n|n},\Sigma^{k,s}_{n|n}], k=1,...,K$ for both DeG-VT-DS and DeG-VT-GT algorithms. For DeG-VT-GT algorithm, the update of $\lambda_{n,k}^s$ and gradient estimates $\boldsymbol{\xi}_{n,k}^{s}=[\boldsymbol{\xi}_{n,k}^{s,1},\boldsymbol{\xi}_{n,k}^{s,2}]$ for jointly optimising the LM-ELBO $\mathcal{L}(\lambda_{n})$ follows \eqref{eq:update equations for each variational parameter}, \eqref{eq:update equation 2}. Specifically, the update equation for each parameter estimate $\mu^{k,s}_{n|n}$, $\Sigma^{k,s}_{n|n}$ and gradient estimates $\boldsymbol{\xi}_{n,k}^{s,1},\boldsymbol{\xi}_{n,k}^{s,2}$
$k=1,...,K$, at iteration $i$ and each sensor $s$, are given by

% For DeG-VT-GT algorithm, the update equations for variational parameters $\lambda_{n,k}^s=[\mu^{k,s}_{n|n},\Sigma^{k,s}_{n|n}]^{\top}, k=1,...,K$ and gradient estimates $\boldsymbol{\xi}_{n}^{s}=[\boldsymbol{\xi}_{n}^{s,1},\boldsymbol{\xi}_{n}^{s,2}]$ at each iteration $i$ at each sensor $s$ are
\begin{align}\label{eq:update lambda 1}
   & \mu^{k,s}_{n|n}(i+1)=\sum_{j=1}^{N_s} w_{sj}(i)  \mu^{k,j}_{n|n}(i) + \alpha \boldsymbol{\xi}_{n,k}^{s,1}(i),  \\\label{eq:update lambda 2}
   & \Sigma^{k,s}_{n|n}(i+1)=\sum_{j=1}^{N_s} w_{sj}(i)  \Sigma^{k,j}_{n|n}(i) + \alpha \boldsymbol{\xi}_{n,k}^{s,2}(i).\\ \label{eq: update xi 1}
    &\boldsymbol{\xi}_{n,k}^{s,1}(i + 1)
      = \sum_{j=1}^{N_s} w_{sj}(i) \boldsymbol{\xi}_{n,k}^{j,1}(i)+\nabla_{\mu^{k,s}_{n|n}}  \mathcal{L}_s(\lambda_{n}^{s}(i+1))-\nabla_{\mu^{k,s}_{n|n}}  \mathcal{L}_s(\lambda_{n}^{s}(i)),\\ \label{eq: update xi 2}
      &\boldsymbol{\xi}_{n,k}^{s,2}(i + 1)
      = \sum_{j=1}^{N_s} w_{sj}(i) \boldsymbol{\xi}_{n,k}^{j,2}(i)+\nabla_{\Sigma^{k,s}_{n|n}}  \mathcal{L}_s(\lambda_{n}^{s}(i+1))-\nabla_{\Sigma^{k,s}_{n|n}}  \mathcal{L}_s(\lambda_{n}^{s}(i)).
\end{align} 
The full procedure of DeG-VT-GT, including prediction and update steps, can be seen in Algorithm \ref{Algo:DeG-VT-GT}.
\begin{algorithm}[htp!]
\caption{{DeG-VT-GT at time step $n$ for each sensor $s$}}
\label{Algo:DeG-VT-GT}
\textbf{Input}: $q^*_{n-1}(X_{n-1,k};\lambda_{n-1,k}^s)= \mathcal{N}(X_{n-1,k};\mu^{k*,s}_{n-1|n-1},\Sigma^{k*,s}_{n-1|n-1})$, $k=1,...,K$, $Y_n^s$, maximum iteration $I_{max}$.\\
\textbf{Output}: $q^*_{n}(X_{n};\lambda_{n}^{s})=\prod_{k=1}^K q^*_{n,k}(X_{n,k};\lambda_{n,k}^{s}) =\prod_{k=1}^K \mathcal{N}(X_{n,k};\mu^{k*,s}_{n|n},\Sigma^{k*,s}_{n|n})$. \\
\For {$k=1,2,...,K$} {Prediction step: $\hat{p}_n(X_{n,k})=\mathcal{N}(X_{n,k};\mu^{k*,s}_{n|n-1},\Sigma^{k*,s}_{n|n-1})$ using \eqref{eq:predictive prior computation}.} 
% \textit{Initialisation}:  $q_n(X_{n,k}; \lambda_{n,k}^{s})= \mathcal{N}(X_{n,k};\mu^{k,s}_{n|n}(0),\Sigma^{k,s}_{n|n}(0))$, $k=1,2,...,K$:\\
% $\mu^{k*,s}_{n|n}(0)= \mu^{k*,s}_{n|n-1}$, 
% $\Sigma^{k*,s}_{n|n}(0)=\Sigma^{k*,s}_{n|n-1}$,$\boldsymbol{\xi}_{n}^{s,1}(0)=0,\boldsymbol{\xi}_{n}^{s,2}(0)=0$\\
{Initialisation}: For $k=1,2,...,K$, set $\mu^{k*,s}_{n|n}(0)= \mu^{k*,s}_{n|n-1}$, $\Sigma^{k*,s}_{n|n}(0)=\Sigma^{k*,s}_{n|n-1}$, $\boldsymbol{\xi}_{n,k}^{s,1}(0)=\nabla_{\lambda_{n,k}^{s,1}}\mathcal{L}_s(\lambda_{n,k}^{s}(0))$, $\boldsymbol{\xi}_{n,k}^{s,2}(0)=\nabla_{\lambda_{n,k}^{s,2}}\mathcal{L}_s(\lambda_{n,k}^{s}(0))$.\\
\For {$i=0,1,...,I_{max}$}  
{
Exchange variables $\mu^{k,s}_{n|n}(i),\Sigma^{k,s}_{n|n}(i)$, $\boldsymbol{\xi}_{n,k}^{s}(i)$ $(k=1,2,...,K)$ with the current neighbors of sensor $s$ in $\mathcal{N}_s(i)$.\\
For $j=1,...,M_n$, compute $q^{s,*}_n(\theta_{n,j})$ using \eqref{eq:qtheta}.\\
Compute the gradients $\nabla_{\mu^{k,s}_{n|n}},\nabla_{\Sigma^{k,s}_{n|n}}$ in \eqref{eq: gradient mu}, \eqref{eq: gradient sigma}.\\
\For {$k=1,2,...,K$}
   {Update $\mu^{k,s}_{n|n}(i+1)$,
   $\Sigma^{k,s}_{n|n}(i+1)$ according to \eqref{eq:update lambda 1}, \eqref{eq:update lambda 2}, \eqref{eq: metro weights}.\\
    Update $\boldsymbol{\xi}_{n,k}^{s,1}(i+1),\boldsymbol{\xi}_{n,k}^{s,2}(i+1)$  according to \eqref{eq: update xi 1}, \eqref{eq: update xi 2}, \eqref{eq: metro weights}.\\
   }
}
After convergence,
$q^*_{n,k}(X_{n,k};\lambda_{n,k}^{s})=\mathcal{N}(X_{n,k};\mu^{k*,s}_{n|n},\Sigma^{k*,s}_{n|n})$, where $\mu^{k*,s}_{n|n},\Sigma^{k*,s}_{n|n}$ are the final updates of $\mu^{k,s}_{n|n}(i),\Sigma^{k,s}_{n|n}(i)$.
\end{algorithm}

\subsection{Decentralised natural gradient
variational multi-object trackers with with diminishing stepsize
(DeNG-VT-DS)}
Recall from \eqref{eq: lambda define normal gradient} that, for both DeNG-VT-DS and DeNG-VT-GT algorithms, the local estimate of the optimised variational parameter is defined as $\lambda_{n,k}^s=[\lambda_{n,k}^{s,1}, \lambda_{n,k}^{s,2}]$ ($k=1,...,K$), with each $\lambda_{n,k}^{s,1}, \lambda_{n,k}^{s,2}$ defined in \eqref{eq: natural parameters definition}. For DeG-VT-DS algorithm, the update of $\lambda_{n,k}^s$ for jointly optimising the LM-ELBO $\mathcal{L}(\lambda_{n})$ follows \eqref{eq:update equation 1}. Specifically, the update equation for each parameter estimate $\lambda_{n,k}^{s,1}, \lambda_{n,k}^{s,2}$, $k=1,...,K$, at iteration $i$ and each sensor $s$, is given by

% For DeNG-VT-DS algorithm, the update equation for each variational parameter $\lambda_{n,k}^s=[\lambda_{n,k}^{s,1}, \lambda_{n,k}^{s,2}]^{\top}, k=1,...,K$ at each iteration $i$ at each sensor $s$ for jointly optimising the LM-ELBO $\mathcal{L}(\lambda_{n})$ is
\begin{align}\label{eq:DeNG-VT-DS 1}
   & \lambda_{n,k}^{s,1}(i+1)=\sum_{j=1}^{N_s} w_{sj}(i) \lambda_{n,k}^{j,1}(i) + \alpha_i \hat{\nabla}_{\lambda_{n,k}^{s,1}} \mathcal{L}_s(\lambda_{n}^{s}(i))\\\label{eq:DeNG-VT-DS 2}
    & \lambda_{n,k}^{s,2}(i+1)=\sum_{j=1}^{N_s} w_{sj}(i) \lambda_{n,k}^{j,2}(i) + \alpha_i \hat{\nabla}_{\lambda_{n,k}^{s,2}} \mathcal{L}_s(\lambda_{n}^{s}(i))
\end{align} 
where each natural gradient component is derived in detail in Appendix \ref{apx: natural gradient derivation} (and briefly outlined in Section \ref{Derivation of the natural gradient}) as
\begin{align} \label{eq: natural gradients 1}
\hat{\nabla}_{\lambda_{n,k}^{s,1}}\mathcal{L}_s(\lambda_{n,k}^{s}(i))&= \frac{1}{N_s}  \left[ (\Sigma^{k*,s}_{n|n-1})^{-1} \mu^{k*,s}_{n|n-1}-  (\Sigma^{k,s}_{n|n}(i))^{-1} \mu^{k,s}_{n|n}\right] +H^\top  (\overbar{R}_n^{k,s}(i))^{-1} \overbar{Y}_n^{k,s}(i) \\ \label{eq: natural gradients 2}
\hat{\nabla}_{\lambda_{n,k}^{s,2}}\mathcal{L}_s(\lambda_{n,k}^{s}(i))&= \frac{1}{2N_s}  [(\Sigma^{k,s}_{n|n}(i))^{-1}-(\Sigma^{k*,s}_{n|n-1})^{-1}] -\frac{1}{2}H^\top  (\overbar{R}_n^{k,s}(i))^{-1}H 
\end{align}
where local pseudo-measurement $\overbar{Y}_n^{k,s}$ and covariance $\overbar{R}_n^{k,s}$ at each sensor $s$ have the same form as in \eqref{eq:y r}.

The full procedure of DeNG-VT-DS, including prediction and update steps, can be seen in Algorithm \ref{Algo:DeNG-VT-DS}.
\begin{algorithm}[htp!]
% \SetAlgoLined
\caption{{DeNG-VT-DS at time step $n$ for each sensor $s$}}
\label{Algo:DeNG-VT-DS}
\textbf{Input}: $q^*_{n-1}(X_{n-1};\lambda_{n-1}^{s}),Y_n^s$, maximum iteration $I_{max}$.\\
\textbf{Output}: $q^*_{n}(X_{n};\lambda_{n}^{s})=\prod_{k=1}^K q^*_{n,k}(X_{n,k};\lambda_{n,k}^{s})$.\\
\For {$k=1,2,...,K$} {Prediction step: $\hat{p}_n(X_{n,k})=\mathcal{N}(X_{n,k};\mu^{k*,s}_{n|n-1},\Sigma^{k*,s}_{n|n-1})$ using \eqref{eq:predictive prior computation}.} 
{Initialisation}: For $k=1,2,...,K$, set
$\lambda_{n,k}^{s,1}(0)= (\Sigma^{k*,s}_{n|n-1})^{-1} \mu^{k*,s}_{n|n-1}$, 
$\lambda_{n,k}^{s,2}(0)=-\frac{1}{2} (\Sigma^{k*,s}_{n|n-1})^{-1}$.\\
\For {$i=0,1,...,I_{max}$}  
{
Exchange variables $\lambda_{n,k}^{s,1}(i),\lambda_{n,k}^{s,2}(i)$ $(k=1,2,...,K)$ with the current neighbors of sensor $s$ in $\mathcal{N}_s(i)$.\\
For $j=1,...,M_n$, compute $q^{s,*}_n(\theta_{n,j})$ using \eqref{eq:qtheta}. \\
Compute the natural gradients $\hat{\nabla}_{\lambda_{n,k}^{s,1}},\hat{\nabla}_{\lambda_{n,k}^{s,2}}$ in \eqref{eq: natural gradients 1}, \eqref{eq: natural gradients 2}.\\
\For {$k=1,2,...,K$}
   {Update $\lambda_{n,k}^{s,1}(i+1)$,
   $\lambda_{n,k}^{s,2}(i+1)$  according to \eqref{eq:DeNG-VT-DS 1}, \eqref{eq:DeNG-VT-DS 2}, \eqref{eq: metro weights}.\\
   }
}
After convergence, $q^*_{n,k}(X_{n,k};\lambda_{n,k}^{s})=\mathcal{N}(X_{n,k};\mu^{k*,s}_{n|n},\Sigma^{k*,s}_{n|n})$, where $\mu^{k*,s}_{n|n}=-\frac{1}{2}(\lambda_{n,k}^{s,2})^{-1}\lambda_{n,k}^{s,1}$,
$\Sigma^{k*,s}_{n|n}=-\frac{1}{2}(\lambda_{n,k}^{s,2})^{-1}$, and $\lambda_{n,k}^{s,1},\lambda_{n,k}^{s,2}$ are the final updates of $\lambda_{n,k}^{s,1}(i),\lambda_{n,k}^{s,2}(i)$.
\end{algorithm}

\subsection{Decentralised natural gradient
variational multi-object trackers with with gradient tracking
(DeNG-VT-GT)}
Recall from \eqref{eq: lambda define normal gradient} that, for both DeNG-VT-DS and DeNG-VT-GT algorithms, the local estimate of the optimised variational parameter is defined as $\lambda_{n,k}^s=[\lambda_{n,k}^{s,1}, \lambda_{n,k}^{s,2}]$ ($k=1,...,K$), with each $\lambda_{n,k}^{s,1}, \lambda_{n,k}^{s,2}$ defined in \eqref{eq: natural parameters definition}. For DeG-VT-GT algorithm, the update of $\lambda_{n,k}^s$ and gradient estimates $ \hat{\boldsymbol{\xi}}_{n,k}^{s}=[ \hat{\boldsymbol{\xi}}_{n,k}^{s,1}, \hat{\boldsymbol{\xi}}_{n,k}^{s,2}]$ for jointly optimising the LM-ELBO $\mathcal{L}(\lambda_{n})$ follows \eqref{eq:update equations for each variational parameter}, \eqref{eq:update equation 2}. Specifically, the update equation for each parameter estimate $\lambda_{n,k}^{s,1}, \lambda_{n,k}^{s,2}$ and gradient estimates $ \hat{\boldsymbol{\xi}}_{n,k}^{s,1}, \hat{\boldsymbol{\xi}}_{n,k}^{s,2}$, $k=1,...,K$, at iteration $i$ and each sensor $s$, are given by

% For DeNG-VT-GT algorithm, the update equation for each variational parameter $\lambda_{n,k}^s=[\lambda_{n,k}^{s,1}, \lambda_{n,k}^{s,2}]^{\top}, k=1,...,K$ and gradient estimates $\hat{\boldsymbol{\xi}}_{n}^{s}=[\hat{\boldsymbol{\xi}}_{n}^{s,1},\hat{\boldsymbol{\xi}}_{n}^{s,2}]$ at each iteration $i$ at each sensor $s$ for jointly optimising the LM-ELBO $\mathcal{L}(\lambda_{n})$ is
\begin{align}\label{eq:DeNG-VT-GT 1}
   & \lambda_{n,k}^{s,1}(i+1)=\sum_{j=1}^{N_s} w_{sj}(i) \lambda_{n,k}^{j,1}(i) + \alpha \hat{\boldsymbol{\xi}}_{n,k}^{s,1}(i)\\\label{eq:DeNG-VT-GT 2}
    & \lambda_{n,k}^{s,2}(i+1)=\sum_{j=1}^{N_s} w_{sj}(i) \lambda_{n,k}^{j,2}(i) + \alpha \hat{\boldsymbol{\xi}}_{n,k}^{s,2} (i)
\end{align} 

\begin{align}\label{eq:gradient estimate GT1}
   \hat{\boldsymbol{\xi}}_{n,k}^{s,1}(i + 1)& = \sum_{j=1}^{N_s} w_{sj}(i) \hat{\boldsymbol{\xi}}_{n,k}^{j,1}(i)+\hat{\nabla}_{\lambda_{n,k}^{s,1}} \mathcal{L}_s(\lambda_{n,k}^{s}(i+1))-\hat{\nabla}_{\lambda_{n,k}^{s,1}} \mathcal{L}_s(\lambda_{n,k}^{s}(i)), \\\label{eq:gradient estimate GT2}
   \hat{\boldsymbol{\xi}}_{n,k}^{s,2}(i + 1)& = \sum_{j=1}^{N_s} w_{sj}(i) \hat{\boldsymbol{\xi}}_{n,k}^{j,2}(i)+\hat{\nabla}_{\lambda_{n,k}^{s,2}} \mathcal{L}_s(\lambda_{n,k}^{s}(i+1))-\hat{\nabla}_{\lambda_{n,k}^{s,2}} \mathcal{L}_s(\lambda_{n,k}^{s}(i)),
\end{align}
The full procedure of DeNG-VT-GT, including prediction and update steps, can be seen in Algorithm \ref{Algo:DeNG-VT-GT}.
\begin{algorithm}[htp!]
% \SetAlgoLined
\caption{{DeNG-VT-GT at time step $n$ for each sensor $s$}}
\label{Algo:DeNG-VT-GT}
\textbf{Input}: $q^*_{n-1}(X_{n-1};\lambda_{n-1}^{s}),Y_n^s$, maximum iteration $I_{max}$.\\
\textbf{Output}: $q^*_{n}(X_{n};\lambda_{n}^{s})=\prod_{k=1}^K q^*_{n,k}(X_{n,k};\lambda_{n,k}^{s})$.\\
\For {$k=1,2,...,K$} {Prediction step: $\hat{p}_n(X_{n,k})=\mathcal{N}(X_{n,k};\mu^{k*,s}_{n|n-1},\Sigma^{k*,s}_{n|n-1})$ using \eqref{eq:predictive prior computation}.} 
{Initialisation}: Set $\lambda_{n,k}^{s,1}(0)= (\Sigma^{k*,s}_{n|n-1})^{-1} \mu^{k*,s}_{n|n-1}$, $\lambda_{n,k}^{s,2}(0)=-\frac{1}{2} (\Sigma^{k*,s}_{n|n-1})^{-1}$, $\hat{\boldsymbol{\xi}}_{n,k}^{s,1}(0)=\hat{\nabla}_{\lambda_{n,k}^{s,1}}\mathcal{L}_s(\lambda_{n,k}^{s}(0))$, $\hat{\boldsymbol{\xi}}_{n,k}^{s,2}(0)=\hat{\nabla}_{\lambda_{n,k}^{s,2}}\mathcal{L}_s(\lambda_{n,k}^{s}(0))$.\\
\For {$i=0,1,...,I_{max}$}  
{
Exchange variables $\mu^{k,s}_{n|n}(i),\Sigma^{k,s}_{n|n}(i)$, $\hat{\boldsymbol{\xi}}_{n,k}^{s}(i)$ $(k=1,2,...,K)$ with the current neighbors of sensor $s$ in $\mathcal{N}_s(i)$.\\
For $j=1,...,M_n$, compute $q^{s,*}_n(\theta_{n,j})$ using \eqref{eq:qtheta}. \\
Compute the natural gradients $\hat{\nabla}_{\lambda_{n,k}^{s,1}},\hat{\nabla}_{\lambda_{n,k}^{s,2}}$ in \eqref{eq: natural gradients 1}, \eqref{eq: natural gradients 2}.\\
\For {$k=1,2,...,K$}
   {Update $\lambda_{n,k}^{s,1}(i+1)$,
   $\lambda_{n,k}^{s,2}(i+1)$  according to \eqref{eq:DeNG-VT-GT 1}, \eqref{eq:DeNG-VT-GT 2}, \eqref{eq: metro weights}.\\
   Update $\hat{\boldsymbol{\xi}}_{n,k}^{s,1}(i + 1)$, $\hat{\boldsymbol{\xi}}_{n,k}^{s,2}(i + 1)$ according to \eqref{eq:gradient estimate GT1}, \eqref{eq:gradient estimate GT2}, \eqref{eq: metro weights}.\\
   }
}
After convergence, $q^*_{n,k}(X_{n,k};\lambda_{n,k}^{s})=\mathcal{N}(X_{n,k};\mu^{k*,s}_{n|n},\Sigma^{k*,s}_{n|n})$, where $\mu^{k*,s}_{n|n}=-\frac{1}{2}(\lambda_{n,k}^{s,2})^{-1}\lambda_{n,k}^{s,1}$,
$\Sigma^{k*,s}_{n|n}=-\frac{1}{2}(\lambda_{n,k}^{s,2})^{-1}$, and $\lambda_{n,k}^{s,1},\lambda_{n,k}^{s,2}$ are the final updates of $\lambda_{n,k}^{s,1}(i),\lambda_{n,k}^{s,2}(i)$.
\end{algorithm}

\section{Handling Non-Convergence: Effective Prior in Decentralised Tracking with Limited Iterations} \label{apx: Tracker Robustness}
Here, we will demonstrate that, as discussed in Section \ref{sec:Robust decentralised tracking}, our decentralised (natural) gradient-based variational trackers still perform sensible inference at time step $n$, even when sensors initially have different predictive priors $\hat{p}_n(X_n;\eta_n^s)$ -- meaning the inference have not yet converged at the previous time step $n-1$, due to the use of limited (natural) gradient descent iterations for efficiency. Specifically, we will show that in this case, the decentralised optimisation at the current time step $n$ still targets the same LM-ELBO in \eqref{eq:locally maximised ELBO form}, but with a different prior $\hat{p}_{eff}(X_n)\propto \prod_{s=1}^{N_s}  \hat{p}_n(X_{n};\eta_{n}^{s} ) ^{1/N_s}$ in place of $\hat{p}_n(X_n)$ in \eqref{eq:locally maximised ELBO form}.

To this end, first recall that our decentralised (natural) gradient-based variational trackers are designed so that all sensors collaboratively optimise $\mathcal{L}(\lambda_{n})=\sum_{s=1}^{N_s}\mathcal{L}_s(\lambda_{n})$, with the LM-ELBO $\mathcal{L}(\lambda_{n})$ and local LM-ELBO $\mathcal{L}_s(\lambda_{n})$ defined in \eqref{eq:locally maximised ELBO form} and \eqref{eq:local LM ELBO}, respectively, as
\begin{align} \label{eq: apx repeat LM-ELBO}
   \mathcal{L}(\lambda_{n})=& \sum_{s=1}^{N_s} \E_{q_n(X_n; \lambda_{n})q_n^*(\theta_n)}\log p(Y_{n}^{s}|\theta_{n}^{s},X_{n}) + 
   \E_{q_n^*(\theta_n)}\log\frac{p(\theta_{n}|M_n)}{q_n^*(\theta_{n})} + \E_{q_n(X_n; \lambda_{n})}\log\frac{\hat{p}_n(X_n)}{q_n(X_n;\lambda_{n})},\\
   \mathcal{L}_s(\lambda_{n})=&\E_{q_n(X_n; \lambda_{n})q_n^*(\theta_n^{s})}\log p(Y_{n}^{s}|\theta_{n}^{s},X_{n}) + \E_{q_n^*(\theta_n^{s})}\log  \frac{p(\theta_{n}^{s}|M_n^{s})}{q_n^*(\theta_{n}^{s})}+ \frac{1}{N_s}\E_{q_n(X_n; \lambda_{n})}\log\frac{\hat{p}_n(X_n)}{q_n(X_n;\lambda_{n})}.
\end{align}
In cases where sensors have different predictive priors $\hat{p}_n(X_{n};\eta_{n}^{s})$ ($s=1,2,...,N_s$) instead of the identical prior $\hat{p}_n(X_n)$, each sensor essentially computes the local (natural) gradient with respect to a different local objective, $\mathcal{L}'_s(\lambda_n)$, defined as:
\begin{align} \label{eq:local LM ELBO limited}
  & \mathcal{L}'_s(\lambda_{n})
     = \E_{q_n(X_n; \lambda_{n})q_n^*(\theta_n^{s})}\log p(Y_{n}^{s}|\theta_{n}^{s},X_{n})   + \E_{q_n^*(\theta_n^{s})}\log  \frac{p(\theta_{n}^{s}|M_n^{s})}{q_n^*(\theta_{n}^{s})}+ \frac{1}{N_s}\E_{q_n(X_n; \lambda_{n})}\log\frac{\hat{p}_n(X_n; \eta_{n}^{s})}{q_n(X_n;\lambda_{n})}.
\end{align}
Consequently, all sensors collaboratively optimise a different objective $\mathcal{L}'(\lambda_n)$, which is the sum of local objectives $\mathcal{L}'_s(\lambda_n)$:
\begin{align}  
\mathcal{L}'(\lambda_{n})& = \sum_{s=1}^{N_s}\mathcal{L}'_s(\lambda_{n})\\
&=\sum_{s=1}^{N_s} \E_{q_n(X_n; \lambda_{n})q_n^*(\theta_n)}\log p(Y_{n}^{s}|\theta_{n}^{s},X_{n})+ 
   \E_{q_n^*(\theta_n)}\log\frac{p(\theta_{n}|M_n)}{q_n^*(\theta_{n})} + \sum_{s=1}^{N_s}  \frac{1}{N_s}\E_{q_n(X_n; \lambda_{n})}\log\frac{\hat{p}_n(X_n; \eta_{n}^{s})}{q_n(X_n;\lambda_{n})}
\end{align}
where the only difference from $\mathcal{L}(\lambda_{n})$ in \eqref{eq:locally maximised ELBO form} (or equivalently \eqref{eq: apx repeat LM-ELBO}) is in the last term, which can be rewritten as follows
\begin{align}  
\sum_{s=1}^{N_s}  \frac{1}{N_s}\E_{q_n(X_n; \lambda_{n})}\log\frac{\hat{p}_n(X_n; \eta_{n}^{s})}{q_n(X_n;\lambda_{n})} & = \E_{q_n(X_n; \lambda_{n})}  \sum_{s=1}^{N_s} \log\frac{\hat{p}_n(X_n; \eta_{n}^{s})^{\frac{1}{N_s}}}{q_n(X_n;\lambda_{n})^{\frac{1}{N_s}}}\\
& = \E_{q_n(X_n; \lambda_{n})}  \log\frac{ \prod_{s=1}^{N_s}\hat{p}_n(X_n; \eta_{n}^{s})^{\frac{1}{N_s}}}{q_n(X_n;\lambda_{n})}\\
& = \E_{q_n(X_n; \lambda_{n})}  \log\frac{ \hat{p}_{eff}(X_n)}{q_n(X_n;\lambda_{n})} + C,
\end{align}
where $\hat{p}_{eff}(X_n)\propto \prod_{s=1}^{N_s}  \hat{p}_n(X_{n};\eta_{n}^{s} ) ^{1/N_s}$ and $C=\log \int \prod_{s=1}^{N_s}  \hat{p}_n(X_{n};\eta_{n}^{s} ) ^{1/N_s} dX_n$ is a constant independent of $X_n$ or $\lambda_n$. Since the constant $C$ can be omitted from the objective function $\mathcal{L}'(\lambda_{n})$, we conclude that our decentralised (natural) gradient-based variational trackers still maximise the same LM-ELBO $\mathcal{L}(\lambda_{n})$ in \eqref{eq:locally maximised ELBO form} (or equivalently \eqref{eq: apx repeat LM-ELBO}), with the only change being the replacement of $\hat{p}_n(X_n)$ by the effective prior $\hat{p}_{eff}(X_n)\propto \prod_{s=1}^{N_s}  \hat{p}_n(X_{n};\eta_{n}^{s} ) ^{1/N_s}$, which is a reasonable geometric average fusion of individual sensors' priors. Therefore, the proposed trackers continue to perform sensible inference at the current time step, even if the sensors' estimates have not fully converged at the previous time step $n-1$ due to the use of limited (natural) gradient descent iterations for efficiency.

% in \eqref{eq:locally maximised ELBO form}, but with a different prior $\hat{p}_{eff}(X_n)\propto \prod_{s=1}^{N_s}  \hat{p}_n(X_{n};\eta_{n}^{s} ) ^{1/N_s}$ in place of $\hat{p}_n(X_n)$ in \eqref{eq:locally maximised ELBO form}.

% under limited DNGD iterations when sensors may hold unique priors $\hat{p}_n(X_{n};\eta_{n}^{s})$, it still maximises LM-ELBO in (36) with 
% $\hat{p}_n(X_n; \eta_{n})$ being replaced by an effective prior $\hat{p}_{eff}(X_n)\propto \prod_{s=1}^{N_s}  \hat{p}_n(X_{n};\eta_{n}^{s} ) ^{1/N_s}$, which is the geometric average fusion of individual sensors' priors. 

% \bibliographystyle{IEEEtran}
% \bibliography{./IEEEsupp}  

\begin{thebibliography}{10}
\providecommand{\url}[1]{#1}
\csname url@samestyle\endcsname
\providecommand{\newblock}{\relax}
\providecommand{\bibinfo}[2]{#2}
\providecommand{\BIBentrySTDinterwordspacing}{\spaceskip=0pt\relax}
\providecommand{\BIBentryALTinterwordstretchfactor}{4}
\providecommand{\BIBentryALTinterwordspacing}{\spaceskip=\fontdimen2\font plus
\BIBentryALTinterwordstretchfactor\fontdimen3\font minus \fontdimen4\font\relax}
\providecommand{\BIBforeignlanguage}[2]{{%
\expandafter\ifx\csname l@#1\endcsname\relax
\typeout{** WARNING: IEEEtran.bst: No hyphenation pattern has been}%
\typeout{** loaded for the language `#1'. Using the pattern for}%
\typeout{** the default language instead.}%
\else
\language=\csname l@#1\endcsname
\fi
#2}}
\providecommand{\BIBdecl}{\relax}
\BIBdecl

\bibitem{rao1993fully}
B.~Rao, H.~F. Durrant-Whyte, and J.~Sheen, ``A fully decentralized multi-sensor system for tracking and surveillance,'' \emph{The International Journal of Robotics Research}, vol.~12, no.~1, pp. 20--44, 1993.

\bibitem{olfati2005distributed}
R.~Olfati-Saber, ``Distributed {K}alman filter with embedded consensus filters,'' in \emph{Proceedings of the 44th IEEE Conference on Decision and Control}.\hskip 1em plus 0.5em minus 0.4em\relax IEEE, 2005, pp. 8179--8184.

\bibitem{sandell2008distributed}
N.~F. Sandell and R.~Olfati-Saber, ``Distributed data association for multi-target tracking in sensor networks,'' in \emph{2008 47th IEEE Conference on Decision and Control}.\hskip 1em plus 0.5em minus 0.4em\relax IEEE, 2008, pp. 1085--1090.

\bibitem{chong2017forty}
C.-Y. Chong, ``Forty years of distributed estimation: A review of noteworthy developments,'' in \emph{2017 Sensor Data Fusion: Trends, Solutions, Applications (SDF)}.\hskip 1em plus 0.5em minus 0.4em\relax IEEE, 2017, pp. 1--10.

\bibitem{coates2004distributed}
M.~Coates, ``Distributed particle filters for sensor networks,'' in \emph{Proceedings of the 3rd international symposium on Information processing in sensor networks}, 2004, pp. 99--107.

\bibitem{oreshkin2010asynchronous}
B.~N. Oreshkin and M.~J. Coates, ``Asynchronous distributed particle filter via decentralized evaluation of {G}aussian products,'' in \emph{2010 13th International Conference on Information Fusion}.\hskip 1em plus 0.5em minus 0.4em\relax IEEE, 2010, pp. 1--8.

\bibitem{chong1990distributed}
C.-Y. Chong, ``Distributed multitarget multisensor tracking,'' \emph{Multitarget-multisensor tracking: Advanced applications}, pp. 247--296, 1990.

\bibitem{clark2010robust}
D.~Clark, S.~Julier, R.~Mahler, and B.~Ristic, ``Robust multi-object sensor fusion with unknown correlations,'' 2010.

\bibitem{li2017generalized}
T.~Li, J.~M. Corchado, and S.~Sun, ``On generalized covariance intersection for distributed {PHD} filtering and a simple but better alternative,'' in \emph{2017 20th International Conference on Information Fusion (Fusion)}.\hskip 1em plus 0.5em minus 0.4em\relax IEEE, 2017, pp. 1--8.

\bibitem{li2021distributed}
T.~Li and F.~Hlawatsch, ``A distributed particle-{PHD} filter using arithmetic-average fusion of {G}aussian mixture parameters,'' \emph{Information Fusion}, vol.~73, pp. 111--124, 2021.

\bibitem{ueney2013distributed}
M.~Ueney, D.~E. Clark, and S.~J. Julier, ``Distributed fusion of {PHD} filters via exponential mixture densities,'' \emph{IEEE Journal of Selected Topics in Signal Processing}, vol.~7, no.~3, pp. 521--531, 2013.

\bibitem{li2020arithmetic}
T.~Li, X.~Wang, Y.~Liang, and Q.~Pan, ``On arithmetic average fusion and its application for distributed multi-{B}ernoulli multitarget tracking,'' \emph{IEEE Transactions on Signal Processing}, vol.~68, pp. 2883--2896, 2020.

\bibitem{li2018computationally}
S.~Li, G.~Battistelli, L.~Chisci, W.~Yi, B.~Wang, and L.~Kong, ``Computationally efficient multi-agent multi-object tracking with labeled random finite sets,'' \emph{IEEE Transactions on Signal Processing}, 2018.

\bibitem{olfati2004consensus}
R.~Olfati-Saber and R.~M. Murray, ``Consensus problems in networks of agents with switching topology and time-delays,'' \emph{IEEE Transactions on automatic control}, vol.~49, no.~9, pp. 1520--1533, 2004.

\bibitem{xiao2005scheme}
L.~Xiao, S.~Boyd, and S.~Lall, ``A scheme for robust distributed sensor fusion based on average consensus,'' in \emph{IPSN 2005. Fourth International Symposium on Information Processing in Sensor Networks, 2005.}\hskip 1em plus 0.5em minus 0.4em\relax IEEE, 2005, pp. 63--70.

\bibitem{gan2022variational}
R.~Gan, Q.~Li, and S.~Godsill, ``A variational {B}ayes association-based multi-object tracker under the non-homogeneous {P}oisson measurement process,'' in \emph{2022 25th International Conference on Information Fusion (FUSION)}.\hskip 1em plus 0.5em minus 0.4em\relax IEEE, 2022, pp. 1--8.

\bibitem{gan2024variational}
R.~Gan, Q.~Li, and S.~J. Godsill, ``Variational tracking and redetection for closely-spaced objects in heavy clutter,'' \emph{IEEE Transactions on Aerospace and Electronic Systems}, 2024.

\bibitem{meyer2020scalable}
F.~Meyer and M.~Z. Win, ``Scalable data association for extended object tracking,'' \emph{IEEE Transactions on Signal and Information Processing over Networks}, vol.~6, pp. 491--507, 2020.

\bibitem{granstrom2019poisson}
K.~Granstr{\"o}m, M.~Fatemi, and L.~Svensson, ``Poisson multi-{B}ernoulli mixture conjugate prior for multiple extended target filtering,'' \emph{IEEE Transactions on Aerospace and Electronic Systems}, vol.~56, no.~1, pp. 208--225, 2019.

\bibitem{li2023adaptive}
Q.~Li, R.~Gan, J.~Liang, and S.~J. Godsill, ``An adaptive and scalable multi-object tracker based on the non-homogeneous {P}oisson process,'' \emph{IEEE Transactions on Signal Processing}, 2023.

\bibitem{gilholm2005poisson}
K.~Gilholm, S.~Godsill, S.~Maskell, and D.~Salmond, ``Poisson models for extended target and group tracking,'' in \emph{Signal and Data Processing of Small Targets 2005}, vol. 5913.\hskip 1em plus 0.5em minus 0.4em\relax International Society for Optics and Photonics, 2005, p. 59130R.

\bibitem{li2023consensus}
Q.~Li, R.~Gan, and S.~Godsill, ``Consensus-based distributed variational multi-object tracker in multi-sensor network,'' in \emph{2023 Sensor Signal Processing for Defence Conference (SSPD)}.\hskip 1em plus 0.5em minus 0.4em\relax IEEE, 2023, pp. 1--5.

\bibitem{li2024}
Q.~Li, R.~Gan, and S.~J. Godsill, ``Decentralised gradient-based variational inference for multi-sensor fusion and tracking in clutter,'' in \emph{in 27th International Conference on Information Fusion}.\hskip 1em plus 0.5em minus 0.4em\relax IEEE, 2024.

\bibitem{hua2015distributed}
J.~Hua and C.~Li, ``Distributed variational {B}ayesian algorithms over sensor networks,'' \emph{IEEE Transactions on Signal Processing}, vol.~64, no.~3, pp. 783--798, 2015.

\bibitem{chang2020distributed}
T.-H. Chang, M.~Hong, H.-T. Wai, X.~Zhang, and S.~Lu, ``Distributed learning in the nonconvex world: From batch data to streaming and beyond,'' \emph{IEEE Signal Processing Magazine}, 2020.

\bibitem{nedic2017achieving}
A.~Nedic, A.~Olshevsky, and W.~Shi, ``Achieving geometric convergence for distributed optimization over time-varying graphs,'' \emph{SIAM Journal on Optimization}, vol.~27, no.~4, pp. 2597--2633, 2017.

\bibitem{amari1998natural}
S.-I. Amari, ``Natural gradient works efficiently in learning,'' \emph{Neural computation}, vol.~10, no.~2, pp. 251--276, 1998.

\bibitem{hensman2012fast}
J.~Hensman, M.~Rattray, and N.~Lawrence, ``Fast variational inference in the conjugate exponential family,'' \emph{Advances in neural information processing systems}, vol.~25, 2012.

\bibitem{hoffman2013stochastic}
M.~D. Hoffman, D.~M. Blei, C.~Wang, and J.~Paisley, ``Stochastic variational inference,'' \emph{Journal of Machine Learning Research}, 2013.

\bibitem{blei2017variational}
D.~M. Blei, A.~Kucukelbir, and J.~D. McAuliffe, ``Variational inference: A review for statisticians,'' \emph{Journal of the American statistical Association}, vol. 112, no. 518, pp. 859--877, 2017.

\bibitem{durante2019conditionally}
D.~Durante and T.~Rigon, ``Conditionally conjugate mean-field variational {B}ayes for logistic models,'' 2019.

\bibitem{king2006fast}
N.~J. King and N.~D. Lawrence, ``Fast variational inference for {G}aussian process models through {KL}-correction,'' in \emph{Machine Learning: ECML 2006: 17th European Conference on Machine Learning Berlin, Germany, September 18-22, 2006 Proceedings 17}.\hskip 1em plus 0.5em minus 0.4em\relax Springer, 2006, pp. 270--281.

\bibitem{lazaro2012overlapping}
M.~L{\'a}zaro-Gredilla, S.~Van~Vaerenbergh, and N.~D. Lawrence, ``Overlapping mixtures of {G}aussian processes for the data association problem,'' \emph{Pattern recognition}, vol.~45, no.~4, pp. 1386--1395, 2012.

\bibitem{hoffman2015structured}
M.~D. Hoffman and D.~M. Blei, ``Structured stochastic variational inference,'' in \emph{Artificial Intelligence and Statistics}, 2015, pp. 361--369.

\bibitem{bonnevie2017difference}
R.~Bonnevie, M.~N. Schmidt \emph{et~al.}, ``Difference-of-convex optimization for variational {KL}-corrected inference in {D}irichlet process mixtures,'' in \emph{2017 IEEE 27th International Workshop on Machine Learning for Signal Processing (MLSP)}.\hskip 1em plus 0.5em minus 0.4em\relax IEEE, 2017, pp. 1--6.

\bibitem{nedic2009distributed}
A.~Nedic and A.~Ozdaglar, ``Distributed subgradient methods for multi-agent optimization,'' \emph{IEEE Transactions on Automatic Control}, vol.~54, no.~1, pp. 48--61, 2009.

\bibitem{zeng2018nonconvex}
J.~Zeng and W.~Yin, ``On nonconvex decentralized gradient descent,'' \emph{IEEE Transactions on signal processing}, vol.~66, no.~11, 2018.

\bibitem{khan2018fast}
M.~Khan, D.~Nielsen, V.~Tangkaratt, W.~Lin, Y.~Gal, and A.~Srivastava, ``Fast and scalable {B}ayesian deep learning by weight-perturbation in adam,'' in \emph{International conference on machine learning}.\hskip 1em plus 0.5em minus 0.4em\relax PMLR, 2018, pp. 2611--2620.

\bibitem{rahmathullah2017generalized}
A.~S. Rahmathullah, {\'A}.~F. Garc{\'\i}a-Fern{\'a}ndez, and L.~Svensson, ``Generalized optimal sub-pattern assignment metric,'' in \emph{2017 20th International Conference on Information Fusion}.\hskip 1em plus 0.5em minus 0.4em\relax IEEE, 2017, pp. 1--8.

\end{thebibliography}


\begin{thebibliography}{1}
\providecommand{\url}[1]{#1}
\csname url@samestyle\endcsname
\providecommand{\newblock}{\relax}
\providecommand{\bibinfo}[2]{#2}
\providecommand{\BIBentrySTDinterwordspacing}{\spaceskip=0pt\relax}
\providecommand{\BIBentryALTinterwordstretchfactor}{4}
\providecommand{\BIBentryALTinterwordspacing}{\spaceskip=\fontdimen2\font plus
\BIBentryALTinterwordstretchfactor\fontdimen3\font minus \fontdimen4\font\relax}
\providecommand{\BIBforeignlanguage}[2]{{%
\expandafter\ifx\csname l@#1\endcsname\relax
\typeout{** WARNING: IEEEtran.bst: No hyphenation pattern has been}%
\typeout{** loaded for the language `#1'. Using the pattern for}%
\typeout{** the default language instead.}%
\else
\language=\csname l@#1\endcsname
\fi
#2}}
\providecommand{\BIBdecl}{\relax}
\BIBdecl

\bibitem{gan2024variationalsuppl}
R.~Gan, Q.~Li, and S.~Godsill, ``Variational tracking and redetection for closely-spaced objects in heavy clutter: Supplementary materials,'' \emph{arXiv preprint arXiv:2309.01774}, 2024.

\bibitem{hensman2012fast}
J.~Hensman, M.~Rattray, and N.~Lawrence, ``Fast variational inference in the conjugate exponential family,'' \emph{Advances in neural information processing systems}, vol.~25, 2012.

\bibitem{hoffman2013stochastic}
M.~D. Hoffman, D.~M. Blei, C.~Wang, and J.~Paisley, ``Stochastic variational inference,'' \emph{Journal of Machine Learning Research}, 2013.

\bibitem{petersen2008matrix}
K.~B. Petersen, M.~S. Pedersen \emph{et~al.}, ``The matrix cookbook,'' \emph{Technical University of Denmark}, vol.~7, no.~15, p. 510, 2008.

\bibitem{khan2018fast}
M.~Khan, D.~Nielsen, V.~Tangkaratt, W.~Lin, Y.~Gal, and A.~Srivastava, ``Fast and scalable bayesian deep learning by weight-perturbation in adam,'' in \emph{International conference on machine learning}.\hskip 1em plus 0.5em minus 0.4em\relax PMLR, 2018, pp. 2611--2620.

\end{thebibliography}
\putbib[IEEEsupp]% Separate bibliography file for the supplemental materials
\end{bibunit}

\end{document}